\documentclass[10pt,a4paper,oneside,notitlepage,fleqn,reqno,english,table]{article}

\usepackage[hidelinks]{hyperref} 
\usepackage{cmap} 				 

\usepackage{multicol} 

\usepackage{longtable} 
\usepackage{multirow} 
\usepackage{rotating} 
\usepackage[cp1252]{inputenc} 
\usepackage{lscape} 
\usepackage{url}    
\usepackage[section]{placeins}  
\usepackage{afterpage} 
\usepackage{capt-of}        
\usepackage{dirtytalk} 

\usepackage[T1]{fontenc}
\usepackage{tabularx, booktabs}   
\usepackage{makecell} 
\usepackage{array} 
\usepackage{xcolor} 
\usepackage{xfrac} 

\graphicspath{{figs/}} 					
\usepackage[parfill]{parskip} 					

\usepackage{amsmath} 			
\usepackage{amsfonts}			
\usepackage{mathtools}			
\everymath{\displaystyle}		
\usepackage{amssymb}  			
\usepackage{nicefrac}  			

\usepackage{tikz}  
\usetikzlibrary{shapes,arrows,calc,shadows,fit,intersections,datavisualization.formats.functions}
\tikzstyle{block}        = [draw, fill=blue!30, rectangle, text centered, minimum height=3em, minimum width=6em] 
\tikzstyle{blockred}     = [draw, fill=red!30, rectangle, text centered, minimum height=3em, minimum width=6em] 
\tikzstyle{blockyellow}  = [draw, fill=yellow!30, rectangle, text centered, minimum height=3em, minimum width=6em] 
\tikzstyle{blockyellow1} = [draw, fill=yellow!15, rectangle, text centered, minimum height=3em, minimum width=6em] 
\tikzstyle{blockyellow2} = [draw, fill=yellow!30, rectangle, text centered, minimum height=3em, minimum width=6em] 
\tikzstyle{blockyellow3} = [draw, fill=yellow!45, rectangle, text centered, minimum height=3em, minimum width=6em] 
\tikzstyle{blockyellow4} = [draw, fill=yellow!60, rectangle, text centered, minimum height=3em, minimum width=6em] 
\tikzstyle{blockgreen}   = [draw, fill=green!30, rectangle, text centered, minimum height=3em, minimum width=6em] 
\tikzstyle{blockgreen1}  = [draw, fill=green!15, rectangle, text centered, minimum height=3em, minimum width=6em] 
\tikzstyle{blockgreen2}  = [draw, fill=green!30, rectangle, text centered, minimum height=3em, minimum width=6em] 
\tikzstyle{blockgreen3}  = [draw, fill=green!45, rectangle, text centered, minimum height=3em, minimum width=6em] 
\tikzstyle{blockgreen4}  = [draw, fill=green!60, rectangle, text centered, minimum height=3em, minimum width=6em] 
\tikzstyle{blockgrey1}   = [draw, fill=black!8,  rectangle, text centered, minimum height=3em, minimum width=6em] 
\tikzstyle{blockgrey2}   = [draw, fill=black!16, rectangle, text centered, minimum height=3em, minimum width=6em] 
\tikzstyle{blockgrey3}   = [draw, fill=black!24, rectangle, text centered, minimum height=3em, minimum width=6em] 
\tikzstyle{blockgrey4}   = [draw, fill=black!32, rectangle, text centered, minimum height=3em, minimum width=6em] 
\tikzstyle{blockgrey5}   = [draw, fill=black!40, rectangle, text centered, minimum height=3em, minimum width=6em] 
\tikzstyle{blocknofill}  = [draw=black!50, line width=1.5pt, rectangle, rounded corners, text centered, minimum height=3em, minimum width=6em]
\tikzstyle{blockhigh}    = [draw, fill=blue!20, rectangle, text centered, minimum height=6em, minimum width=6em]
\tikzstyle{noblock}      = [draw=black!50, line width=1.5pt, rectangle, rounded corners, text centered, minimum height=3em, minimum width=6em]
\tikzstyle{sum}          = [draw, fill=blue!20, circle, node distance=1cm]
\tikzstyle{sumrest}      = [draw, fill=black, circle, radius=0.5cm]
\tikzstyle{pinstyle}     = [pin edge={to-,thin,black}]
\tikzstyle{title}        = [text centered]
\pgfdeclarelayer{background}
\pgfdeclarelayer{foreground}
\pgfsetlayers{background,main,foreground}

\usepackage{pgfplots}
\pgfplotsset{compat=1.13}
\pgfplotsset{colormap/bluered}
\usepackage{pgfplotstable}
\usepgfplotslibrary{polar}
\usepgfplotslibrary{colormaps}
\usepgfplotslibrary{external}

\usepackage{graphicx, color} 	
\usepackage{epstopdf}  			

\allowdisplaybreaks 

\newcommand{\sss}   {\scriptscriptstyle}    

\newcommand{\nm}   		[1] {\ensuremath{\mathrm{#1}}} 									
\newcommand{\neweq}     [2] {\begin{equation} \mathrm{#1}\label{#2} \end{equation}} 	

\newcommand{\CC}		{{C\nolinebreak[4]\hspace{-.05em}\raisebox{.4ex}{\tiny\bf ++}}}

\renewcommand{\vec}		[1] {\mbox{\boldmath{\ensuremath{\mathrm{#1}}}}}	
\newcommand{\deriv}     [1] {\dfrac{d{#1}}{dt}}             				
\newcommand{\lrp}       [1] {\left(#1\right)}								
\newcommand{\lrsb}      [1] {\left[#1\right]}								

\newcommand{\muj}        [1] {\mu_{{#1}j}}
\newcommand{\sigmaj}     [1] {\sigma_{{#1}j}}

\newcommand{\mun}        [1] {\mu_{{#1}n}}
\newcommand{\sigman}     [1] {\sigma_{{#1}n}}

\newcommand{\DeltarBest}        {\hat{\vec r}^{\sss B}}
\newcommand{\DeltarBestnorm}    {\|\hat{\vec r}^{\sss B}\|}

\newcommand{\Deltat}     	{\Delta t}

\newcommand{\DeltatEST}  	{\Delta t_{\sss EST}}

\newcommand{\second}        {\nm{2^{\sss nd}}}   

\newcommand{\deltaCNTR}  	{\vec{\delta}_{\sss CNTR}}		
\newcommand{\deltaTARGET}  	{\vec{\delta}_{\sss TARGET}}	    

\newcommand{\xvec}   			{\vec x}
\newcommand{\xveczero}	 		{\vec x_0}
\newcommand{\xvecestzero}		{\hat{\vec x}_0}

\newcommand{\xvecest}	  		{\hat{\vec x}}
\newcommand{\xvectilde} 		{\widetilde{\vec x}}

\newcommand{\xTRUTH}			{\vec x_{\sss TRUTH}}
\newcommand{\xSENSED}  			{\vec x_{\sss SENSED}}
\newcommand{\xEST}  			{\vec x_{\sss EST}}
\newcommand{\xREF}	  			{\vec x_{\sss REF}}

\newcommand{\FE}		 {F_{\sss E}}

\newcommand{\iEiii}      {\vec i_{\sss3}^{\sss E}}

\newcommand{\xEgdt}      {\vec x_{\sss GDT}}

\newcommand{\FN} 		  {F_{\sss N}}

\newcommand{\FB}		  {F_{\sss B}}

\newcommand{\FI}		 {F_{\sss I}}							

\newcommand{\gN}          {\vec g^{\sss N}}

\newcommand{\ac}          {\vec a_c}
\newcommand{\gc}     	  {\vec g_c}

\newcommand{\gcN}         {\vec g_c^{\sss N}}
\newcommand{\gcNMODEL}    {\vec g_{c,\sss{MOD}}^{\sss N}}
\newcommand{\gcNMODELest} {\hat{\vec g}_{c,\sss{MOD}}^{\sss N}}
\newcommand{\gcNREAL}     {\vec g_{c,\sss{REAL}}^{\sss N}}

\newcommand{\vvec} 				{\vec v}
\newcommand{\vEN}  				{\vec v_{\sss EN}}
\newcommand{\vEB}  				{\vec v_{\sss EB}}

\newcommand{\vB} 	   			{\vec v^{\sss B}}

\newcommand{\vN}	    		{\vec v^{\sss N}}
\newcommand{\vNi}        		{v_{\sss 1}^{\sss N}}
\newcommand{\vNii}      	  	{v_{\sss 2}^{\sss N}}
\newcommand{\vNiii} 	      	{v_{\sss 3}^{\sss N}}

\newcommand{\vIB}           	{\vec v_{\sss IB}}

\newcommand{\vEBI}				{\vec v_{\sss EB}^{\sss I}}

\newcommand{\vEBB}				{\vec v_{\sss EB}^{\sss B}}
\newcommand{\vEBN}				{\vec v_{\sss EB}^{\sss N}}
\newcommand{\vENN}				{\vec v_{\sss EN}^{\sss N}}

\newcommand{\vIBIdot}			{\dot{\vec v}_{\sss IB}^{\sss I}}
\newcommand{\vEBEdot}			{\dot{\vec v}_{\sss EB}^{\sss E}}
\newcommand{\vEBBdot}			{\dot{\vec v}_{\sss EB}^{\sss B}}
\newcommand{\vENEdot}			{\dot{\vec v}_{\sss EN}^{\sss E}}
\newcommand{\vENNdot}			{\dot{\vec v}_{\sss EN}^{\sss N}}
\newcommand{\vBdot}				{\dot{\vec v}^{\sss B}}

\newcommand{\vNdot}	    		{\dot{\vec v}^{\sss N}}
\newcommand{\vNest}	    		{\hat{\vec v}^{\sss N}}

\newcommand{\vNesti}       		{\hat{v}_{\sss 1}^{\sss N}}
\newcommand{\vNestii}      	  	{\hat{v}_{\sss 2}^{\sss N}}
\newcommand{\vNestiii} 	      	{\hat{v}_{\sss 3}^{\sss N}}
\newcommand{\vNdotest}			{\hat{\dot{\vec v}}^{\sss N}}

\newcommand{\hdotest}           {\hat{\dot{h}}}

\newcommand{\vtas}          {v_{\sss TAS}}
\newcommand{\vtasINI}       {v_{\sss TAS,INI}}
\newcommand{\vtasEND}       {v_{\sss TAS,END}}
\newcommand{\vtastilde}     {\widetilde{v}_{\sss TAS}}

\newcommand{\vTASBdot}      {\dot{\vec v}_{\sss TAS}^{\sss B}}
\newcommand{\vTASNdot}      {\dot{\vec v}_{\sss TAS}^{\sss N}}
\newcommand{\vtasest}       {\hat{v}_{\sss TAS}}

\newcommand{\vtasdotest}    {\hat{\dot v}_{\sss TAS}}
\newcommand{\vTASN}    		{\vec v_{\sss TAS}^{\sss N}}
\newcommand{\vTASB}    		{\vec v_{\sss TAS}^{\sss B}}

\newcommand{\vTASBest} 		{\hat{\vec v}_{\sss TAS}^{\sss B}}
\newcommand{\vTASNest} 		{\hat{\vec v}_{\sss TAS}^{\sss N}}
\newcommand{\vTASNestiii}  	{\hat{v}_{\sss TAS,3}^{\sss N}}
\newcommand{\vTASNdotest}   {\hat{\dot{\vec v}}_{\sss TAS}^{\sss N}}
\newcommand{\vTASBdotest}   {\hat{\dot{\vec v}}_{\sss TAS}^{\sss B}}

\newcommand{\vWIND}       	{\vec v_{\sss WIND}}

\newcommand{\vWINDB}      	{\vec v_{\sss WIND}^{\sss B}}
\newcommand{\vWINDN}      	{\vec v_{\sss WIND}^{\sss N}}
\newcommand{\vWINDNdot}     {\dot{\vec v}_{\sss WIND}^{\sss N}}
\newcommand{\vWINDBdot}     {\dot{\vec v}_{\sss WIND}^{\sss B}}

\newcommand{\vWINDNest}    	{\hat{\vec v}_{\sss WIND}^{\sss N}}
\newcommand{\vwindINI}     	{v_{\sss WIND,INI}}
\newcommand{\vwindEND}     	{v_{\sss WIND,END}}

\newcommand{\vWINDNINI}     	{\vec v_{\sss WIND,INI}^{\sss N}}
\newcommand{\vWINDNEND}     	{\vec v_{\sss WIND,END}^{\sss N}}

\newcommand{\vWINDNdotest}      {\hat{\dot{\vec v}}_{\sss WIND}^{\sss N}}

\newcommand{\vTURBBdot}   {\dot{\vec v}_{\sss TURB}^{\sss B}}
\newcommand{\vTURBNdot}   {\dot{\vec v}_{\sss TURB}^{\sss N}}
\newcommand{\vTURBN}      {\vec v_{\sss TURB}^{\sss N}}

\newcommand{\vTURBB}      {\vec v_{\sss TURB}^{\sss B}}

\newcommand{\omegaE}		{\omega_{\sss E}}

\newcommand{\wIE}			{\vec \omega_{\sss IE}}

\newcommand{\wIEN}			{\vec \omega_{\sss IE}^{\sss N}}

\newcommand{\wIEIskew}		{\widehat{\vec \omega}_{\sss IE}^{\sss I}}
\newcommand{\wIENskew}		{\widehat{\vec \omega}_{\sss IE}^{\sss N}}
\newcommand{\alphaIEIskew} 	{\widehat{\vec \alpha}_{\sss IE}^{\sss I}}

\newcommand{\wEN}			{\vec \omega_{\sss EN}}

\newcommand{\wENN}			{\vec \omega_{\sss EN}^{\sss N}}
\newcommand{\wENNskew}		{\widehat{\vec \omega}_{\sss EN}^{\sss N}}
\newcommand{\wENNest}		{\hat{\vec \omega}_{\sss EN}^{\sss N}}
\newcommand{\wENNestskew}	{\widehat{\hat{\vec \omega}}_{\sss EN}^{\sss N}}

\newcommand{\wEB}			{\vec \omega_{\sss EB}}

\newcommand{\wEBBskew}		{\widehat{\vec \omega}_{\sss EB}^{\sss B}}

\newcommand{\wNB}			{\vec \omega_{\sss NB}}

\newcommand{\wNBB}			{\vec \omega_{\sss NB}^{\sss B}}

\newcommand{\wNBBskew}		{\widehat{\vec \omega}_{\sss {NB}}^{\sss B}}
\newcommand{\wNBBest}		{\hat{\vec \omega}_{\sss NB}^{\sss B}}

\newcommand{\wIBB}				{\vec \omega_{\sss IB}^{\sss B}}

\newcommand{\wIBBtilde}			{\widetilde{\vec \omega}_{\sss IB}^{\sss B}}

\newcommand{\atrs}			{\vec a_{trs}}
\newcommand{\atrsI}			{\vec a_{trs}^{\sss I}}
\newcommand{\atrsN}			{\vec a_{trs}^{\sss N}}
\newcommand{\acor}			{\vec a_{cor}}
\newcommand{\acorN}			{\vec a_{cor}^{\sss N}}
\newcommand{\acorNest}		{\hat{\vec a}_{cor}^{\sss N}}

\newcommand{\REN}			{\vec R_{\sss EN}}

\newcommand{\RIE}			{\vec R_{\sss IE}}

\newcommand{\REB}			{\vec R_{\sss EB}}

\newcommand{\RIB}			{\vec R_{\sss IB}}
\newcommand{\RBI}			{\vec R_{\sss BI}}
\newcommand{\RIN}			{\vec R_{\sss IN}}
\newcommand{\RNI}			{\vec R_{\sss NI}}

\newcommand{\REBdot}		{\dot{\vec R}_{\sss EB}}
\newcommand{\RENdot}		{\dot{\vec R}_{\sss EN}}

\newcommand{\qNB}			{\vec q_{\sss NB}}

\newcommand{\qNBdot}		{\dot{\vec q}_{\sss NB}}

\newcommand{\qNBast}		{\vec q_{\sss NB}^{\ast}}

\newcommand{\qNBest}		{\hat{\vec q}_{\sss NB}}
\newcommand{\qNBastest}		{\hat{\vec q}_{\sss NB}^{\ast}}

\newcommand{\chiWINDINI}       	{\chi_{\sss WIND,INI}}
\newcommand{\chiWINDEND}       	{\chi_{\sss WIND,END}}

\newcommand{\psiest}			{\hat \psi}

\newcommand{\thetaest}			{\hat \theta}

\newcommand{\xiest}				{\hat \xi}

\newcommand{\alphaest}          {\hat{\alpha}}

\newcommand{\alphadotest}       {\hat{\dot \alpha}}
\newcommand{\betaest}			{\hat \beta}

\newcommand{\betadotest}        {\hat{\dot \beta}}

\newcommand{\alphatilde}        {\widetilde{\alpha}}
\newcommand{\betatilde}         {\widetilde{\beta}}
\newcommand{\chiINI}            {\chi_{\sss INI}}
\newcommand{\chiEND}            {\chi_{\sss END}}

\newcommand{\xiTURN}            {\xi_{\sss TURN}}

\newcommand{\gammaTASCLIMB}     {\gamma_{\sss TAS,CLIMB}}

\newcommand{\tTURN}				{t_{\sss TURN}}

\newcommand{\tGNSS}				{t_{\sss GNSS}}
\newcommand{\tEND}				{t_{\sss END}}

\newcommand{\Hp}            {H_{\sss P}}
\newcommand{\HpINI}         {H_{\sss P,INI}}
\newcommand{\HpEND}         {H_{\sss P,END}}

\newcommand{\Hpest}         {\hat{H}_{\sss P}}

\newcommand{\DeltaT}        {\Delta T}
\newcommand{\Deltap}        {\Delta p}
\newcommand{\Ttilde}    	{\widetilde{T}}
\newcommand{\ptilde} 	    {\widetilde{p}}
\newcommand{\Test}			{\hat{T}}
\newcommand{\hest}			{\hat{h}}

\newcommand{\DeltaTest}		{\Delta\hat{T}}
\newcommand{\Deltapest}		{\Delta\hat{p}}
\newcommand{\DeltaTINI}     {\Delta T_{\sss INI}}
\newcommand{\DeltaTEND}     {\Delta T_{\sss END}}

\newcommand{\DeltapINI}     {\Delta p_{\sss INI}}
\newcommand{\DeltapEND}     {\Delta p_{\sss END}}

\newcommand{\Tzero}         {T_{\sss 0}}                 
\newcommand{\pzero}         {p_{\sss 0}}                 

\newcommand{\betaT}         {\beta_{\sss T}}         	
\newcommand{\gzero}     	{g_{\sss 0}} 				
\newcommand{\RE}     	    {R_{\sss E}}   				
\newcommand{\gBR}           {-\dfrac{\gzero}{\betaT R}}	

\newcommand{\FAER}  		{\vec F_{\sss AER}}
\newcommand{\FAERB}  		{\vec F_{\sss AER}^{\sss B}}

\newcommand{\FPRO}  		{\vec F_{\sss PRO}}
\newcommand{\FPROB}  		{\vec F_{\sss PRO}^{\sss B}}

\newcommand{\fIBB}				{\vec f_{\sss IB}^{\sss B}}
\newcommand{\fIBN}				{\vec f_{\sss IB}^{\sss N}}
\newcommand{\fIBBtilde}			{\widetilde{\vec f}_{\sss IB}^{\sss B}}

\newcommand{\fIBBest}			{\hat{\vec f}_{\sss IB}^{\sss B}}

\newcommand{\EACC}		    {\vec E_{\sss ACC}}

\newcommand{\EACCest}	    {\hat{\vec E}_{\sss ACC}}

\newcommand{\BzeroACC}		{B_{0\sss{ACC}}}

\newcommand{\sigmauACC}		{\sigma_{u\sss{ACC}}}
\newcommand{\sigmavACC}		{\sigma_{v\sss{ACC}}}
\newcommand{\EGYR}			{\vec E_{\sss GYR}}

\newcommand{\EGYRii}		{E_{\sss GYR,2}}

\newcommand{\EGYRest}		{\hat{\vec E}_{\sss GYR}}

\newcommand{\EGYRestii}		{\hat{E}_{\sss GYR,2}}

\newcommand{\BzeroGYR}		{B_{0\sss{GYR}}}

\newcommand{\sigmauGYR}		{\sigma_{u\sss{GYR}}}
\newcommand{\sigmavGYR}		{\sigma_{v\sss{GYR}}}

\newcommand{\sACC}			{s_{\sss ACC}}

\newcommand{\sGYR}			{s_{\sss GYR}}

\newcommand{\mACC}			{m_{\sss ACC}}

\newcommand{\mGYR}			{m_{\sss GYR}}

\newcommand{\sigmaAOA}		{\sigma_{\sss AOA}}
\newcommand{\sigmaAOS}		{\sigma_{\sss AOS}}
\newcommand{\sigmaOSP}		{\sigma_{\sss OSP}}
\newcommand{\sigmaOAT}		{\sigma_{\sss OAT}}
\newcommand{\sigmaTAS}		{\sigma_{\sss TAS}}
\newcommand{\BzeroAOA}		{B_{0\sss{AOA}}}
\newcommand{\BzeroAOS}		{B_{0\sss{AOS}}}
\newcommand{\BzeroOSP}		{B_{0\sss{OSP}}}
\newcommand{\BzeroOAT}		{B_{0\sss{OAT}}}
\newcommand{\BzeroTAS}		{B_{0\sss{TAS}}}

\newcommand{\BNMODEL}    	{\vec B_{\sss MOD}^{\sss N}}

\newcommand{\BNREAL}    	{\vec B_{\sss REAL}^{\sss N}}

\newcommand{\BNDEV}			{\vec B_{\sss DEV}^{\sss N}}

\newcommand{\BNDEVest}		{\hat{\vec B}_{\sss DEV}^{\sss N}}

\newcommand{\BBtilde}		{\widetilde{\vec B}^{\sss B}}

\newcommand{\EMAG}	    		{\vec E_{\sss MAG}}

\newcommand{\EMAGest}	    	{\hat{\vec E}_{\sss MAG}}

\newcommand{\BzeroMAG}			{B_{0,\sss MAG}}

\newcommand{\BhiMAG}			{B_{\sss{HI,MAG}}}

\newcommand{\sigmavMAG}		{\sigma_{v,\sss MAG}}
\newcommand{\sMAG}			{s_{\sss MAG}}
\newcommand{\mMAG}			{m_{\sss MAG}}

\newcommand{\pvec}						{\vec p}
\newcommand{\yvec}						{\vec y}

\newcommand{\Deltaxhorest}				{\Delta \hat{x}_{\sss HOR}}

\newcommand{\Deltaxlongest}				{\Delta \hat{x}_{\sss LONG}}

\newcommand{\Deltaxcrossest}			{\Delta \hat{x}_{\sss CROSS}}

\begin{document}

\title{Minimization of GNSS-Denied Inertial Navigation Errors for Fixed Wing Autonomous Unmanned Air Vehicles}
\author{Eduardo Gallo \footnote{Eduardo Gallo is a PhD. student for the Polytechnic University of Madrid (Faculty of Mechanical Engineering, Center for Automation and Robotics) working on his thesis titled ``Autonomous Unmanned Air Vehicle GNSS-Denied Navigation'', advised by Dr. Antonio Barrientos. Eduardo Gallo holds a MSc. in Aerospace Engineering by the Polytechnic University of Madrid and has twenty-two years of experience working on aircraft trajectory prediction, modeling, and flight simulation. He is currently a Senior Trajectory Prediction and Aircraft Performance Engineer at Boeing Research \& Technology Europe (BR\&T-E), although he is publishing this article in his individual capacity and time as part of his PhD thesis.} \footnote{Contact: edugallo@yahoo.com, e.gallo@alumnos.upm.es, \url{https://orcid.org/0000-0002-7397-0425}} \footnote{Postal address: Juan Ramon Jimenez 4 - 3 - 3A, Las Rozas, Madrid, 28232, Spain.} \ and Antonio Barrientos\footnote{Antonio Barrientos received his MSc. and PhD. Degrees from the Polytechnic University of Madrid in 1982 and 1986, respectively. In 2002 he obtained the MSc Degree in Biomedical Engineering by the National University of Distance Education. Since 1988 he is a professor at the Polytechnic University of Madrid, where he is presently a full professor teaching robotics and automatic control. He has worked in robotics for more than 30 years, developing industrial and service robots for different areas. He is a permanent staff member of the Center for Automation and Robotics (UPM-CSIC) where he is the head of the Applied Robotics Unit. He is a Senior Member of IEEE and former member of the steering committee of CEA (Spanish Automatics Committee - IFAC Spanish chapter).} \footnote{Contact: antonio.barrientos@upm.es, \url{https://orcid.org/0000-0003-1691-3907}}} 
\date{August 2021}
\maketitle


\section*{Abstract}

This article proposes an inertial navigation algorithm intended to lower the negative consequences of the absence of GNSS (Global Navigation Satellite System) signals on the navigation of autonomous fixed wing low SWaP (Size, Weight, and Power) UAVs (Unmanned Air Vehicles). In addition to accelerometers and gyroscopes, the filter takes advantage of sensors usually present onboard these platforms, such as magnetometers, Pitot tube, and air vanes, and aims to minimize the attitude error and reduce the position drift (both horizontal and vertical) with the dual objective of improving the aircraft GNSS-Denied inertial navigation capabilities as well as facilitating the fusion of the inertial filter with visual odometry algorithms. Stochastic high fidelity Monte Carlo simulations of two representative scenarios involving the loss of GNSS signals are employed to evaluate the results, compare the proposed filter with more traditional implementations, and analyze the sensitivity of the results to the quality of the onboard sensors. The author releases the \nm{\CC} implementation of both the navigation filter and the high fidelity simulation as open-source software \cite{Gallo2020_simulation}.


\textbf{\emph{Keywords}}: GNSS-Denied, GPS-Denied, inertial navigation, autonomous navigation, UAV


\section{Introduction and Outline}\label{sec:Outline}

The extreme dependency of an autonomous UAV (Unmanned Air Vehicle) inertial navigation to the presence of GNSS (Global Navigation Satellite System) signals, without which it incurs in a slow but unavoidable position drift that may ultimately lead to the loss of the platform, is discussed in section \ref{sec:GNSS-Denied}. This dependency does not only constitute a significant obstacle for the widespread usage of these platforms in civil airspace, but also represents a threat to military missions.

This article focuses on autonomous fixed wing low SWaP (Size, Weight, and Power) aircraft, and proposes a navigation filter specifically designed for GNSS-Denied conditions. The proposed filter takes advantage of sensors usually present onboard the aircraft but only used for control purposes instead of navigation, the particularities of fixed wing flight, and the atmospheric and wind estimations that can be obtained before the GNSS signals are lost. The objectives of the proposed navigation system are two-fold. The first is to diminish the GNSS-Denied position drift so the vehicle has a higher probability of returning to base, although the drift can not be fully eliminated without the use of additional sensors. The second objective is to improve the GNSS-Denied estimation of the aircraft attitude, specially during maneuvers, with the aim of facilitating the addition of cameras and visual odometry algorithms to the navigation system, which constitutes one of the most promising approaches to further reduce the position drift. The objectives and novelty of the proposed filter are described in section \ref{sec:Objectives} in the context of the different existing approaches to improving GNSS-Denied navigation. 

Section \ref{sec:Simulation} introduces the stochastic high fidelity simulation employed to evaluate the navigation results by means of Monte Carlo executions of two scenarios representative of the challenges of GNSS-Denied navigation, which are described in detail in \cite{SIMULATION}. Its reading is highly recommended in order to understand the rest of this article, as is that of \cite{SENSORS}, which describes the different error sources modeled for each of the onboard sensors. 

Section \ref{sec:Filter} describes the proposed inertial navigation system, characterized by its separation into three complimentary filters, an implementation of the specific force observation equation that minimizes the negative effects of the lack of position and ground velocity GNSS measurements, and the freezing of the values for the wind field and atmospheric pressure offset estimations at the time the GNSS signals are lost.

A detailed analysis of the ability of the filter to track the aircraft attitude, its ground velocity, as well as its vertical and horizontal positions, is presented in section \ref{sec:Results}. They are compared in section \ref{sec:Comparison} with the results obtained based on more traditional approaches employed to estimate the aircraft pose (attitude plus position), representative of inertial filters designed to work with GNSS signals but that can no longer rely on them. 

The analysis of the results concludes in section \ref{sec:Influence}, which discusses the influence of the quality or grade of the different sensors on the aircraft GNSS-Denied inertial navigation capabilities; by ensuring that the results obtained when employing sensors of inferior quality are qualitatively the same and quantitatively slightly inferior to those presented in section \ref{sec:Results}, the results are safeguarded from possible errors introduced when modeling the performances of the various sensors. Last, section \ref{sec:Conclusions} presents the conclusions summarizing the results.


\section{GNSS-Denied Navigation}\label{sec:GNSS-Denied}

An autonomous UAV relies on onboard computers to execute the previously uploaded mission objectives. It can employ the communications channel to provide information to the ground, which may decide to update the mission and upload it to the platform, but as its own name implies, it can also operate without any kind of communication with its operator. It just continues executing its mission until the flight concludes or the mission is modified.

The number, variety, and applications of UAVs have grown exponentially in the last few years, and the trend is expected to continue in the future. This is particularly true in the case of low SWaP vehicles because their reduced cost makes them suitable for a wide range of applications, both civil and military. A significant percentage of these vehicles are capable of operating autonomously. With small variations, these platforms rely on a suite of sensors that continuously provides noisy data about the airframe state, a navigation algorithm to estimate the aircraft pose (position plus attitude), and a control system that, based on the navigation output, adjusts the aircraft control mechanisms to successfully execute the preloaded mission.

This article focuses on fixed wing autonomous platforms, which are generally equipped with a GNSS receiver, accelerometers, gyroscopes, magnetometers, a Pitot tube, and air vanes. The combination of accelerometers and gyroscopes is known as the Inertial Measurement Unit (IMU). The errors introduced by all sensors grow significantly as their SWaP decreases, in particular in the case of the IMU. The recent introduction of solid state accelerometers and gyroscopes has dramatically improved the performance of low SWaP IMUs, with new models showing significant improvements when compared to those fabricated only a few years ago. The problem of noisy measurement readings is compounded in the case of low SWaP vehicles by the low mass of the platforms, which results in less inertia and hence more high frequency accelerations and rotations caused by the atmospheric turbulence.

Aircraft navigation has traditionally relied on the measurements provided by accelerometers, gyroscopes, and magnetometers, incurring in an slow but unbounded position drift that could only be stopped by triangulation with the use of external navigation (radio) aids. More recently, the introduction of satellite navigation has completely removed the position drift and enabled autonomous inertial navigation in low SWaP platforms \cite{Farrell2008, Groves2008, Chatfield1997}. On the negative side, low SWaP inertial navigation exhibits an extreme dependency on the availability of GNSS signals. If not present, inertial systems rely on dead reckoning, which results in velocity and position drift, with the aircraft slowly but steadily deviating from its intended route.

The availability of GNSS signals cannot be guaranteed by any means. In addition to the (unlikely) event of one or various GNSS satellites ceasing to broadcast (voluntarily or not), the GNSS signals can be accidentally blocked by nearby objects (mountains, buildings, trees), corrupted by their own reflections in those same objects, or maliciously interfered with by broadcasting a different signal in the same frequency (jamming or spoofing). Any of the above results in what is known as \emph{GNSS-Denied navigation}. In that event, the vehicle is unable to fly its intended route or even return to a safe recovery location, which leads to the uncontrolled loss of the airframe if the GNSS signals are not recovered before the aircraft runs out of fuel (or battery in case of electric vehicles). 

The extreme dependency on GNSS availability is not only one of the main impediments for the introduction of small UAVs in civil airspace, where it is not acceptable to have uncontrolled vehicles causing personal or material damage, but it also presents a significant drawback for military applications, as a single hull loss may compromise the onboard technology. At this time there are no comprehensive solutions to the operation of low SWaP autonomous UAVs in GNSS-Denied scenarios, although the use of onboard cameras to provide an additional relative pose measurement seems to be one of the most promising routes. Bigger and more expensive UAVs, this is, with less stringent SWaP requirements, can rely to some degree on more accurate IMUs (at the expense of SWaP) and additional communications equipment to overcome this problem, but for most autonomous UAVs, the permanent loss of the GNSS signal is equivalent to losing the airframe in an uncontrolled way.


\section{Objectives and Novelty}\label{sec:Objectives}

The main objective of this article is to improve the GNSS-Denied inertial navigation capabilities of autonomous fixed wing low SWaP vehicles, so in case GNSS signals become unavailable, they can continue their mission or safely fly to a predetermined recovery location. This article does not only pursue the diminution of the position drift inherent to the lack of GNSS signals, but also the facilitation of the fusion of the inertial navigation system with visual odometry algorithms. Three remarks should be added to this objective:
\begin{itemize}
\item The focus on low SWaP autonomous UAVs rules out the use of high quality sensors, which in general are bigger and have more weight, heavier platforms with more inertia against atmospheric turbulence, as well as the assistance of any kind of communications between the platform and the ground. Although the proposed algorithms would work if installed onboard higher SWaP platforms, there may exist better solutions for high SWaP GNSS-Denied navigation that take advantage of the better quality of the onboard sensors as well as the higher platform inertia against turbulence.

\item Platforms that generate their lift by means of rotating blades (helicopters and multirotors) are excluded as the proposed algorithms take advantage of the special characteristics of fixed wing flight, as explained below. Vertical Take Off and Landing (VTOL) as well as Short Take Off and Landing (STOL) platforms, which use their rotors to generate lift when the airspeed is low, but are capable of rotating them to fly like conventional fixed wing aircraft, can rely on the proposed algorithms only when behaving like fixed wing platforms.

\item Although the modeled platform employs a piston engine to power the propeller, there would be no significant difference in the results if it were replaced by an electric motor.
\end{itemize}

\emph{Inertial navigation} employs the periodic readings provided by triads of accelerometers and gyroscopes (known as inertial sensors) to estimate the pose (position and attitude) of a moving object by means of dead reckoning or integration. On aircraft, inertial sensors are complemented by magnetometers and a barometer to add robustness to the inertial solution. Fixed wing aircraft are also equipped with a Pitot tube and air vanes required by their control system, although their measurements are usually not employed for navigation. Absolute references such as that provided by navigation radio aids\footnote{The most widely used radio aids for aircraft navigation are VHF Omnidirectional Radio Range (VOR), Non Directional Beacon (NDB), and Instrument Landing System (ILS).} or GNSS are required to remove the position drift inherent to inertial navigation.

Low SWaP autonomous aircraft are too small to incorporate navigation aid receivers, which in any case are not available over vast regions of the Earth, exhibiting an extreme dependency on the availability of GNSS signals. A summary of the challenges of GNSS-Denied navigation and the research efforts intended to improve its performance is provided by \cite{Tippitt2020}. 

One approach to mitigate this problem is the establishment of alternative navigation signals, such as the ground-wave radio frequency location signals of the Theater Positioning System (TPS) for military applications \cite{Smith2005, Ma2011, Ma2013}, or the use of pseudo satellites generating GNSS-like signals from fixed ground locations \cite{Wang2002, Driscoll2011, Kuusniemi2012, Kim2014}, also known as pseudolites.

An alternative approach is to triangulate the aircraft position using existing signals originally intended for other purposes, such as those of television and cellular networks \cite{Raquet2007, Coluccia2014, Zheng2011}. Receivers for these signals, known as Signals of Opportunity (SoOP), have a sufficiently low SWaP to be mounted on most autonomous aircraft, although the quality and quantity of the available signals varies enormously depending on the location where the flight takes place.

\emph{Georegistration} matches landmarks or terrain features as scanned or imaged by vehicles to preloaded data \cite{Tippitt2020}, and can work based on Synthetic Aperture Radar (SAR) \cite{White2020, Sjanic2015, Nitti2015}, Light Detection and Ranging (LIDAR) \cite{Hemann2016, Miller2010, Csaba2018}, or camera visual systems, in what is known as \emph{image registration} \cite{Pritt2013, Ziaei2019, Conte2007, Wang2016}. While SAR suffers from being memory and computationally expensive and LIDAR is restricted for aviation purposes by its limited range, image registration is a potentially valid solution for completely removing the position drift in GNSS-Denied scenarios. Its main challenge is obtaining a high percentage of positive matches when supplied with onboard generated images that display the terrain not only at different altitudes and attitudes than those of the database images, but which are also obtained in different seasons, illumination, and weather conditions.

Another approach is to assist the inertial navigation system with the images taken by an onboard camera, but without the use of any prerecorded image database. This solution does not aim to completely eliminate the position drift as it lacks absolute references, but instead tries to reduce it to acceptable levels. \emph{Visual Odometry} (VO) has been employed for navigation of ground robots, road vehicles, and multirotors flying both indoors and outdoors. It relies exclusively on the digital images generated by one or more onboard cameras, incrementally estimating the vehicle pose based on the changes that its motion induces on the images \cite{Scaramuzza2011, Fraundorfer2012, Scaramuzza2012}. The incremental concatenation of relative poses results in a slow but unbounded pose drift, which can only be eliminated if aided by SLAM \cite{Scaramuzza2017, Cadena2016}, a particular case of VO in which the map of the already viewed terrain is stored and employed for loop closure in case it is revisited by the vehicle during its motion. Modern stand-alone VO algorithms such as SVO (semi direct visual odometry) \cite{Forster2014, Forster2016} and DSO (direct sparse odometry) \cite{Engel2016} are robust and exhibit a limited drift, while LSD-SLAM (large scale semi dense SLAM) \cite{Engel2014} and ORB-SLAM (large scale feature based SLAM)\footnote{ORB stands for Oriented FAST and rotated BRIEF, a type of blob feature.} \cite{Mur2015, Mur2017, Mur2017bis} may be more appropriate if the vehicle revisits an already imaged area. Note however that these methods were developed for road vehicles and multirotors and have not been tested for high speed fixed wing aircraft motion.

\emph{Visual Inertial Odometry} (VIO) is a navigation technique that combines inertial navigation with VO to reduce the position drift \cite{Scaramuzza2019, Yang2019}. Current VIO implementations are also primarily intended for ground robots, multirotors, and road vehicles, and hence rely exclusively on the readings of the vehicle inertial sensors and the images taken by the onboard camera.

The algorithms proposed in this article aim to take advantage of sensors already present onboard fixed wing aircraft and the special characteristics of the flight of these platforms to not only improve their GNSS-Denied navigation capabilities without the use of images, but also to facilitate the integration between the inertial and visual algorithms to obtain a VIO solution better suited to low SWaP fixed wing aircraft:
\begin{itemize}
\item Onboard sensors. It is possible to employ the magnetic field readings provided by a triad of magnetometers (useful as an absolute reference of body yaw to complement that of pitch and roll obtained from gravity and the accelerometers), the pressure altitude measurements obtained with the barometer (employed to limit the error in vertical position), and the airspeed recordings supplied by the Pitot tube and air vanes (used to limit the error in ground velocity and hence reduce the horizontal position drift).

\item Motion. In contrast with ground robots and road vehicles, the ground velocity of an aircraft is the combination of its airspeed, a low frequency quasi stationary wind field, plus high frequency turbulence. While GNSS receivers provide ground speed readings, only airspeed can be independently measured by the Pitot tube and air vanes. The proposed filter takes advantage of the airspeed measurements and the fact that the wind field has a significant effect on the aircraft kinematics and resulting trajectory but very little on the dynamics, in contrast with the turbulence, which produces strong accelerations that have a major influence on the aircraft dynamics and its control system, but that evens itself out over time showing little effect on the trajectory.

\item GNSS-Denied environment. While the indoors motion of ground robots and multirotors can never rely on GNSS signals, GNSS-Denied scenarios for road vehicles and outdoors multirotors are likely short lived and caused by building or terrain blockages. All these platforms can make use of additional distance sensors to avoid collisions and smart algorithms to remain in the road, and can ultimately stop or land if the GNSS signals are not recovered. In the case of fixed wing aircraft, it can be assumed that GNSS signals are present at the beginning of the flight, and that if they disappear the reason is likely to be technical error or intentional action (jamming or spamming), and the vehicle needs to be capable of flying for long periods of time in GNSS-Denied conditions until reaching a recovery location where it can be landed by remote control. The consequences of failing to do so are much more severe than for all other platforms mentioned above.
\end{itemize}


\section{High Fidelity Stochastic Simulation, Sensors, and Scenarios}\label{sec:Simulation}

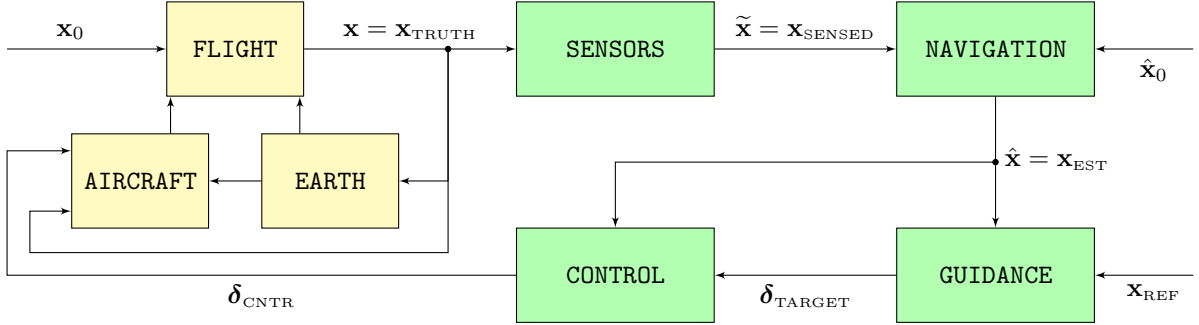
\begin{figure}[h]
\centering
\begin{tikzpicture}[auto, node distance=2cm,>=latex']
	\node [coordinate](x0input) {};
	\node [coordinate, below of=x0input, node distance=3.0cm] (deltaCNTRinput){};
	\node [blockyellow, right of=x0input, minimum width=1.8cm, node distance=3.0cm, align=center, minimum height=1.25cm] (FLIGHT) {\texttt{FLIGHT}};
	\node [coordinate, below of=FLIGHT, node distance=1.75cm] (midblocks){};
	\node [blockyellow, right of=midblocks, minimum width=1.8cm, node distance=1.25 cm, align=center, minimum height=1.25cm] (EARTH) {\texttt{EARTH}};
	\node [blockyellow, left of=midblocks, minimum width=1.8cm, node distance=1.25 cm, align=center, minimum height=1.25cm] (AIRCRAFT) {\texttt{AIRCRAFT}};
	\node [coordinate, right of=FLIGHT, node distance=2.8cm] (crosspoint1){};
	\node [coordinate, below of=crosspoint1, node distance=2.7cm] (crosspoint2){};
	\node [coordinate, left of=crosspoint2, node distance=5.5cm] (crosspoint3){};
	\node [blockgreen, right of=FLIGHT, minimum width=2.6cm, node distance=5.0cm, align=center, minimum height=1.25cm] (SENSORS) {\texttt{SENSORS}};
	\node [blockgreen, right of=SENSORS, minimum width=2.6cm, node distance=5.0cm, align=center, minimum height=1.25cm] (NAVIGATION) {\texttt{NAVIGATION}};
	\node [coordinate, right of=NAVIGATION, node distance=2.6cm] (x0estinput){};
	\node [coordinate, below of=NAVIGATION, node distance=1.5cm](tmp1) {};
	\node [blockgreen, below of=SENSORS, minimum width=2.6cm, node distance=3.0cm, align=center, minimum height=1.25cm] (CONTROL) {\texttt{CONTROL}};
	\node [blockgreen, below of=NAVIGATION, minimum width=2.6cm, node distance=3.0cm, align=center, minimum height=1.25cm] (GUIDANCE) {\texttt{GUIDANCE}};
	\node [coordinate, right of=GUIDANCE, node distance=2.6cm](xrefinput) {};

	\draw [->] ($(EARTH.north)-(0.40cm,0cm)$) -- ($(FLIGHT.south)+(0.85cm,0cm)$);
	\draw [->] (EARTH.west) -- (AIRCRAFT.east);
	\draw [->] ($(AIRCRAFT.north)+(0.40cm,0cm)$) -- ($(FLIGHT.south)-(0.85cm,0cm)$);
	\draw [->] (x0input) -- node[pos=0.4] {\nm{\xveczero}} (FLIGHT.west);
	\draw [->] (x0estinput) -- node[pos=0.4] {\nm{\xvecestzero}} (NAVIGATION.east);
	\draw [->] (FLIGHT.east) -- node[pos=0.5] {\nm{\xvec = \xTRUTH}} (SENSORS.west);
	\draw [->] (SENSORS.east) -- node[pos=0.5] {\nm{\xvectilde = \xSENSED}} (NAVIGATION.west);
	\filldraw [black] (tmp1) circle [radius=1pt];
	\draw [->] (NAVIGATION.south) -- node[pos=0.95] {\nm{\xvecest = \xEST}} (tmp1) -| (CONTROL.north);
	\draw [->] (tmp1) -- (GUIDANCE.north);
	\draw [->] (xrefinput) -- node[pos=0.4] {\nm{\xREF}} (GUIDANCE.east);
	\draw [->] (GUIDANCE.west) -- node[pos=0.5] {\nm{\deltaTARGET}} (CONTROL.east);
	\draw [->] (CONTROL.west) -| node[pos=0.25] {\nm{\deltaCNTR}} (deltaCNTRinput) |- ($(AIRCRAFT.west)+(0cm,0.4cm)$);
	\filldraw [black] (crosspoint1) circle [radius=1pt];
	\draw [->] (crosspoint1) |- (EARTH.east);
	\draw [->] (crosspoint1) -- (crosspoint2) -- (crosspoint3) |- ($(AIRCRAFT.west)-(0cm,0.4cm)$);
\end{tikzpicture}
\caption{Components of the high fidelity simulation}
\label{fig:flow_diagram}
\end{figure}

To evaluate the performance of the proposed inertial navigation algorithms, this article relies on Monte Carlo simulations consisting of one hundred runs of two different scenarios based on the high fidelity stochastic flight simulation graphically depicted in figure \ref{fig:flow_diagram}. The simulation models the flight in varying weather and turbulent conditions of a low SWaP fixed wing piston engine autonomous UAV, and has been specifically developed to analyze the performances of different navigation systems in GNSS-Denied conditions. Described in detail in \cite{SIMULATION}, the simulation models the influence on the resulting aircraft trajectory of many different factors, such as the guidance objectives that make up the mission, the atmospheric conditions, the wind field, the air turbulence, the local perturbations to the Earth gravity and magnetic fields, the aircraft aerodynamic and propulsive performances, its onboard sensors and their error sources, the control system that moves the throttle and the aerodynamic controls so the trajectory conforms to the guidance objectives, and the navigation system that processes the data obtained by the sensors and feeds the control system. The open source \nm{\CC} implementation of the high fidelity simulation and the inertial navigation algorithms proposed below is available in \cite{Gallo2020_simulation}.

The simulation consists on two distinct processes. The first, represented by the yellow blocks on the left of figure \ref{fig:flow_diagram}, focuses on the physics of flight and the interaction between the aircraft and its surroundings that results in the actual or real aircraft trajectory \nm{\xvec = \xTRUTH}; the second, represented by the green blocks on the right, contains the aircraft systems in charge of ensuring that the resulting trajectory adheres as much as possible to the mission objectives. It includes the different sensors whose output comprise the sensed trajectory \nm{\xvectilde = \xSENSED}, the navigation system in charge of filtering it to obtain the estimated trajectory \nm{\xvecest = \xEST}, the guidance system that converts the reference objectives \nm{\xREF} into the control targets \nm{\deltaTARGET}, and the control system that adjusts the position of the throttle and aerodynamic control surfaces \nm{\deltaCNTR} so the estimated trajectory \nm{\xvecest} is as close as possible to the reference objectives \nm{\xREF}. As shown in the figure, the two parts of the simulation are not independent. The \emph{total system error} (TSE) or difference between \nm{\xvec} and \nm{\xREF} is the combination of the \emph{navigation system error} (NSE) or difference between \nm{\xvec} and \nm{\xvecest} and the \emph{flight technical error} (FTE) or difference between \nm{\xvecest} and \nm{\xREF}. 

All components of the simulation have been modeled with as few simplifications as possible to increase the realism of the results, as explained in \cite{SIMULATION}. With the exception of the aircraft performances and its control system, which are deterministic, all other simulation components are treated as stochastic and hence vary from one execution to the next, enhancing the significance of the Monte Carlo simulation results.


\subsubsection*{Sensors}

The representation of the different errors introduced by the onboard sensors is of particular interest for inertial navigation. The following tables provide the performance values employed to simulate the different sensors, although the reader should refer to \cite{SENSORS} for a detailed description of the expressions that provide the measurement error introduced by the various sensors, as well as for an explanation of the process employed to obtain the values shown at the tables from the sensors data sheets.
\begin{center}
\begin{tabular}{l|clc|clc}
	\hline
	GYR-ACC Baseline & Variable & \multicolumn{1}{c}{Value} & Unit & Variable & \multicolumn{1}{c}{Value} & Unit \\
	\hline
	Bias Drift		& \nm{\sigmauGYR}	& \nm{1.42 \cdot 10^{-4}}	& [\nm{deg/sec^{1.5}}] & \nm{\sigmauACC}	& \nm{6.86 \cdot 10^{-5}}	& [\nm{m/sec^{2.5}}] \\ 
	System Noise	& \nm{\sigmavGYR}	& \nm{4.30 \cdot 10^{-3}}	& [\nm{deg/sec^{0.5}}] & \nm{\sigmavACC}	& \nm{4.83 \cdot 10^{-4}}	& [\nm{m/sec^{1.5}}] \\
	Scale Factor  	& \nm{\sGYR}		& \nm{1.50 \cdot 10^{-5}}	& [-] 				   & \nm{\sACC}			& \nm{5.00 \cdot 10^{-5}}	& [-] \\
	Cross Coupling	& \nm{\mGYR}		& \nm{4.35 \cdot 10^{-5}}	& [-]                  & \nm{\mACC}		    & \nm{3.05 \cdot 10^{-5}}	& [-] \\
	Bias Offset		& \nm{\BzeroGYR}	& \nm{2.00 \cdot 10^{-1}}	& [\nm{deg/sec}]       & \nm{\BzeroACC}	    & \nm{1.57 \cdot 10^{-1}}	& [\nm{m/sec^2}] \\
	\hline
\end{tabular}
\captionof{table}{Performance of ``Baseline'' gyroscopes and accelerometers} \label{tab:Sensors_gyr_acc}
\end{center}

The modeling of the gyroscope and accelerometer triads includes in-run error sources such as bias drift and system noise, run-to-run contributions such as bias offsets, as well as fixed scale factor and cross coupling errors, together with uncertainties in the exact position and attitude of the IMU with respect to the aircraft structure. The baseline performance values shown in table \ref{tab:Sensors_gyr_acc} correspond to the MEMS gyroscopes and accelerometers installed inside the Analog Devices ADIS16488A IMU \cite{ADIS16488A}.

The magnetometers triad is modeled similarly, but without bias drift. The baseline magnetometer features employed in the simulation are shown in table \ref{tab:Sensors_mag}, where the system noise has been taken from \cite{Groves2008} and the rest of the parameters correspond to the magnetometers present inside the Analog Devices ADIS16488A IMU \cite{ADIS16488A}.
\begin{center}
\begin{tabular}{l|clc}
	\hline
	MAG Baseline & Variable & \multicolumn{1}{c}{Value} & Unit \\
	\hline
	System Noise		& \nm{\sigmavMAG}	& \nm{5.00 \cdot 10^{0}}	& [\nm{nT \cdot sec^{0.5}}] \\
	Scale Factor \& Soft Iron		& \nm{\sMAG}		& \nm{7.50 \cdot 10^{-4}}	& [-] \\
	Cross Coupling \& Soft Iron		& \nm{\mMAG}		& \nm{9.16 \cdot 10^{-4}}	& [-] \\
	Hard Iron			& \nm{\BhiMAG}		& \nm{1.75 \cdot 10^{2}}    & [\nm{nT}] \\
	Bias Offset			& \nm{\BzeroMAG}	& \nm{5.00 \cdot 10^{2}}    & [\nm{nT}] \\
	\hline
\end{tabular}
\captionof{table}{Performance of ``Baseline'' magnetometers} \label{tab:Sensors_mag}
\end{center}

The airspeed sensors (TAS for true airspeed, AOA for angle of attack, AOS for angle of sideslip) and air data sensors (OSP for outside static pressure, OAT for outside air temperature) only consider system noise and bias offset in their measurements, and are usually provided by a Pitot tube, air vanes, barometer, and thermometer. The baseline parameters shown in table \ref{tab:Sensors_vtasb_air} correspond to the Aeroprobe air data system \cite{Aeroprobe}, which is equipped with a multi hole Pitot tube.
\begin{center}
\begin{tabular}{l|lrc}
	\hline
	Baseline					& \multicolumn{1}{c}{Variable} & \multicolumn{1}{c}{Value} & Unit \\
	\hline
	TAS System Noise		    & \nm{\sigmaTAS}	& \nm{3.33 \cdot 10^{-1}}	& [\nm{m/sec}] \\
	TAS Bias Offset        	    & \nm{\BzeroTAS}	& \nm{3.33 \cdot 10^{-1}}	& [\nm{m/sec}] \\
	AOA System Noise		    & \nm{\sigmaAOA}	& \nm{3.33 \cdot 10^{-1}}	& [\nm{deg}] \\
	AOA Bias Offset        	    & \nm{\BzeroAOA}	& \nm{3.33 \cdot 10^{-1}}	& [\nm{deg}] \\
	AOS System Noise		    & \nm{\sigmaAOS}	& \nm{3.33 \cdot 10^{-1}}	& [\nm{deg}] \\
	AOS Bias Offset			    & \nm{\BzeroAOS}	& \nm{3.33 \cdot 10^{-1}}	& [\nm{deg}] \\
	OSP System Noise		    & \nm{\sigmaOSP}	& \nm{1.00 \cdot 10^{+2}}	& [\nm{pa}] \\
	OSP Bias Offset        	    & \nm{\BzeroOSP}	& \nm{1.00 \cdot 10^{+2}}	& [\nm{pa}] \\
	OAT System Noise			& \nm{\sigmaOAT}	& \nm{5.00 \cdot 10^{-2}}   & [\nm{^{\circ}K}] \\
	OAT Bias Offset       	    & \nm{\BzeroOAT}	& \nm{5.00 \cdot 10^{-2}}	& [\nm{^{\circ}K}] \\
	\hline
\end{tabular}
\captionof{table}{Performance of ``Baseline'' airspeed and air data sensors} \label{tab:Sensors_vtasb_air}
\end{center}


\subsubsection*{Scenarios}

Two different scenarios are employed to analyze the consequences of losing the GNSS signals. Although a short summary is included below, detailed descriptions of the mission, weather, and wind field employed in each scenario can be found in \cite{SIMULATION}. Most parameters comprising the scenario are defined stochastically, resulting in different values for every execution. Note that all results shown in sections \ref{sec:Results}, \ref{sec:Comparison}, and \ref{sec:Influence} are based on Monte Carlo simulations comprising one hundred runs of each scenario, testing the sensitivity of the proposed algorithms to a wide variety of values in the parameters.
\begin{itemize}
\item Scenario \#1 has been defined with the objective of adequately representing the challenges faced by an autonomous fixed wing UAV that suddenly cannot rely on GNSS and hence changes course to reach a predefined recovery location situated at approximately one hour of flight time. In the process, in addition to executing an altitude and airspeed adjustment, the autonomous aircraft faces significant weather and wind field changes that make its GNSS-Denied navigation even more challenging. 

With respect to the mission, the stochastic parameters include the initial airspeed, pressure altitude, and bearing (\nm{\vtasINI, \HpINI, \chiINI}), their final values (\nm{\vtasEND, \HpEND, \chiEND}), and the time at which each of the three maneuvers is initiated\footnote{Turns are executed with a bank angle of \nm{\xiTURN = \pm 10 \lrsb{deg}}, altitude changes employ an aerodynamic path angle of \nm{\gammaTASCLIMB = \pm 2 \lrsb{deg}}, and airspeed modifications are automatically executed by the control system as setpoint changes.}. The scenario lasts for \nm{\tEND = 3800 \lrsb{sec}}, while the GNSS signals are lost at \nm{\tGNSS = 100 \lrsb{sec}}, which does not involve any loss of generality as the accuracy of the aircraft pose (attitude and position) estimation does not degrade with time when GNSS signals are available.

The wind field is also defined stochastically, as its two parameters (speed and bearing) are constant both at the beginning (\nm{\vwindINI, \chiWINDINI}) and conclusion (\nm{\vwindEND, \chiWINDEND}) of the scenario, with a linear transition in between. The specific times at which the wind change starts and concludes also vary stochastically among the different simulation runs. As described in \cite{SIMULATION}, the turbulence remains strong throughout the whole scenario, but its specific values also vary stochastically from one execution to the next.

A similar linear transition occurs with the temperature and pressure offsets that define the atmospheric properties \cite{INSA}, as they are constant both at the start (\nm{\DeltaTINI, \DeltapINI}) and end (\nm{\DeltaTEND, \DeltapEND}) of the flight. In contrast with the wind field, the specific times at which the two transitions start and conclude are not only stochastic but also different from each other.

Although quite generic, there are two reasons why scenario \#1 can not constitute the only means employed to evaluate the behavior of different GNSS-Denied navigation algorithms. On one side, its mission includes three maneuvers (a change of bearing, a change of airspeed, and a change of pressure altitude), but for the most part it consists of a long straight level flight in which the aircraft systems have lots of time to recover from estimation errors induced by the maneuvers, which are executed quite apart from each other. Additionally, the GNSS-Denied position errors incurred by the proposed inertial navigation system may exhibit a significant dependency on the weather and wind changes that occur since the aircraft can no longer employ the GNSS signals for navigation. The prevalence of these errors coupled with the significant scenario \#1 weather and wind changes prevents the detection of smaller error sources when analyzing the Monte Carlo simulation results in section \ref{sec:Results}. These two reasons indicate the need for a scenario \#2.

\item Scenario \#2 represents the challenges involved in continuing with the original mission upon the loss of the GNSS signals, executing a series of continuous turn maneuvers over a relatively short period of time with no atmospheric or wind variations. As in scenario \nm{\#1}, the GNSS signals are lost at \nm{\tGNSS = 100 \lrsb{sec}}, but the scenario duration is shorter (\nm{\tEND = 500 \lrsb{sec}}). The initial airspeed and pressure altitude (\nm{\vtasINI, \HpINI}) are defined stochastically and do not change throughout the whole scenario; the bearing however changes a total of eight times between its initial and final values, with all intermediate bearing values as well as the time for each turn varying stochastically from one execution to the next. Although the same turbulence is employed as in scenario \nm{\#1}, the wind and atmospheric parameters (\nm{\vwindINI, \chiWINDINI, \DeltaTINI, \DeltapINI}) remain constant throughout scenario \nm{\#2}.
\end{itemize}


\section{Proposed Navigation Filter}\label{sec:Filter}

The navigation system takes the sensor outputs or sensed trajectory \nm{\xvectilde = \xSENSED} and processes them to obtain the estimated trajectory \nm{\xvecest = \xEST} , which then passes to the guidance and control systems. Note that in the simulation all sensors as well as the navigation system operate at \nm{100\lrsb{hz}}, with the exception of the GNSS receiver (until the GNSS signals are lost), which works at \nm{1\lrsb{hz}}. With the objective of minimizing the NSE (difference between \nm{\xvec} and \nm{\xvecest}) in GNSS-Denied conditions, the proposed architecture divides the navigation filter into three smaller complimentary filters named the air data, attitude, and position filters, respectively.


\subsubsection*{Air Data Filter}

As the ground speed is the sum of the airspeed and the wind speed, and only the former can be observed in GNSS-Denied conditions, the variables describing the aircraft motion with respect to the air (airspeed, aerodynamic angles, atmospheric conditions) are isolated from the estimation point of view from those that link the aircraft to the ground (attitude, ground speed, position), and it is hence more computationally efficient to implement a smaller independent filter for these variables.
\begin{eqnarray}
\nm{\xvec_{\sss AIR}\lrp{t}} & = & \nm{\lrsb{\vtas, \ \alpha, \ \beta, \ T, \ \Hp}^T}\label{eq:filter_air_st} \\
\nm{\yvec_{{\sss AIR},n}} & = & \nm{\lrsb{\vtastilde, \ \alphatilde, \ \betatilde, \ \Ttilde, \ \ptilde}^T}\label{eq:filter_air_y}
\end{eqnarray}

The air data filter state vector \nm{\xvec_{\sss AIR}} is composed by the airspeed \nm{\vtas}, angle of attack \nm{\alpha}, angle of sideslip \nm{\beta}, atmospheric temperature \nm{T}, and pressure altitude \nm{\Hp}, while the observation vector \nm{\yvec_{{\sss AIR},n}} includes the measurements provided by the Pitot tube \nm{\vtastilde}, air vanes (\nm{\alphatilde} and \nm{\betatilde}), and atmospheric sensors (\nm{\Ttilde} and \nm{\ptilde}). Although implemented as an extended Kalman filter (EKF) \cite{Simon2006}, the absence of relationships among the state variables implies that for all practical purposes it behaves as a low pass filter. The state and observation equations are the following:
\begin{eqnarray}
\nm{{\dot {\vec x}}_{\sss AIR}} & = & \nm{\vec 0}\label{eq:filter_air_state} \\
\nm{\vtastilde}  & = & \nm{\vtas}\label{eq:filter_air_obs_vtas} \\
\nm{\alphatilde} & = & \nm{\alpha}\label{eq:filter_air_obs_alpha} \\
\nm{\betatilde}  & = & \nm{\beta}\label{eq:filter_air_obs_beta} \\
\nm{\Ttilde}     & = & \nm{T}\label{eq:filte_air_obs_T} \\
\nm{\ptilde}     & = & \nm{\pzero \ \lrp{1 + \frac{\betaT}{\Tzero} \ \Hp}^{\sss \gBR}}\label{eq:filter_air_obs_Hp}
\end{eqnarray}

Expression (\ref{eq:filter_air_obs_Hp}), together with the standard mean sea level temperature and pressure values (\nm{\Tzero} and \nm{\pzero}), the temperature gradient \nm{\betaT}, air specific constant \nm{R}, and standard acceleration of free fall \nm{\gzero}, are all taken from the International Civil Aviation Organization (ICAO) Standard Atmosphere model or ISA \cite{ISA}.

The navigation filter tracks the weather by continuously estimating the temperature and pressure offsets (\nm{\DeltaT} and \nm{\Deltap}), which indicate the deviation of the atmosphere through which the aircraft flies from the static model provided by ISA. The variation with time and horizontal position of these two offsets is the basis of the non standard ISA model or INSA \cite{INSA}, which establishes accurate relationships between the atmospheric properties (temperature, pressure, density) and altitude that simultaneously comply with local observations while verifying all ISA hypotheses. Note that the INSA model converges to ISA when both offsets are zero.

Estimating the pressure offset \nm{\Deltapest} is the task of the position filter, but the temperature offset \nm{\DeltaTest} is estimated as follows after each execution of the air data filter. Refer to \cite{INSA} for more details on (\ref{eq:filter_air_DeltaT}):
\neweq{\DeltaTest = \Test - \Tzero - \betaT \ \Hpest}{eq:filter_air_DeltaT}


\subsubsection*{Attitude Filter}

The main objective of the attitude filter is the estimation of the aircraft attitude \nm{\qNBest} or rotation between the North-East-Down (NED) and body frames represented by its unit quaternion \cite{Sola2017}, although it also estimates the angular velocity \nm{\wNBBest} (from NED to body viewed in body), the full gyroscope error \nm{\EGYRest}\footnote{\nm{\EGYR} includes all gyroscopes error sources except system noise as explained in \cite{SENSORS}.}, the full magnetometer error \nm{\EMAGest}\footnote{\nm{\EMAG} includes all magnetometers error sources except system noise as explained in \cite{SENSORS}.}, and the difference \nm{\BNDEVest} between the Earth magnetic field provided by the onboard model \nm{\BNMODEL} and the real one \nm{\BNREAL}. The observations are provided by the gyroscopes, which measure the angular velocity \nm{\wIBBtilde} from the inertial frame\footnote{In the simulation, the inertial frame is centered at the Earth center of mass and its axes do not rotate with respect to any stars other than the Sun.} to body frame viewed in body, the magnetometers that measure the magnetic field \nm{\BBtilde} in the body frame, and the accelerometers that provide the specific force \nm{\fIBBtilde} or non gravitational acceleration from the inertial to the body frame viewed in body.
\begin{eqnarray}
\nm{\xvec_{\sss ATT}\lrp{t}} & = & \nm{\lrsb{\qNB, \ \wNBB, \ \EGYR, \ \EMAG, \ \BNDEV}^T}\label{eq:filter_att_st} \\
\nm{\yvec_{{\sss ATT},n}} & = & \nm{\lrsb{\wIBBtilde, \ \BBtilde, \ \fIBBtilde}^T}\label{eq:filter_att_y}
\end{eqnarray}

While the only non zero state equation is that of the unit quaternion (\ref{eq:filter_att_state_qNB}) (refer to \cite{Sola2017} for the relationship between the time derivative of the unit quaternion and the angular velocity), the main novelty of this article lies in the selection of the observation equations, which are intended to minimize the negative influence on the estimation of the aircraft attitude of the inaccurate estimations of ground velocity and position in the position filter caused by the lack of GNSS signals. This design should not be employed when GNSS signals are available as it results in inferior performance when the GNSS observations are added to the position filter, but it isolates and protects the attitude filter estimations in GNSS-Denied conditions.
\begin{eqnarray}
\nm{\qNBdot} & = & \nm{\dfrac{1}{2} \; \qNB \otimes \vec \wNBB} \label{eq:filter_att_state_qNB} \\
\nm{{\dot {\vec x}}_{\sss ATT,OTHER}} & = & \nm{\vec 0} \label{eq:filter_att_state_other} \\
\nm{\pvec_n}    & = & \nm{\lrsb{\gcNMODEL, \ \BNMODEL, \ \vN, \vWINDN, \ \wIEN, \ \wENN, \ \acorN, \ \EACC}^T} \label{eq:filter_att_obs_pvec} \\
\nm{\wIBBtilde} & = & \nm{\wNBB + \qNBast \otimes \lrp{\wIEN + \wENN} \otimes \qNB+ \EGYR}\label{eq:filter_att_obs_wIBB} \\
\nm{\BBtilde}   & = & \nm{\qNBast \otimes \lrp{\BNMODEL - \BNDEV} \otimes \qNB + \EMAG} \label{eq:filter_att_obs_BB} \\
\nm{\fIBBtilde} & = & \nm{\wNBBskew \, \lrsb{\qNBast \otimes \lrp{\vN - \vWINDN} \otimes \qNB} + \qNBast \otimes \lrp{\wENNskew \, \vN + \acorN - \gcNMODEL} \otimes \qNB + \EACC} \label{eq:filter_att_obs_fIBB}
\end{eqnarray}

The negative influence of the inaccurate velocity and position estimations is restricted to the \nm{\pvec_n} vector, which contains the gravity \nm{\gcNMODEL} and magnetism \nm{\BNMODEL} employed in the filter (which differ slightly from the real ones experienced by the aircraft), the aircraft ground and wind velocities (\nm{\vN, \vWINDN}) viewed in NED, the Earth and motion angular velocities \nm{\wIEN} and \nm{\wENN}\footnote{The Earth angular velocity \nm{\wIEN} is that caused by the rotation of the Earth, represented by the Earth Centered Earth Fixed (ECEF) frame, with respect to the inertial frame, and is viewed in NED; the motion angular velocity \nm{\wENN}, also viewed in NED, represents the angular velocity caused by the motion of the aircraft above the curved Earth surface.}, the Coriolis acceleration \nm{\acorN}, and the full accelerometer error \nm{\EACC}\footnote{\nm{\EACC} includes all accelerometers error sources except system noise as explained in \cite{SENSORS}.}. The \nm{\pvec_n} vector is taken from the previous step execution of the position filter (every \nm{\pvec_n} member can be obtained from the position filter outputs including the geodetic coordinates \nm{\xEgdt}, as shown in (\ref{eq:filter_att_obs_pvec_dependencies})), and the measures taken to minimize its errors and how they spread by means of the attitude filter observation equations are the key to improving the estimation of the aircraft attitude \nm{\qNBest}.
\neweq{\pvec_n = \vec f\lrp{\xEgdt, \ \vN, \ \vWINDN, \ \EACC}} {eq:filter_att_obs_pvec_dependencies}

Note that (\ref{eq:filter_att_obs_wIBB}), which combines the gyroscope error with the three angular velocities that together make up the inertial rotation speed \nm{\wIBB}, and (\ref{eq:filter_att_obs_BB}), which adds the magnetometers errors to the true magnetic field, do not include any simplifications and are hence quite precise, although they are subject to the errors introduced by the use of \nm{\pvec_n} when evaluating some of its members.
 
Expression (\ref{eq:filter_att_obs_fIBB}) is the observation equation with the highest influence on the estimation results as well as the least precise. It is based on the fact that the specific force measured by the accelerometers is equal to the non gravitational (aerodynamic and propulsive) accelerations, but modified as described below to minimize the negative influence of the errors present in the \nm{\pvec_n} vector. The obtainment of (\ref{eq:filter_att_obs_fIBB}) starts with the true expressions of the specific force viewed in both the body and NED frames (\ref{eq:specific_force_definition}, \ref{eq:specific_force_definition_ned}), which are based on the (\ref{eq:EquationsMotion_equations_force8}, \ref{eq:EquationsMotion_equations_force8_N}) expressions for the time derivative of the ground velocity obtained in appendix \ref{sec:DerivativeVelocity}:
\begin{eqnarray}
\nm{\fIBB} & = & \nm{\wEBBskew \; \vB + \vBdot + \qNBast \otimes \lrp{\acorN - \gcN} \otimes \qNB} \nonumber \\
           & = & \nm{\wEBBskew \; \lrp{\vTASB + \vTURBB + \vWINDB} + \lrp{\vTASBdot + \vTURBBdot + \vWINDBdot} + \qNBast \otimes \lrp{\acorN - \gcN} \otimes \qNB}\label{eq:specific_force_definition} \\
\nm{\fIBN} & = & \nm{\wENNskew \; \vN + \vNdot + \acorN - \gcN} \nonumber \\
           & = & \nm{\wENNskew \; \lrp{\vTASN + \vTURBN + \vWINDN} + \lrp{\vTASNdot + \vTURBNdot + \vWINDNdot} + \acorN - \gcN}\label{eq:specific_force_definition_ned}
\end{eqnarray}

The process leading to (\ref{eq:filter_att_obs_fIBB}) from (\ref{eq:specific_force_definition}) is the following: 
\begin{itemize}

\item Discard the airspeed and turbulence time derivatives viewed in body (\nm{\vTASBdot, \, \vTURBBdot}) as they are unknown to the filter. The omission of the turbulence acceleration is of particular importance, as it is usually significantly bigger than the errors introduced by the accelerometer \nm{\EACC} and hence one of the reasons why \nm{\EACCest} can not be estimated in the attitude filter without GNSS observations. This results in (\ref{eq:specific_force_definition_middle}), which only represents a middle step in the obtainment of (\ref{eq:filter_att_obs_fIBB}) and is shown for added clarity:
\begin{eqnarray}
\nm{\fIBB} & \nm{\sim} & \nm{\wEBBskew \; \lrp{\vTASB + \vTURBB + \vWINDB} + \vWINDBdot + \qNBast \otimes \lrp{\acorN - \gcN} \otimes \qNB}\nonumber \\
           & = & \nm{\wNBBskew \; \lrp{\vTASB + \vTURBB + \vWINDB} + \vWINDBdot}\label{eq:specific_force_definition_middle} \\
           &   & \nm{+ \ \qNBast \otimes \lrsb{\wENNskew \; \lrp{\vTASN + \vTURBN + \vWINDN} + \acorN - \gcN} \otimes \qNB}\nonumber
\end{eqnarray}

Discarding the instantaneous accelerations of the airspeed and the turbulence viewed in body has negative consequences for the filter accuracy. However, both \nm{\vTASBdot} and \nm{\vTURBBdot} roughly oscillate around zero (airspeed tends to be constant when viewed in body while turbulence oscillates around zero also in body) and hence are easily absorbed into the EKF measurement noise respecting its zero mean white noise character \cite{Simon2006}. This implies that although less accurate, the EKF continues to provide an unbiased estimation of the state vector.

\item Discard the wind field time derivatives viewed in NED (\nm{\vWINDNdot}) as the wind is approximately quasi stationary and hence its time derivatives are small. Although wind variations may have a significant influence in the resulting aircraft trajectory, they are too slow and short lived to influence its dynamics. The wind acceleration \nm{\vWINDNdot} can hence be assumed to oscillate around zero and be absorbed into the measurement noise, resulting in the filter being less accurate but behaving properly and providing unbiased estimations of the state variables \cite{Simon2006}. Expression (\ref{eq:specific_force_definition_middle}) thus gets transformed into:
\neweq{\fIBB \sim \wNBBskew \; \lrp{\vTASB + \vTURBB} + \qNBast \otimes \lrsb{\wENNskew \; \lrp{\vTASN + \vTURBN + \vWINDN} + \acorN - \gcN} \otimes \qNB} {eq:specific_force_definition_middle_2}

The obtainment of (\ref{eq:filter_att_obs_fIBB}) from (\ref{eq:specific_force_definition_middle_2}) is straightforward. Note that the product of the motion angular velocity \nm{\wEN} by the wind speed \nm{\vWIND} has not been discarded, while that of the aircraft angular velocity \nm{\wNB} has. This may look counter-intuitive, but is correct and better understood by looking at (\ref{eq:specific_force_definition_ned}).

If GNSS signals are available, it is not necessary to eliminate the speed derivatives with time from the specific force observation equation. However, in GNSS-Denied conditions, the estimation of \nm{\vNdot} or \nm{\vBdot} in the position filter is not accurate enough, and hence its usage in (\ref{eq:filter_att_obs_fIBB}) would introduce artificial biases resulting in a deterioration of the attitude estimation.

The distinction between eliminating the airspeed and turbulence time derivatives in body and the wind field time derivative in NED is key for the minimization of the attitude filter estimation errors; as shown in the results below, other options such as discarding the whole ground speed time derivative \nm{\vBdot} when operating in GNSS-Denied scenarios result in the introduction of artificial biases during maneuvers and the growth of the attitude estimation error.

\item Include the difference between the real gravity field \nm{\gcNREAL} measured by the accelerometers and \nm{\gcNMODEL}, which is computed onboard based on the gravity model. These differences are too small to be observed during flight and have a minor but not negligible influence on the filter behavior.

\item Although the terms \nm{\vN}, \nm{\vWINDN}, \nm{\wENN}, \nm{\acorN}, and \nm{\EACC} are correct as written, they are not state variables but taken or computed from the results of the position filter, and hence may include potentially significant errors in GNSS-Denied conditions.

\end{itemize}


\subsubsection*{Position Filter}

Without the absolute position and velocity measurements provided by the GNSS receiver, the position filter does not have enough observability to avoid errors in its estimations, but nevertheless has two main objectives. On one side, the estimation of the ground position (horizontal and vertical) needs to result in errors that are as small as possible; on the other, it needs to supply the attitude filter with contained \nm{\pvec_n} estimations (\ref{eq:filter_att_obs_pvec}, \ref{eq:filter_att_obs_pvec_dependencies}) so they do not destabilize the attitude filter.
\begin{eqnarray}
\nm{\xvec_{\sss POS}\lrp{t}} & = & \nm{\lrsb{\ \fIBB, \ \EACC}^T}\label{eq:filter_pos_st} \\
\nm{\yvec_{{\sss POS},n}}  & = & \nm{\fIBBtilde}\label{eq:filter_pos_y} \\
\nm{{\dot {\vec x}}_{\sss POS}} & = & \nm{\vec 0} \label{eq:filter_pos_state} \\
\nm{\fIBBtilde}  & = & \nm{\fIBB + \EACC} \label{eq:filter_pos_obs_fIBB}
\end{eqnarray}

The EKF filter itself is extremely simple and in fact can not separate between \nm{\fIBB} and \nm{\EACC}, which are hence employed together in (\ref{eq:filter_att_obs_fIBB}). The interesting part lies in those variables that are estimated after each filter step. In the case of the pressure offset \nm{\Deltap} \cite{INSA} and the wind velocity \nm{\vWINDN}, there are not enough observations in GNSS-Denied conditions to make any proper estimation, so they are kept constant from their last GNSS aided estimations. Note that the presence of absolute position and velocity observations provided by the GNSS enables the estimation of \nm{\Deltapest_{GNSS-BASED}} and \nm{\hat{\vec v}_{{\sss WIND},GNSS-BASED}^{\sss N}} by reversing the process described next for GNSS-Denied conditions.
\begin{eqnarray}
\nm{\Deltapest} & = & \nm{\Deltapest_{GNSS-BASED,LAST}}\label{eq:filter_pos_Deltap} \\
\nm{\vWINDNest} & = & \nm{\hat{\vec v}_{{\sss WIND},GNSS-BASED,LAST}^{\sss N}} \label{eq:filter_pos_vWINDN}
\end{eqnarray}

With the wind estimation kept constant, the ground velocity viewed in NED can be estimated by making use of the airspeed and angles of attack and sideslip estimated by the air data filter:
\begin{eqnarray}
\nm{\vTASNest} & = & \nm{\qNBest \otimes \vTASB\lrp{\vtasest, \, \alphaest, \, \betaest} \otimes \qNBastest}\label{eq:filter_pos_vTASN} \\
\nm{\vNest} & = & \nm{\vWINDNest + \vTASNest}\label{eq:filter_pos_vN}
\end{eqnarray}

The vertical position \nm{h} is estimated better from the pressure altitude \nm{\Hpest} than by integrating the ground velocity \nm{\vN}. It is a two step process in which the geopotential altitude H is first estimated based on estimations of the pressure altitude \nm{\Hp}, temperature offset \nm{\DeltaT}, and pressure offset \nm{\Deltap} \cite{INSA}, followed by its conversion to geometric altitude according to the spherical Earth model \cite{ISA}, where the Earth radius \nm{\RE} is taken from \cite{SMITHSONIAN}.
\begin{eqnarray}
\nm{\hat H} & = & \nm{f\lrp{\Hpest, \, \DeltaTest, \, \Deltapest}}\label{eq:GNC_Navigation_AirDataFilterGNSSDenied_H} \\
\nm{\hat h} & = & \nm{\dfrac{\RE \cdot \hat H}{\RE - \hat H}}\label{eq:GNC_Navigation_AirDataFilterGNSSDenied_h}
\end{eqnarray}

Longitude and latitude are finally obtained by integrating the ground velocity estimated above.
 

\section{Navigation System Error in GNSS-Denied Conditions}\label{sec:Results}

This section presents the results obtained with the proposed navigation filter when executing Monte Carlo simulations of the two GNSS-Denied scenarios\footnote{All trajectories are executed at the location indicated by the \say{mix} (MX) terrain type, as described in \cite{SIMULATION}.}, each consisting of one hundred executions. \cite{SIMULATION} provides detailed descriptions of the different modules of the high fidelity simulation together with the definition of the two scenarios, which are summarized in section \ref{sec:Simulation}. \cite{SENSORS} explains the expressions employed to model the measurement errors introduced by the different sensors, as well as their \say{baseline} values employed in the simulation. A short summary can also be found in section \ref{sec:Simulation}. 

The navigation system error (NSE) incurred by the proposed navigation system when estimating the value of the different variables in GNSS-Denied conditions is evaluated in the tables below according to the aggregated metrics and the aggregated final state metrics, both defined in \cite{SIMULATION}. The \emph{aggregated metrics} include the mean, standard deviation, and maximum value taking as inputs the NSE of every navigation filter step and simulation run\footnote{As the navigation system works at \nm{100 \lrsb{hz}} and there are \nm{100} runs, this amounts to \nm{100 \times \lrp{100 \times 3800 + 1}} points for scenario \nm{\#{1}} and \nm{100 \times \lrp{100 \times 500 + 1}} points for scenario \nm{\#2}.}, while the \emph{aggregated final state metrics} do the same but only for the last filter evaluation in each run\footnote{This amounts to \nm{100} points for each scenario.}. When depicting the NSE, the figures below show the variation with time of the mean and standard deviation of the one hundred different executions of each scenario.

As as example on how to read the tables below, the three right hand columns of table \ref{tab:attitude_errors} show the aggregated body attitude NSE for the two scenarios. In the case of scenario \nm{\#1}, the first step involves computing the means, standard deviations, and maximum values (trajectory metrics) of each of the one hundred executions independently, each taking as input the \nm{100 \times 3800 + 1} evaluations of the NSE after each navigation filter step. The aggregated metrics are then obtained by again computing the means, standard deviations, and maximum values, but this time taking as input only the one hundred values of the corresponding trajectory metric. The standard deviation of the one hundred trajectory means results in \nm{0.072 \lrsb{deg}}, while the mean of the one hundred trajectory standard deviations is \nm{0.078 \lrsb{deg}}.


\subsubsection*{Air Data Estimation}

All variables estimated by the air data filter, this is, those contained in (\ref{eq:filter_air_st}) plus \nm{\DeltaT}, share the same qualitative NSE results, which are independent of the scenario. Measured by the Pitot tube, air vanes, and atmospheric sensors, their measurements contain system noise plus a run-to-run constant bias offset\footnote{The temperature offset \nm{\DeltaT} is not directly measured, but per (\ref{eq:filter_air_DeltaT}) is a linear combination of the temperature and pressure altitude, and hence combines their measurement errors.}.

The individual trajectory metrics of their estimation errors are biased (not zero mean) and the aggregated ones are zero mean or unbiased, meaning that the bias origin is random and present in the data itself, not in the navigation algorithms. No drift is present in either the trajectory or the aggregated metrics. The repeatability of the results is elevated, as the standard deviation and maximum absolute value vary little among different runs. The air data filter hence behaves like a low pass filter, removing the system noise but unable to separate the bias offsets introduced by the sensors, which are incorporated in their totality into the estimated values.

The main consequence of the biases lack of observability is that the specific bias offset realizations with which the sensors operate in a given run become thresholds that can not be overcome by the navigation system. Note that the bias offsets are based on the sensors exclusively, so the estimation errors can be considered as bounded or limited by the quality of the Pitot tube, air vanes, and atmospheric sensors.


\subsubsection*{Body Attitude Estimation}

Minimizing the aircraft body attitude estimation error \nm{\DeltarBest = \qNBest \ominus \qNB}, where \nm{\ominus} represents the rotations minus operator \cite{Sola2017}, is one of the primary objectives of the inertial navigation system, and a necessary condition for maintaining flight stability as well as for the fusion between the inertial and visual navigation algorithms. Experiencing no drift in attitude is indispensable for flight viability, as if this is not the case, in particular for pitch and roll, at some point the aircraft will breach its flight envelope and the control system will lose its hold on the platform.

The behavior of the attitude filter is intrinsically different to that of the air data filter. In addition to the fact that there is no sensor directly measuring the body attitude, the errors introduced by the attitude filter sensors include contributions such as bias drift, scale factors, and cross coupling errors, that make them significantly more challenging to estimate. On the positive side, there is some observability of the full gyroscope error \nm{\EGYR} that enables its estimation, in contrast with the air data system biases that can not be observed.
\begin{center}
\begin{tabular}{llrrrrrrrrr}
\hline
\multicolumn{2}{l}{Scenario} & \multicolumn{2}{c}{\nm{\psiest - \psi \lrsb{deg}}} & \multicolumn{2}{c}{\nm{\thetaest - \theta \lrsb{deg}}} & \multicolumn{2}{c}{\nm{\xiest - \xi \lrsb{deg}}} & \multicolumn{3}{c}{\nm{\DeltarBestnorm \lrsb{deg}}} \\
\multicolumn{2}{l}{} & \multicolumn{1}{c}{mean} & \multicolumn{1}{c}{std} & \multicolumn{1}{c}{mean} & \multicolumn{1}{c}{std} & \multicolumn{1}{c}{mean} & \multicolumn{1}{c}{std} & \multicolumn{1}{c}{mean} & \multicolumn{1}{c}{std} & \multicolumn{1}{c}{max \nm{| \cdot |}} \\
\hline
\multirow{3}{*}{\nm{\#1}} & mean & +0.036 & 0.104 & -0.022 & 0.036 & +0.000 & 0.038 & \textcolor{red}{\textbf{0.152}} & \textbf{0.078} & \textbf{0.521} \\
                          & std  &  0.131 & 0.015 &  0.036 & 0.014 &  0.027 & 0.010 & \textbf{0.072} & \textbf{0.021} & \textbf{0.151} \\
						  & max  & +0.371 & 0.161 & -0.209 & 0.083 & -0.123 & 0.063 & \textbf{0.448} & \textbf{0.168} & \textbf{1.321} \\
\hline
\multirow{3}{*}{\nm{\#2}} & mean & +0.005 & 0.107 & -0.002 & 0.043 & +0.004 & 0.043 & \textcolor{red}{\textbf{0.118}} & \textbf{0.074} & \textbf{0.376} \\
                          & std  &  0.058 & 0.021 &  0.021 & 0.009 &  0.020 & 0.010 & \textbf{0.023} & \textbf{0.015} & \textbf{0.069} \\
                          & max  & +0.170 & 0.162 & +0.065 & 0.069 & +0.074 & 0.076 & \textbf{0.206} & \textbf{0.124} & \textbf{0.632} \\
\hline
\end{tabular}
\end{center}
\captionof{table}{Aggregated body attitude NSE (100 runs)} \label{tab:attitude_errors}

On a run by run basis, the trajectory metrics indicate that the attitude estimation is lightly biased (non zero mean), both for each Euler angle (yaw \nm{\psi}, pitch \nm{\theta}, roll \nm{\xi}) as well as for the attitude error \nm{\DeltarBestnorm}, defined as the norm of the rotation vector between the estimated attitude and the real one \cite{Shuster1993}. Table \ref{tab:attitude_errors} shows the aggregated metrics of the attitude estimation results for the hundred runs of both scenarios. The estimation of each individual Euler angle tends to be unbiased (zero mean) when aggregated\footnote{Unbiased means that the aggregated mean of means tends to zero as the number of simulation runs increases.}; the aggregated mean of the attitude error is not zero as it is a norm. The aggregated mean of standard deviations is clearly inferior to the mean of means, indicating that the body attitude estimation errors do not deviate far from their mean value as the trajectories progress with time. On the other side, the small values of the aggregated standard deviations when compared with the aggregated means shows that the results are very repeatable and highly independent from the stochastic characteristics of both scenarios that differentiate one run from another. As such, they can be considered bounded. It is also worth noting that the aggregated maximums, although logically bigger, do not indicate any potential problem when considering the wide variety of conditions tested throughout the hundred runs.
\begin{figure}[h]
\centering
\begin{tikzpicture}
\begin{axis}[
colormap name = bluered,
cycle list={[of colormap]}, 
width=8.0cm,
height=5.0cm,
xmin=0, xmax=0.5, xtick={0,0.1,0.2,0.3,0.4,0.5},
xlabel={\nm{\muj{\DeltarBestnorm} \, \lrsb{deg}}},
xmajorgrids,
ymin=0, ymax=0.2, ytick={0,0.1,0.2},
ylabel={\nm{\sigmaj{\DeltarBestnorm} \, \lrsb{deg}}},
ymajorgrids,
axis lines=left,
axis line style={-stealth},
] 
\pgfplotstableread{figs/ch09_sim/error_filter_att/error_filter_att_euler_scatter.txt}\mytable
\addplot+ [scatter, only marks, mark=*,       scatter src=explicit] table [header=false, x index=1,y index=2, meta index=0] {\mytable};
\path node [draw, shape=rectangle, fill=white] at (0.35,0.05) {\footnotesize Scenario \nm{\#1}};
\end{axis}	
\end{tikzpicture}
\hskip 10pt
\begin{tikzpicture}
\begin{axis}[
colormap name = bluered,
cycle list={[of colormap]}, 
width=8.0cm,
height=5.0cm,
xmin=0, xmax=0.5, xtick={0,0.1,0.2,0.3,0.4,0.5},
xlabel={\nm{\muj{\DeltarBestnorm} \, \lrsb{deg}}},
xmajorgrids,
ymin=0, ymax=0.2, ytick={0,0.1,0.2},
ylabel={\nm{\sigmaj{\DeltarBestnorm} \, \lrsb{deg}}},
ymajorgrids,
axis lines=left,
axis line style={-stealth},
] 
\pgfplotstableread{figs/ch09_sim/error_filter_att/error_filter_att_alter_euler_scatter.txt}\mytable
\addplot+ [scatter, only marks, mark=*,       scatter src=explicit] table [header=false, x index=1,y index=2, meta index=0] {\mytable};
\path node [draw, shape=rectangle, fill=white] at (0.35,0.15) {\footnotesize Scenario \nm{\#2}};
\end{axis}	
\end{tikzpicture}
\caption{Trajectory body attitude NSE (100 runs)}
\label{fig:Sim_NSE_Results_euler_scatter}
\end{figure}
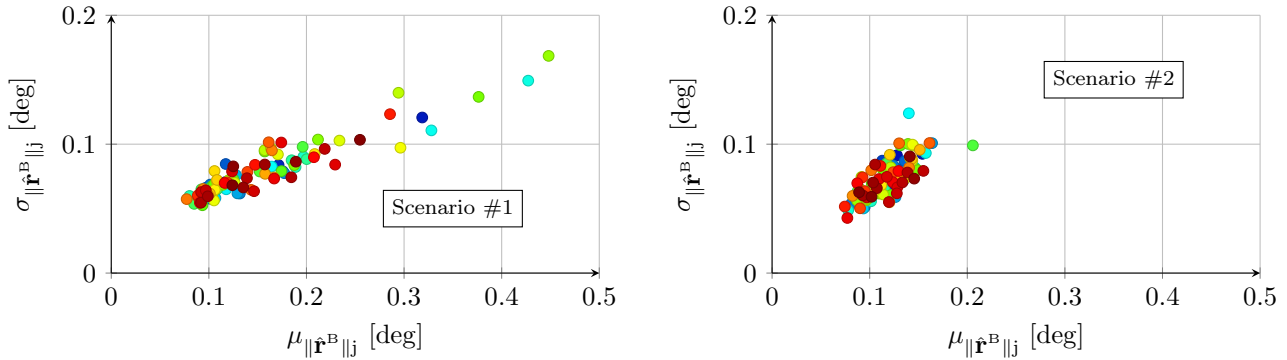

These results show that the proposed attitude filter is capable of closely tracking the aircraft attitude during turbulent flight both when the flight is mostly straight (scenario \nm{\#1}) as when continuously executing maneuvers (scenario \nm{\#2}). Maneuvers increase observability and help the filter to identify and reduce attitude errors, but they also increase the errors in the filter observation equations, so balancing both requirements is one of the main challenges when designing the filter and choosing the filter process and measurement noise values.

Figure \ref{fig:Sim_NSE_Results_euler_scatter} shows the individual body attitude mean and standard deviation for each of the hundred runs and two scenarios. Most are located within a relatively narrow area of similar performance, while in a few cases, which correspond to particularly negative combinations of the random inputs that make up the scenarios, the attitude filter has significantly more difficulty tracking the body attitude.
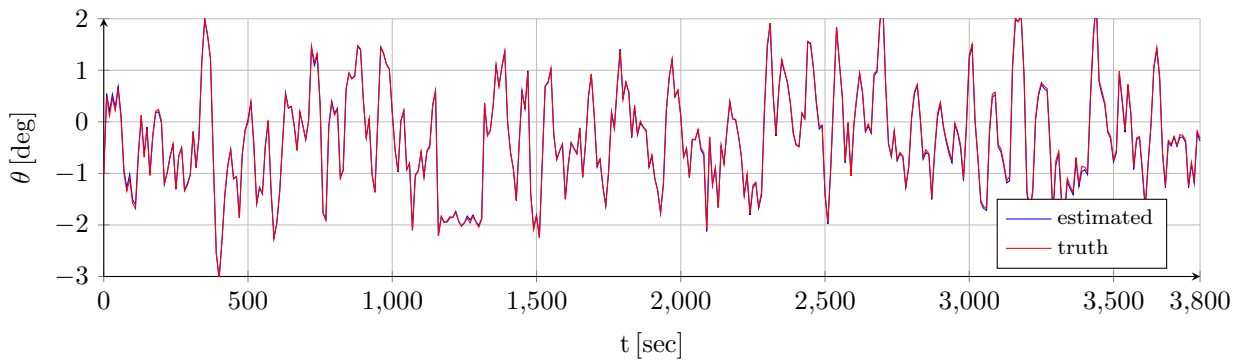
\begin{figure}[h]
\centering
\begin{tikzpicture}
\begin{axis}[
cycle list={{blue,no markers},{red,no markers}},
width=16.0cm,
height=5.0cm,
xmin=0, xmax=3800, xtick={0,500,...,3500,3800},
xlabel={\nm{t \lrsb{sec}}},
xmajorgrids,
ymin=-3, ymax=2, ytick={-3,-2,-1,0,1,2},
ylabel={\nm{\theta \lrsb{deg}}},
ymajorgrids,
axis lines=left,
axis line style={-stealth},
legend entries={estimated, truth},
legend pos=south east,
legend style={font=\footnotesize},
legend cell align=left,
]
\pgfplotstableread{figs/ch09_sim/error_filter_att/error_filter_att_single_euler_deg.txt}\mytable
\addplot table [header=false, x index=0,y index=1] {\mytable};
\addplot table [header=false, x index=0,y index=2] {\mytable};
\end{axis}   
\end{tikzpicture}
\caption{Estimated and true body pitch angles \nm{\theta} for run \nm{\#28} of scenario \nm{\#1}}
\label{fig:Sim_NSE_Results_att_single_euler}
\end{figure}

Figure \ref{fig:Sim_NSE_Results_att_single_euler} plots the body pitch \nm{\theta} and its estimation \nm{\thetaest} for a single run of scenario \nm{\#1}\footnote{The trajectory mean for this run is \nm{0.185 \, \lrsb{deg}}, which is worse than the aggregated mean of \nm{0.152 \, \lrsb{deg}} shown in table \ref{tab:attitude_errors}.}, showing both the significant body pitch oscillations required to maintain pressure altitude with the aircraft subjected to an elevated turbulence level, as well as the accurate estimation performed by the filter.
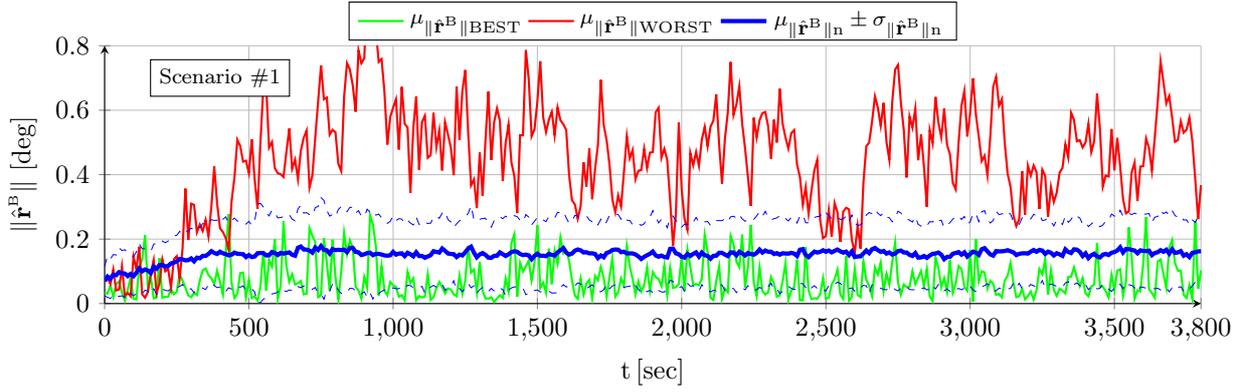
\begin{figure}[h]
\centering
\pgfplotsset{
	every axis legend/.append style={
		at={(0.5,1.02)},
		anchor=south,
	},
}
\begin{tikzpicture}
\begin{axis}[
cycle list={{green,no markers,thick},
            {red,no markers,thick},
            {blue,no markers,ultra thick},
            {blue,dashed,no markers,ultra thin},{blue,dashed,no markers,ultra thin}},
width=16.0cm,
height=5.0cm,
xmin=0, xmax=3800, xtick={0,500,...,3500,3800},
xlabel={\nm{t \lrsb{sec}}},
xmajorgrids,
ymin=0, ymax=0.8, ytick={0,0.2,0.4,0.6,0.8},
ylabel={\nm{\DeltarBestnorm \, \lrsb{deg}}},
ymajorgrids,
axis lines=left,
axis line style={-stealth},
legend entries={\nm{\mu_{\DeltarBestnorm BEST}},
                \nm{\mu_{\DeltarBestnorm WORST}},
				\nm{\mun{\DeltarBestnorm} \pm \sigman{\DeltarBestnorm}}},
legend columns=3,
legend style={font=\footnotesize},
legend cell align=left,
]
\pgfplotstableread{figs/ch09_sim/error_filter_att/error_filter_att_euler_deg.txt}\mytable
\addplot table [header=false, x index=0,y index=4] {\mytable};
\addplot table [header=false, x index=0,y index=5] {\mytable};
\addplot table [header=false, x index=0,y index=1] {\mytable};
\addplot table [header=false, x index=0,y index=2] {\mytable};
\addplot table [header=false, x index=0,y index=3] {\mytable};
\path node [draw, shape=rectangle, fill=white] at (400,0.7) {\footnotesize Scenario \nm{\#1}};
\end{axis}   
\end{tikzpicture}
\caption{Body attitude NSE for scenario \nm{\#1} (100 runs)}
\label{fig:Sim_NSE_Results_euler}
\end{figure}

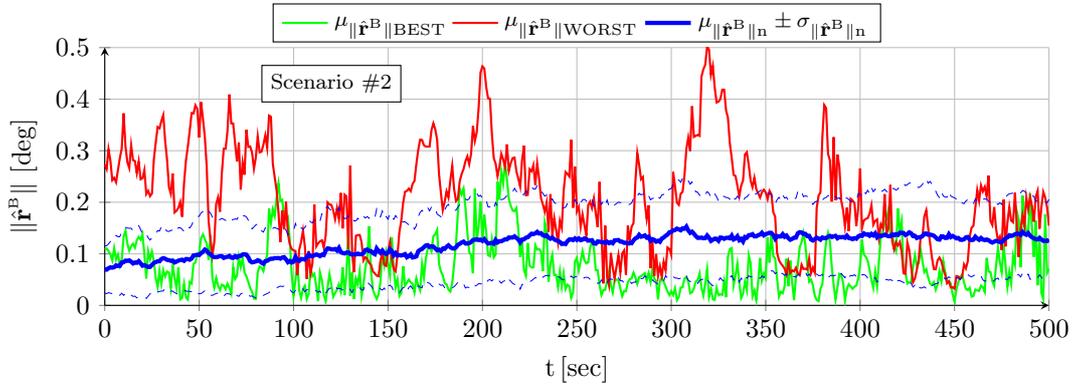
\begin{figure}[h]
\centering
\pgfplotsset{
	every axis legend/.append style={
		at={(0.5,1.02)},
		anchor=south,
	},
}
\begin{tikzpicture}
\begin{axis}[
cycle list={{green,no markers,thick},
            {red,no markers,thick},
            {blue,no markers,ultra thick},
            {blue,dashed,no markers,ultra thin},{blue,dashed,no markers,ultra thin}},
width=14.0cm,
height=5.0cm,
xmin=0, xmax=500, xtick={0,50,...,500},
xlabel={\nm{t \lrsb{sec}}},
xmajorgrids,
ymin=0, ymax=0.5, ytick={0,0.1,0.2,0.3,0.4,0.5},
ylabel={\nm{\DeltarBestnorm \, \lrsb{deg}}},
ymajorgrids,
axis lines=left,
axis line style={-stealth},
legend entries={\nm{\mu_{\DeltarBestnorm BEST}},
                \nm{\mu_{\DeltarBestnorm WORST}},
				\nm{\mun{\DeltarBestnorm} \pm \sigman{\DeltarBestnorm}}},
legend columns=3,
legend style={font=\footnotesize},
legend cell align=left,
]
\pgfplotstableread{figs/ch09_sim/error_filter_att/error_filter_att_alter_euler_deg.txt}\mytable
\addplot table [header=false, x index=0,y index=4] {\mytable};
\addplot table [header=false, x index=0,y index=5] {\mytable};
\addplot table [header=false, x index=0,y index=1] {\mytable};
\addplot table [header=false, x index=0,y index=2] {\mytable};
\addplot table [header=false, x index=0,y index=3] {\mytable};
\path node [draw, shape=rectangle, fill=white] at (120,0.43) {\footnotesize Scenario \nm{\#2}};
\end{axis}   
\end{tikzpicture}
\caption{Body attitude NSE for scenario \nm{\#2} (100 runs)}
\label{fig:Sim_NSE_Results_euler_alter}
\end{figure}

Figures \ref{fig:Sim_NSE_Results_euler} and \ref{fig:Sim_NSE_Results_euler_alter} show the variation with time of the body attitude NSE for both scenarios, complemented with the two runs that result in the best and worst estimations of body attitude. They clearly show that the body attitude estimation has no drift with time, ensuring that the aircraft can remain aloft in GNSS-Denied conditions for as long as there is fuel available.


\subsubsection*{Full Gyroscope Errors Estimation}

Given that the aircraft attitude \nm{\qNB} is the result of integrating the aircraft angular velocity \nm{\wNBB}, which accounts for the majority of the inertial angular velocity \nm{\wIBB}, and that the gyroscope measurements \nm{\wIBBtilde} also include their full error \nm{\EGYR}, the attitude filter would not be able to estimate the body attitude if it were not simultaneously estimating the full gyroscopes error \nm{\EGYRest}.

\nm{\EGYR} encompasses all error sources except system noise \cite{SENSORS}, including the bias offset (constant in each run), the bias drift (which behaves as a random walk), the scale factor error, and the cross coupling error. The last two sources make tracking the error by the attitude filter significantly more challenging, specially during maneuvers, when a swift change in the inertial angular velocity about a given axis does not only result in measurement changes in that same axis, but also in the other two.

The gyroscope error estimation results are qualitatively the same as those of the body attitude for both scenarios, this is, lightly biased for each trajectory, and unbiased or zero mean for each component when aggregated. Being a norm, the full error is also biased when aggregated, but as in the case of the body attitude, it can be considered bounded, and the results are highly repeatable and vary little among different runs.


\subsubsection*{Vertical Position Estimation}

\begin{figure}[h]
\centering
\pgfplotsset{
	every axis legend/.append style={
		at={(0.5,1.02)},
		anchor=south,
	},
}
\begin{tikzpicture}
\begin{axis}[
cycle list={{green,no markers,thick},
            {red,no markers,thick},
            {blue,no markers,ultra thick},
            {blue,dashed,no markers,ultra thin},
            {blue,dashed,no markers,ultra thin}},
width=16.0cm,
height=5.0cm,
xmin=0, xmax=3800, xtick={0,500,...,3500,3800},
xlabel={\nm{t \lrsb{sec}}},
xmajorgrids,
ymin=-80, ymax=40, ytick={-80,-60,...,40},
ylabel={\nm{\hest - h \, \lrsb{m}}},
ymajorgrids,
axis lines=left,
axis line style={-stealth},
legend entries={\nm{\left({\hest \, \lrp{\tEND} - h \, \lrp{\tEND}}\right) \sss{BEST}},
                \nm{\left({\hest \, \lrp{\tEND} - h \, \lrp{\tEND}}\right) \sss{WORST}},
				\nm{\mun{h} \pm \sigman{h}}}, 
legend columns=3,
legend style={font=\footnotesize},
legend cell align=left,
]
\pgfplotstableread{figs/ch09_sim/error_filter_pos/error_filter_pos_h_m.txt}\mytable
\addplot table [header=false, x index=0,y index=4] {\mytable};
\addplot table [header=false, x index=0,y index=5] {\mytable};
\addplot table [header=false, x index=0,y index=1] {\mytable};
\addplot table [header=false, x index=0,y index=2] {\mytable};
\addplot table [header=false, x index=0,y index=3] {\mytable};
\path node [draw, shape=rectangle, fill=white] at (500,-60) {\footnotesize Scenario \nm{\#1}};
\end{axis}   
\end{tikzpicture}
\caption{Vertical position (geometric altitude) NSE for scenario \nm{\#1} (100 runs)}
\label{fig:Sim_NSE_Results_Position_h}
\end{figure}
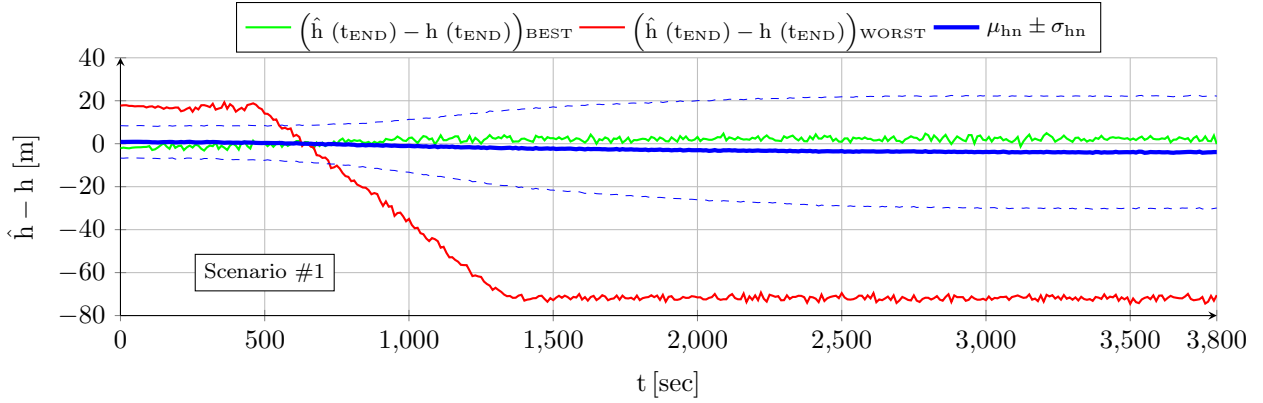

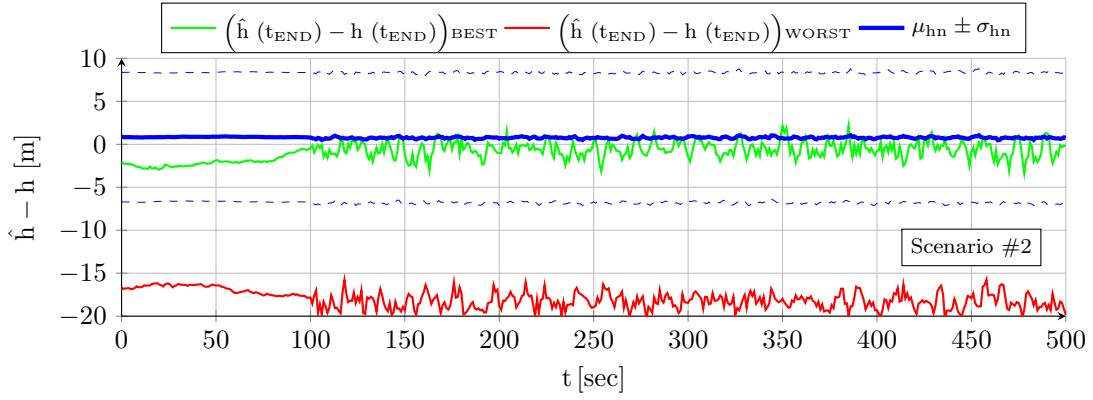
\begin{figure}[h]
\centering
\pgfplotsset{
	every axis legend/.append style={
		at={(0.5,1.02)},
		anchor=south,
	},
}
\begin{tikzpicture}
\begin{axis}[
cycle list={{green,no markers,thick},
            {red,no markers,thick},
            {blue,no markers,ultra thick},
            {blue,dashed,no markers,ultra thin},
            {blue,dashed,no markers,ultra thin}},
width=14.0cm,
height=5.0cm,
xmin=0, xmax=500, xtick={0,50,...,500},
xlabel={\nm{t \lrsb{sec}}},
xmajorgrids,
ymin=-20, ymax=10, ytick={-20,-15,...,10},
ylabel={\nm{\hest - h \, \lrsb{m}}},
ymajorgrids,
axis lines=left,
axis line style={-stealth},
legend entries={\nm{\left({\hest \, \lrp{\tEND} - h \, \lrp{\tEND}}\right) \sss{BEST}},
                \nm{\left({\hest \, \lrp{\tEND} - h \, \lrp{\tEND}}\right) \sss{WORST}},
				\nm{\mun{h} \pm \sigman{h}}}, 
legend columns=3,
legend style={font=\footnotesize},
legend cell align=left,
]
\pgfplotstableread{figs/ch09_sim/error_filter_pos/error_filter_pos_alter_h_m.txt}\mytable
\addplot table [header=false, x index=0,y index=4] {\mytable};
\addplot table [header=false, x index=0,y index=5] {\mytable};
\addplot table [header=false, x index=0,y index=1] {\mytable};
\addplot table [header=false, x index=0,y index=2] {\mytable};
\addplot table [header=false, x index=0,y index=3] {\mytable};
\path node [draw, shape=rectangle, fill=white] at (450,-12) {\footnotesize Scenario \nm{\#2}};
\end{axis}   
\end{tikzpicture}
\caption{Vertical position (geometric altitude) NSE for scenario \nm{\#2} (100 runs)}
\label{fig:Sim_NSE_Results_Position_h_alter}
\end{figure}

Figures \ref{fig:Sim_NSE_Results_Position_h} and \ref{fig:Sim_NSE_Results_Position_h_alter} show the variation with time of the vertical position NSE for both scenarios, complemented with the two runs that result in the lowest and highest altitude final error, respectively. As the scenario \nm{\#1} estimation shows an obvious accuracy degradation with time, the aggregated metrics are not useful \cite{SIMULATION} and are replaced by the aggregated final state metrics, which are shown in table \ref{tab:h_errors}.

From the point of view of the geometric altitude estimation, the only difference between the two scenarios is that \nm{\#1} includes varying atmospheric conditions, while the weather is constant when executing scenario \nm{\#2}. Note that in both cases the aggregated final mean is much smaller than the standard deviation and maximum absolute value, tending to zero as the number of runs grows, which means that the final error is unbiased or zero mean. There are not any intrinsic errors introduced by the estimation process and the individual trajectory errors are caused by random processes either in the observations or in the estimation process itself.
\begin{center}
\begin{tabular}{lrr}
\hline
\nm{\hest \lrp{\tEND} - h \lrp{\tEND}\, \lrsb{m}} & Scenario \nm{\#1} & Scenario \nm{\#2} \\
\hline
mean  & -3.89 & +0.76 \\
std   & \textcolor{red}{\textbf{26.03}} & \textcolor{red}{\textbf{7.55}} \\
max   & \textbf{ -70.49} & \textbf{ -19.86} \\
\hline
\end{tabular}
\end{center}
\captionof{table}{Aggregated final vertical position (geometric altitude) NSE (100 runs)} \label{tab:h_errors}

Careful examination of the above figures reveals several facts. First of all, the geometric altitude estimation error does not increase when the GNSS signals are lost at \nm{\tGNSS = 100 \lrsb{sec}}, although it becomes noisier. Second, the error at the beginning of the trajectories is not only constant, but it is also exactly the same for both scenarios. However, while in the case of scenario \nm{\#2} it remains constant for the whole trajectory, in \nm{\#1} the error increases between two clearly defined time stamps and then remains constant again until the end of the trajectory.

The deviation with time in figure \ref{fig:Sim_NSE_Results_Position_h} is constrained to in between two time stamps, which is the case in all scenario \nm{\#1} runs, as it is not related to the mission altitude change, but to its linear pressure offset \nm{\Deltap} weather change, as explained in \cite{SIMULATION} and section \ref{sec:Simulation}. If the pressure offset variation were modeled differently, then the geometric altitude deviation with time would also vary accordingly. Figure \ref{fig:Sim_NSE_Results_Position_h_alter} does not show any deviation with time because \nm{\Deltap} is constant in scenario \nm{\#2}.

At the beginning of both scenarios, when GNSS signals are present, the position filter employs additional position and ground velocity observation equations, and hence the ionospheric effects constitute the main source of error for the vertical position estimation. In addition, the navigation filter does not only estimate the temperature offset \nm{\DeltaT} but also the pressure offset \nm{\Deltap}, incurring in relatively small errors caused not only by the ionospheric effects but also by the thermometer and barometer bias offsets.

When the GNSS signals are lost, the filter does not have enough observations to properly estimate \nm{\Deltap}, so it keeps the estimation constant per (\ref{eq:filter_pos_Deltap}). This process thus incorporates the pressure offset variation between the time at which the geometric altitude is estimated and the time at which the GNSS signals are lost as the fourth primary error source. Finally, the estimation of the geometric altitude in GNSS-Denied conditions by means of (\ref{eq:GNC_Navigation_AirDataFilterGNSSDenied_H}) and (\ref{eq:GNC_Navigation_AirDataFilterGNSSDenied_h}) effectively reverses the process performed to obtain \nm{\Deltap} when GNSS was available, eliminating the dependencies with the bias offsets of the thermometer and barometer, which are constant for a given run.

The final vertical position estimation error hence depends on two factors, the GNSS ionospheric errors and the variation in atmospheric pressure offset \nm{\DeltapEND - \DeltapINI}, both bounded by atmospheric physics. However, while the ionospheric effects represent a lower error threshold that can not be avoided even by using GNSS, the pressure offset change is only a factor in the case of GNSS-Denied navigation and disappears when employing the GNSS signals. The geometric altitude estimation hence does not experience an unbounded drift but an error with known bounded sources.

To verify this explanation, the one hundred runs of scenario \nm{\#1} have been repeated with the only difference being that the GNSS signals are always available (\nm{\tGNSS > 3800 \lrsb{sec}}), resulting in \nm{+0.99}, \nm{7.51}, and \nm{+17.09 \lrsb{m}} for the aggregated final mean, standard deviation, and maximum absolute value, respectively. Note that these values are very similar to those shown for scenario \nm{\#2} in table \ref{tab:h_errors}.

The values of the final state metrics shown in table \ref{tab:h_errors} are adequate for all navigation applications except landing. As explained in section \ref{sec:Comparison}, estimating the geometric altitude this way is intrinsically superior to integrating the vertical speed, in which the error grows with time without bounds.


\subsubsection*{Ground Velocity Estimation}

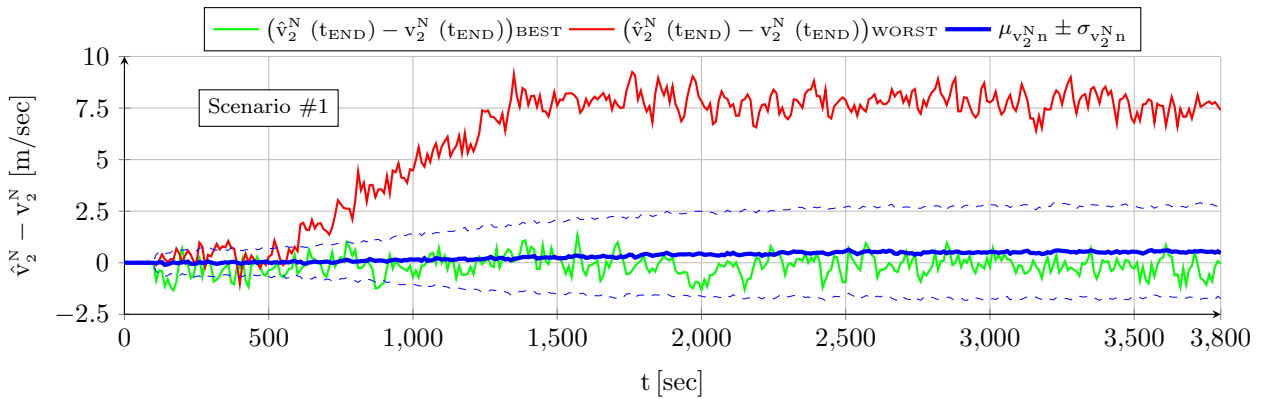
\begin{figure}[h]
\centering
\pgfplotsset{
	every axis legend/.append style={
		at={(0.5,1.02)},
		anchor=south,
	},
}
\begin{tikzpicture}
\begin{axis}[
cycle list={{green,no markers,thick},
            {red,no markers,thick},
            {blue,no markers,ultra thick},
            {blue,dashed,no markers,ultra thin},
   			{blue,dashed,no markers,ultra thin}},
width=16.0cm,
height=5.0cm,
xmin=0, xmax=3800, xtick={0,500,...,3500,3800},
xlabel={\nm{t \lrsb{sec}}},
xmajorgrids,
ymin=-2.5, ymax=10, ytick={-2.5,0,2.5,5,7.5,10.0},
ylabel={\nm{\vNestii - \vNii \, \lrsb{m/sec}}},
ymajorgrids, 
axis lines=left,
axis line style={-stealth},
legend entries={\nm{\left({\vNestii \, \lrp{\tEND} - \vNii \, \lrp{\tEND}}\right) \sss{BEST}},
                \nm{\left({\vNestii \, \lrp{\tEND} - \vNii \, \lrp{\tEND}}\right) \sss{WORST}},
				\nm{\mun{\vNii} \pm \sigman{\vNii}}}, 
legend columns=3,
legend style={font=\footnotesize},
legend cell align=left,
]
\pgfplotstableread{figs/ch09_sim/error_filter_pos/error_filter_pos_vned_mps.txt}\mytable
\addplot table [header=false, x index=0,y index=9] {\mytable};
\addplot table [header=false, x index=0,y index=10] {\mytable};
\addplot table [header=false, x index=0,y index=6] {\mytable};
\addplot table [header=false, x index=0,y index=7] {\mytable};
\addplot table [header=false, x index=0,y index=8] {\mytable};
\path node [draw, shape=rectangle, fill=white] at (500,7.5) {\footnotesize Scenario \nm{\#1}};
\end{axis}   
\end{tikzpicture}
\caption{East ground speed NSE for scenario \nm{\#1} (100 runs)}
\label{fig:Sim_NSE_Results_Position_vnedii}
\end{figure}

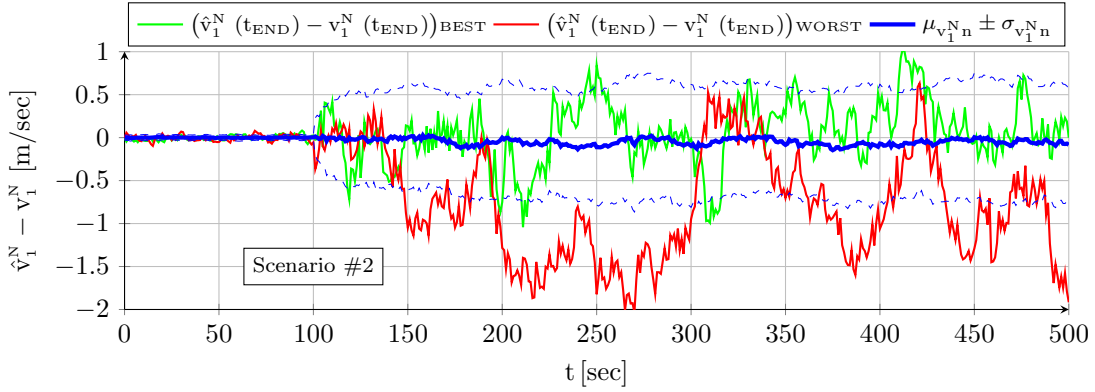
\begin{figure}[h]
\centering
\pgfplotsset{
	every axis legend/.append style={
		at={(0.5,1.02)},
		anchor=south,
	},
}
\begin{tikzpicture}
\begin{axis}[
cycle list={{green,no markers,thick},
            {red,no markers,thick},
            {blue,no markers,ultra thick},
            {blue,dashed,no markers,ultra thin},
   			{blue,dashed,no markers,ultra thin}},
width=14.0cm,
height=5.0cm,
xmin=0, xmax=500, xtick={0,50,...,500},
xlabel={\nm{t \lrsb{sec}}},
xmajorgrids,
ymin=-2, ymax=1, ytick={-2,-1.5,-1,-0.5,0,0.5,1},
ylabel={\nm{\vNesti - \vNi \, \lrsb{m/sec}}},
ymajorgrids,
axis lines=left,
axis line style={-stealth},
legend entries={\nm{\left({\vNesti \, \lrp{\tEND} - \vNi \, \lrp{\tEND}}\right) \sss{BEST}},
                \nm{\left({\vNesti \, \lrp{\tEND} - \vNi \, \lrp{\tEND}}\right) \sss{WORST}},
				\nm{\mun{\vNi} \pm \sigman{\vNi}}}, 
legend columns=3,
legend style={font=\footnotesize},
legend cell align=left,
]
\pgfplotstableread{figs/ch09_sim/error_filter_pos/error_filter_pos_alter_vned_mps.txt}\mytable
\addplot table [header=false, x index=0,y index=4] {\mytable};
\addplot table [header=false, x index=0,y index=5] {\mytable};
\addplot table [header=false, x index=0,y index=1] {\mytable};
\addplot table [header=false, x index=0,y index=2] {\mytable};
\addplot table [header=false, x index=0,y index=3] {\mytable};
\path node [draw, shape=rectangle, fill=white] at (100,-1.5) {\footnotesize Scenario \nm{\#2}};
\end{axis}   
\end{tikzpicture}
\caption{North ground speed NSE for scenario \nm{\#2} (100 runs)}
\label{fig:Sim_NSE_Results_Position_vnedi_alter}
\end{figure}

The explanation of the ground velocity NSE results is very similar to that of the vertical position in the previous section. Figure \ref{fig:Sim_NSE_Results_Position_vnedii} shows the variation with time of the East ground velocity NSE for scenario \nm{\#1}, while figure \ref{fig:Sim_NSE_Results_Position_vnedi_alter} represents that of the North velocity NSE for scenario \nm{\#2}. As an accuracy degradation with time is present in scenario \nm{\#1}, table \ref{tab:v_errors} employs aggregated final state metrics to show the results \cite{SIMULATION}. From the point of view of the ground velocity estimation, the only difference between both scenarios is that \nm{\#1} includes varying wind, which is constant in scenario \nm{\#2}.

As with the vertical position, the estimation process does not introduce any errors, so the final errors are unbiased or zero mean. These figures reveal that the ground velocity estimation error is extremely small when GNSS signals are available, but suddenly increases when the GNSS signals are lost at \nm{\tGNSS = 100 \lrsb{sec}}, after which they are approximately stable. These errors are exactly the same for both scenarios. However, while in the case of scenario \nm{\#2} the errors remain approximately constant for the whole trajectory, in \nm{\#1} they increase again between two clearly defined time stamps, after which they remain constant until the end of the trajectory.

The deviation with time in figure \ref{fig:Sim_NSE_Results_Position_vnedii} is constrained to in between two time stamps, which is the case in all scenario \nm{\#1} runs, as it is not related to the mission airspeed change but to its linear wind field change, as explained in \cite{SIMULATION} and section \ref{sec:Simulation}. If the wind field variation were modeled differently, then the ground speed deviation with time would also vary accordingly. Figure \ref{fig:Sim_NSE_Results_Position_vnedi_alter} does not show any deviation with time because \nm{\vWINDN} is constant in scenario \nm{\#2}.
\begin{center}
\begin{tabular}{lrrrr}
\hline
\nm{\vNest \lrp{\tEND} - \vN \lrp{\tEND}} & \multicolumn{2}{c}{Scenario \nm{\#1}} & \multicolumn{2}{c}{Scenario \nm{\#2}} \\
\nm{\lrsb{m/sec}} & \nm{\vNesti - \vNi} & \nm{\vNestii - \vNii} & \nm{\vNesti - \vNi} & \nm{\vNestii - \vNii} \\
\hline
mean & +0.32           & +0.46          & -0.08          &   -0.14 \\
std  & \textbf{2.59}   & \textbf{2.20}  & \textbf{0.65}  & \textbf{0.63}\\
max  & \textbf{ +9.32} & \textbf{+7.39} & \textbf{-1.92} & \textbf{-1.79} \\
\hline
\end{tabular}
\end{center}
\captionof{table}{Aggregated final ground speed NSE (100 runs)} \label{tab:v_errors}

The estimation of the airspeed \nm{\vTASNest} per (\ref{eq:filter_pos_vTASN}) is the same independently of whether GNSS signals are available or not. It is a combination of the estimated body attitude \nm{\qNBest} and the airspeed vector viewed in body \nm{\vTASBest}\footnote{\nm{\vTASB} is a function of the airspeed \nm{\vtas}, the angle of attack \nm{\alpha}, and the angle of sideslip \nm{\beta}, all estimated by the air data filter.}. Both are biased but drift-less on a trajectory basis, with their errors bounded by the quality of the sensors, and hence so is \nm{\vTASNest}.

If GNSS signals are available, the position filter accurately estimates the ground velocity \nm{\vNest} mostly based on the unbiased observations provided by the GNSS receiver and then subtracts \nm{\vTASNest} to estimate the wind field \nm{\vWINDNest}, so the wind estimation errors are mostly those of \nm{\vTASNest} but with the opposite sign. The process is reversed in GNSS-Denied conditions (\ref{eq:filter_pos_vN}), as the wind estimation is frozen from its value when the GNSS signals are lost (\ref{eq:filter_pos_vWINDN}), and the ground velocity is estimated by adding \nm{\vTASNest} and the frozen wind estimation. As such, the GNSS-Denied \nm{\vNest} error also includes any change in the wind velocity that occurs from the time the GNSS signals are lost to the time the ground speed is estimated.

The ground velocity estimation error is caused by the three factors mentioned above: the errors in the airspeed vector estimation \nm{\vTASBest}, the errors in the body attitude estimation \nm{\qNBest}, and the horizontal wind variation experienced by the aircraft from the time the GNSS signals are lost to the time its ground speed is estimated \nm{\vWINDNEND - \vWINDNINI}. The first two are responsible for the sudden increase in the error at \nm{\tGNSS = 100 \lrsb{sec}} that can be observed in both scenarios, while the third is the culprit of the additional error that only occurs in \nm{\#1}.

In practical terms, the first two error sources are bounded by sensor quality and establish a threshold for the ground velocity estimation accuracy, while the wind variation, although bounded by atmospheric physics, has the most potential to induce errors in the ground speed estimation, which hence does not present an unbounded drift but a bounded error with known bounded sources. As explained in section \ref{sec:Comparison}, estimating the ground velocity this way is intrinsically superior to integrating the specific force, in which the error grows with time without any limitations.


\subsubsection*{Horizontal Position Estimation}

\begin{figure}[h]
\centering
\begin{tikzpicture}
\begin{axis}[
cycle list={{green,no markers,thick},
			{red,no markers,thick},
			{blue,no markers,ultra thick},
			{blue,dashed,no markers,ultra thin},
			{blue,dashed,no markers,ultra thin}},
width=16.0cm,
height=5.0cm,
xmin=0, xmax=3800, xtick={0,500,...,3500,3800},
xlabel={\nm{t \lrsb{sec}}},
xmajorgrids, 
ymin=0, ymax=24, ytick={0,4,8,12,16,20,24},
ylabel={\nm{\Deltaxhorest \, \lrsb{km}}},
ymajorgrids,
axis lines=left,
axis line style={-stealth},
legend entries={\nm{\lrp{\Deltaxhorest \, \lrp{\tEND}} \sss{BEST}},
                \nm{\lrp{\Deltaxhorest \, \lrp{\tEND}} \sss{WORST}},
				\nm{\mun{\Deltaxhorest} \pm \sigman{\Deltaxhorest}}},
 legend pos=north west,
legend style={font=\footnotesize},
legend cell align=left,
]
\pgfplotstableread{figs/ch09_sim/error_filter_pos/error_filter_pos_hor_m_pc.txt}\mytable
\addplot table [header=false, x index=0,y index=4] {\mytable};
\addplot table [header=false, x index=0,y index=5] {\mytable};
\addplot table [header=false, x index=0,y index=1] {\mytable};
\addplot table [header=false, x index=0,y index=2] {\mytable};
\addplot table [header=false, x index=0,y index=3] {\mytable};
\path node [draw, shape=rectangle, fill=white] at (500,4) {\footnotesize Scenario \nm{\#1}};
\end{axis}   
\end{tikzpicture}
\caption{Horizontal position NSE for scenario \nm{\#1} (100 runs)}
\label{fig:Sim_NSE_Results_xhor}
\end{figure}
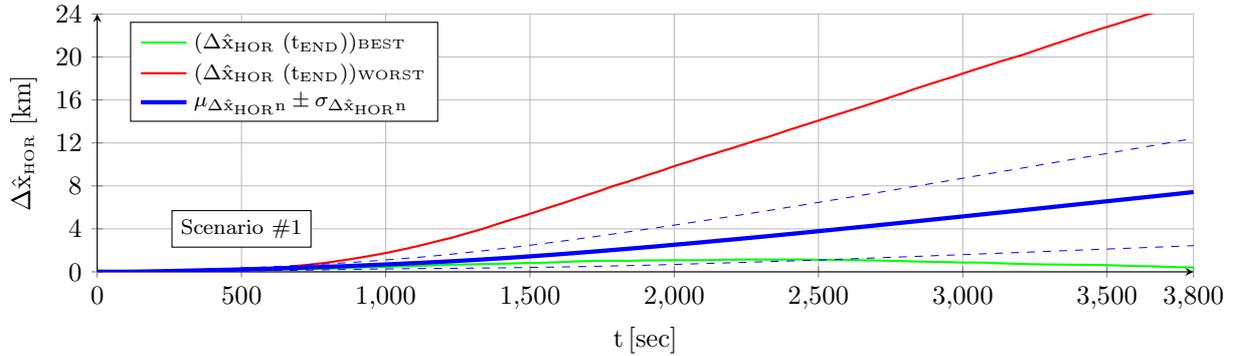

In GNSS-Denied conditions the horizontal position is estimated by the position filter through the integration of the ground velocity without any observations to reset the integration errors. Figures \ref{fig:Sim_NSE_Results_xhor} and \ref{fig:Sim_NSE_Results_xhor_alter} show the variation with time of the horizontal position NSE for both scenarios, together with the two runs with the highest and lowest final error, respectively.

The approximately linear position drift observed in both figures is caused by the integration of the approximately constant horizontal velocity error, and can not be avoided unless additional observations are included into the GNSS-Denied position filter. The only way to eliminate the position drift with a standard suite of sensors in GNSS-Denied conditions is to reduce the ground velocity error, whose origins are explained in detail in the previous section.

As in previous cases, aggregated final state metrics are employed since drift is present, and these are shown in table \ref{tab:hor_errors} for both scenarios. Note that this table includes the horizontal distance error both in absolute terms as well as a relative percentage with respect to the distance flown since the GNSS signals are lost. In addition to the horizontal error, which is a norm and hence always positive, table \ref{tab:hor_errors} also includes the cross track and long track errors, which are positive when the estimated position lies to the right or ahead of the actual position, and negative otherwise. These are unbiased or zero mean, indicating that the error source is in the data (the ground velocity) and not in the algorithm itself (the integration).
\begin{figure}[h]
\centering
\begin{tikzpicture}
\begin{axis}[
cycle list={{green,no markers,thick},
			{red,no markers,thick},
			{blue,no markers,ultra thick},
			{blue,dashed,no markers,ultra thin},
			{blue,dashed,no markers,ultra thin}},
width=14.0cm,
height=5.0cm,
xmin=0, xmax=500, xtick={0,50,...,500},
xlabel={\nm{t \lrsb{sec}}},
xmajorgrids,
ymin=0, ymax=0.6, ytick={0,0.1,0.2,0.3,0.4,0.5,0.6},
ylabel={\nm{\Deltaxhorest \, \lrsb{km}}},
ymajorgrids,
axis lines=left,
axis line style={-stealth},
legend entries={\nm{\lrp{\Deltaxhorest \, \lrp{\tEND}}_{BEST}},
				\nm{\lrp{\Deltaxhorest \, \lrp{\tEND}}_{WORST}},
				\nm{\mun{\Deltaxhorest} \pm \sigman{\Deltaxhorest}}},
legend pos=north west,
legend style={font=\footnotesize},
legend cell align=left,
]
\pgfplotstableread{figs/ch09_sim/error_filter_pos/error_filter_pos_alter_hor_m_pc.txt}\mytable
\addplot table [header=false, x index=0,y index=4] {\mytable};
\addplot table [header=false, x index=0,y index=5] {\mytable};
\addplot table [header=false, x index=0,y index=1] {\mytable};
\addplot table [header=false, x index=0,y index=2] {\mytable};
\addplot table [header=false, x index=0,y index=3] {\mytable};
\path node [draw, shape=rectangle, fill=white] at (100,0.2) {\footnotesize Scenario \nm{\#2}};
\end{axis}   
\end{tikzpicture}
\caption{Horizontal position NSE for scenario \nm{\#2} (100 runs)}
\label{fig:Sim_NSE_Results_xhor_alter}
\end{figure}
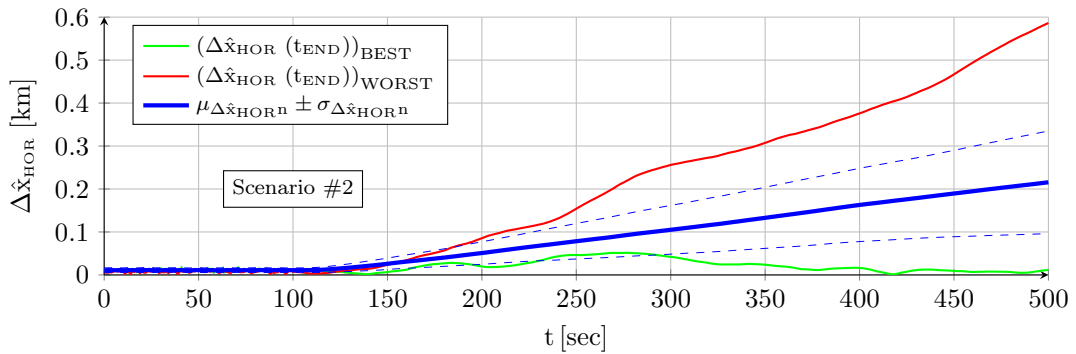
\begin{center}
\begin{tabular}{llrrrrrrr}
\hline
\multicolumn{2}{l}{Scenario} & \multicolumn{1}{c}{Distance} & \multicolumn{2}{c}{\nm{\Deltaxcrossest \lrp{\tEND}}} & \multicolumn{2}{c}{\nm{\Deltaxlongest \lrp{\tEND}}} & \multicolumn{2}{c}{\nm{\Deltaxhorest \lrp{\tEND}}} \\
\multicolumn{2}{c}{} & \multicolumn{1}{c}{\nm{\lrsb{m}}} & \multicolumn{1}{c}{\nm{\lrsb{m}}} &  \multicolumn{1}{c}{\nm{\lrsb{\%}}} & \multicolumn{1}{c}{\nm{\lrsb{m}}} &  \multicolumn{1}{c}{\nm{\lrsb{\%}}} & \multicolumn{1}{c}{\nm{\lrsb{m}}} & \multicolumn{1}{c}{\nm{\lrsb{\%}}} \\
\hline
\multirow{3}{*}{\nm{\#1}} & mean	& 108623 &        +59 &   +0.13 &        +41 &   +0.08 & \textbf{      7428} & \textcolor{red}{ \textbf{  7.18}} \\
                          & std     &  19935 &      +5615 &   +5.07 &      +6965 &   +7.66 & \textbf{      4988} & \textbf{  5.73} \\
                          & max     & 172842 &     -18954 &  -16.40 &     +20064 &  +32.22 & \textbf{     25288} & \textbf{ 32.38} \\
\hline
\multirow{3}{*}{\nm{\#2}} & mean    &  14198 &        -37 &   -0.26 &        +22 &   +0.15 & \textbf{       216} & \textcolor{red}{ \textbf{  1.52}} \\
                          & std     &   1176 &       +162 &   +1.15 &       +180 &   +1.28 & \textbf{       119} & \textbf{  0.86} \\
                          & max     &  18253 &       +562 &   +4.20 &       +537 &   +4.12 & \textbf{       586} & \textbf{  4.38} \\
\hline
\end{tabular}
\end{center}
\captionof{table}{Aggregated final horizontal position NSE (100 runs)} \label{tab:hor_errors}

The results for scenario \nm{\#1}, which contains a significant wind field variation, show that in most cases the deviations from the intended route are not severe, but a mean error of \nm{7.18 \lrsb{\%}} and standard deviation of \nm{5.73 \lrsb{\%}} are not sufficient to ensure safe GNSS-Denied navigation under any circumstances. They translate into absolute position errors that are in the order of several kilometers after one hour of flight, which is not acceptable as the risk of collision with the terrain, structures, or other aircraft is too high. The results for scenario \nm{\#2}, which are optimistic since it assumes zero wind variation, can be considered as a best case threshold showing the influence of the remaining sources of error: the airspeed sensors quality influencing \nm{\vTASBest}, the inertial sensors quality for \nm{\qNBest}, and the ground velocity integration errors.
\begin{center}
\begin{tabular}{lrrrrr}
\hline
& \multicolumn{1}{c}{Distance} & \multicolumn{2}{c}{\nm{\Deltaxhorest \lrp{\tEND}}} & \multicolumn{2}{c}{\nm{\vWIND} accum.} \\
& \multicolumn{1}{c}{\nm{\lrsb{m}}} & \multicolumn{1}{c}{\nm{\lrsb{m}}} &  \multicolumn{1}{c}{\nm{\lrsb{\%}}} & \multicolumn{1}{c}{\nm{\lrsb{m}}} & \multicolumn{1}{c}{\nm{\lrsb{\%}}} \\
\hline
mean   &      108623 &       7428 &   7.18 &       7121 &   6.96 \\
std    &       19935 &       4988 &   5.73 &       4988 &   5.96 \\
max    &      172842 &      25288 &  32.38 &      23630 &  32.95 \\
\hline
\end{tabular}
\end{center}
\captionof{table}{Final horizontal position NSE analysis for scenario \nm{\#1}} \label{tab:Sim_NSE_Results_compare_xhor}

Table \ref{tab:Sim_NSE_Results_compare_xhor} compares the scenario \nm{\#1} results with those obtained by means of a simple time integral of the wind variation with respect to that existing when the GNSS signals are lost. The similarity with the aggregated final state metrics shown in table \ref{tab:hor_errors} proves that the wind variation is by far the main responsible for the horizontal position drift. The selected approach to estimating the horizontal position, although resulting in a drift with time, is intrinsically superior to a double integration of the acceleration, as explained in section \ref{sec:Comparison}.


\section{Comparison with Other Navigation Algorithms}\label{sec:Comparison}

This section compares the results obtained with the proposed navigation filter, referred below as the \say{baseline}, against those obtained when employing different algorithms for the estimation of the horizontal position, vertical position, and body attitude. These algorithms are more representative of those employed by a traditional GNSS-Based inertial navigation filter that at a given time is forced to operate without the observations provided by its GNSS receiver.


\subsubsection*{Different Horizontal Position Algorithms}

The position filter relies on freezing the wind speed estimation (\ref{eq:filter_pos_vWINDN}) at the value it has when the GNSS signals are lost, which may not appear as optimum as it allows the horizontal position estimation error to grow linearly with time with a slope proportional to the wind change since the beginning of GNSS-Denied navigation.

In this section the baseline approach is compared with two alternative implementations that provide better results for the first few minutes of GNSS-Denied navigation. However, as they have a higher reliance on integration, their error growth with time is of higher order than in the baseline, resulting in much higher estimation errors. As a matter of fact, the ground velocity errors are so significant as time grows that they destabilize the attitude filter by means of (\ref{eq:filter_att_obs_pvec}) and (\ref{eq:filter_att_obs_pvec_dependencies}), and hence it is necessary to shorten the scenario \nm{\#1} simulations to \nm{\tEND = 1000 \, \lrsb{sec}}. This could have been mitigated by further modifications to the attitude filter, which have not been implemented as the results are in any case clearly inferior to those of the baseline.

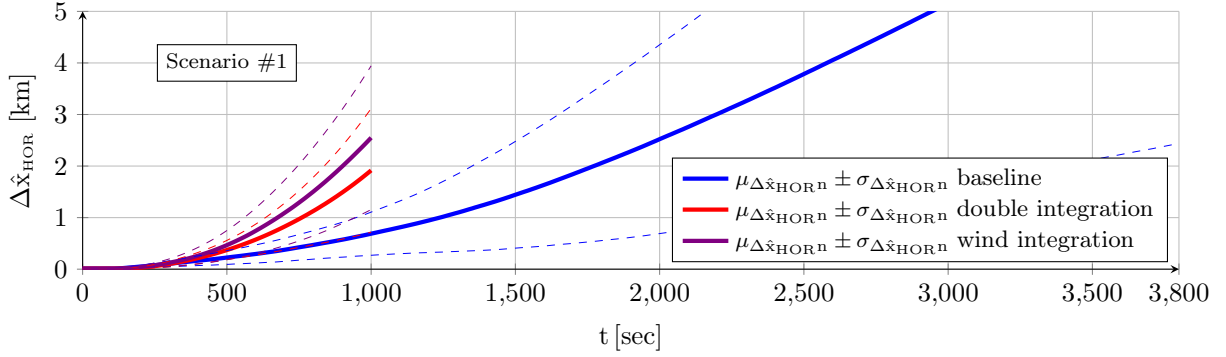
\begin{figure}[h]
\centering
\pgfplotsset{
	every axis legend/.append style={
		at={(0.99,0.23)},
		anchor=east,
	},
}
\begin{tikzpicture}
\begin{axis}[
cycle list={{blue,no markers,ultra thick},
			{red,no markers,ultra thick},
			{violet,no markers,ultra thick},
			{blue,dashed,no markers,ultra thin},{blue,dashed,no markers,ultra thin},
            {red,dashed,no markers,ultra thin}, {red,dashed,no markers,ultra thin}, 
			{violet,dashed,no markers,ultra thin}, {violet,dashed,no markers,ultra thin}},
width=16.0cm,
height=5.0cm,
xmin=0, xmax=3800, xtick={0,500,...,3500,3800},
xlabel={\nm{t \lrsb{sec}}},
xmajorgrids,
ymin=0, ymax=5, ytick={0,1,...,5},
ylabel={\nm{\Deltaxhorest \, \lrsb{km}}},
ymajorgrids,
axis lines=left,
axis line style={-stealth},
legend entries={\nm{\mun{\Deltaxhorest} \pm \sigman{\Deltaxhorest}} \small{baseline},
                \nm{\mun{\Deltaxhorest} \pm \sigman{\Deltaxhorest}} \small{double integration},
				\nm{\mun{\Deltaxhorest} \pm \sigman{\Deltaxhorest}} \small{wind integration}},
legend style={font=\footnotesize},
legend columns=1,
legend cell align=left,
]
\pgfplotstableread{figs/ch10_ana/versus_integr/versus_integr_pos_hor_m_pc_part1.txt}\mytable
\addplot table [header=false, x index=0,y index=1] {\mytable};
\pgfplotstableread{figs/ch10_ana/versus_integr/versus_integr_pos_hor_m_pc_part2.txt}\mytable
\addplot table [header=false, x index=0,y index=1] {\mytable};
\addplot table [header=false, x index=4,y index=5] {\mytable};
\pgfplotstableread{figs/ch10_ana/versus_integr/versus_integr_pos_hor_m_pc_part1.txt}\mytable
\addplot table [header=false, x index=0,y index=2] {\mytable};
\addplot table [header=false, x index=0,y index=3] {\mytable};
\pgfplotstableread{figs/ch10_ana/versus_integr/versus_integr_pos_hor_m_pc_part2.txt}\mytable
\addplot table [header=false, x index=0,y index=2] {\mytable};
\addplot table [header=false, x index=0,y index=3] {\mytable};
\addplot table [header=false, x index=4,y index=6] {\mytable};
\addplot table [header=false, x index=4,y index=7] {\mytable};
\path node [draw, shape=rectangle, fill=white] at (500,4) {\footnotesize Scenario \nm{\#1}};
\end{axis}   
\end{tikzpicture}
\caption{Influence of filter algorithms on horizontal position NSE for scenario \nm{\#1} (100 runs)}
\label{fig:Ana_NSE_Results_xhor_integr}
\end{figure}

\begin{itemize}
\item The first alternative, named \say{double integration}, behaves as a standard GNSS-Based inertial filter \cite{Farrell2008, Groves2008} working without the periodic ground speed and position observations provided by the GNSS receiver. It relies on discarding (\ref{eq:filter_pos_vWINDN}), (\ref{eq:filter_pos_vTASN}), and (\ref{eq:filter_pos_vN}), and instead estimates the ground speed \nm{\vNest} by integrating its time derivative, which in turn is obtained based on (\ref{eq:specific_force_definition_ned}) from the estimations of the specific force \nm{\fIBBest} and the body attitude \nm{\qNBest}:
\begin{eqnarray}
\nm{\vNdotest} & = & \nm{\qNBest \otimes \fIBBest \otimes \qNBastest - \wENNestskew \; {\hat{\vec v}}_{n-1}^{\sss N} + \gcNMODELest - \acorNest} \label{eq:Ana_PosFilter_xhor_double_vndot} \\
\nm{\vNest}    & = & \nm{{\hat{\vec v}_{n-1}^{\sss N}} + \DeltatEST \cdot \vNdotest}\label{eq:Ana_PosFilter_xhor_double_vn}
\end{eqnarray}

Note that the motion angular velocity \nm{\wENNest} and the Coriolis acceleration \nm{\acorNest} depend on position and velocity, while the model gravity acceleration \nm{\gcNMODELest} is computed based exclusively on position. In (\ref{eq:Ana_PosFilter_xhor_double_vndot}) all three are evaluated based on the previous step (\nm{n-1}). As in the baseline, the position is then estimated by integrating the ground velocity. The main weakness of this approach is the low accuracy of the specific force estimation \nm{\fIBBest} in the GNSS-Denied position filter, which is integrated twice to obtain the aircraft position.

This approach is valid for GNSS-Based conditions, not only because the velocity and position estimations on which \nm{\wENNest}, \nm{\acorNest}, and \nm{\gcNMODELest} rely are more accurate, but also because the GNSS observations impede the accumulation of integration errors. Without these observations, nothing prevents the errors from accumulating as the time without GNSS signals increases.

\item The second alternative, named \say{wind integration}, also estimates the ground speed time derivative per (\ref{eq:Ana_PosFilter_xhor_double_vndot}). This approach however tries to reduce the integration errors required to obtain \nm{\vNest} per (\ref{eq:Ana_PosFilter_xhor_double_vn}) by only integrating the wind derivative \nm{\vWINDNdotest} instead of the complete ground speed derivative \nm{\vNdotest} \cite{PATENT}. As the proposed filter, it takes advantage of the airspeed sensors present in fixed wing aircraft that however are generally employed for control purposes exclusively, and not for navigation. To do so, it is first necessary to directly estimate the true airspeed time derivative viewed in body \nm{\vTASBdotest} by using the airspeed vector time derivative estimations (\nm{\vtasdotest}, \nm{\alphadotest}, \nm{\betadotest}) provided by the air data filter. From there, the true airspeed derivative with time in NED (\nm{\vTASNdotest}) is obtained per (\ref{eq:Ana_PosFilter_xhor_wind_vtasN}). The estimated ground speed \nm{\vNest} can then be obtained by combining the true airspeed with the integrated wind speed (\ref{eq:Ana_PosFilter_xhor_double_vN}). As in the first alternative,the position is also obtained by integrating the resulting ground velocity.
\begin{eqnarray}
\nm{\vTASNdotest} & = & \nm{\qNBest \otimes \lrp{\vTASBdotest + \wNBBest \, \vTASBest} \otimes \qNBastest}\label{eq:Ana_PosFilter_xhor_wind_vtasN} \\
\nm{\vWINDNdotest} & = & \nm{\vNdotest - \vTASNdotest} \label{eq:Ana_PosFilter_xhor_wind_vwinddotN} \\
\nm{\vWINDNest} & = & \nm{{\hat{\vec v}_{{\sss{WIND}},n-1}^{\sss N}} + \DeltatEST \cdot \vWINDNdotest}\label{eq:Ana_PosFilter_xhor_double_vwindN} \\
\nm{\vNest} & = & \nm{\vWINDNest + \qNBest \otimes \vTASB\lrp{\vtasest, \, \alphaest, \, \betaest} \otimes \qNBastest}\label{eq:Ana_PosFilter_xhor_double_vN}
\end{eqnarray}
\end{itemize}
 
\begin{figure}[h]
\centering
\pgfplotsset{
	every axis legend/.append style={
		at={(0.01,0.75)},
		anchor=west,
	},
}
\begin{tikzpicture}
\begin{axis}[
cycle list={{blue,no markers,ultra thick},
			{red,no markers,ultra thick},
			{violet,no markers,ultra thick},
			{blue,dashed,no markers,ultra thin},{blue,dashed,no markers,ultra thin},
            {red,dashed,no markers,ultra thin}, {red,dashed,no markers,ultra thin},
			{violet,dashed,no markers,ultra thin}, {violet,dashed,no markers,ultra thin}},
width=16.0cm,
height=5.0cm,
xmin=0, xmax=300, xtick={0,50,...,300},
xlabel={\nm{t \lrsb{sec}}},
xmajorgrids,
ymin=0, ymax=0.2, ytick={0,0.05,0.1,0.15,0.2},
ylabel={\nm{\Deltaxhorest \, \lrsb{km}}},
ymajorgrids,
axis lines=left,
axis line style={-stealth},
legend entries={\nm{\mun{\Deltaxhorest} \pm \sigman{\Deltaxhorest}} \small{baseline},
                \nm{\mun{\Deltaxhorest} \pm \sigman{\Deltaxhorest}} \small{double integration},
				\nm{\mun{\Deltaxhorest} \pm \sigman{\Deltaxhorest}} \small{wind integration}},
legend style={font=\footnotesize},
legend columns=1,
legend cell align=left,
]
\pgfplotstableread{figs/ch10_ana/versus_integr/versus_integr_pos_hor_m_pc_part1.txt}\mytable
\addplot table [header=false, x index=0,y index=1] {\mytable};
\pgfplotstableread{figs/ch10_ana/versus_integr/versus_integr_pos_hor_m_pc_part2.txt}\mytable
\addplot table [header=false, x index=0,y index=1] {\mytable};
\addplot table [header=false, x index=4,y index=5] {\mytable};
\pgfplotstableread{figs/ch10_ana/versus_integr/versus_integr_pos_hor_m_pc_part1.txt}\mytable
\addplot table [header=false, x index=0,y index=2] {\mytable};
\addplot table [header=false, x index=0,y index=3] {\mytable};
\pgfplotstableread{figs/ch10_ana/versus_integr/versus_integr_pos_hor_m_pc_part2.txt}\mytable
\addplot table [header=false, x index=0,y index=2] {\mytable};
\addplot table [header=false, x index=0,y index=3] {\mytable};
\addplot table [header=false, x index=4,y index=6] {\mytable};
\addplot table [header=false, x index=4,y index=7] {\mytable};
\path node [draw, shape=rectangle, fill=white] at (50,0.05) {\footnotesize Scenario \nm{\#1}};
\end{axis}   
\end{tikzpicture}
\caption{Short term influence of filter algorithms on horizontal position NSE for scenario \nm{\#1} (100 runs)}
\label{fig:Ana_NSE_Results_xhor_integr_detail}
\end{figure}
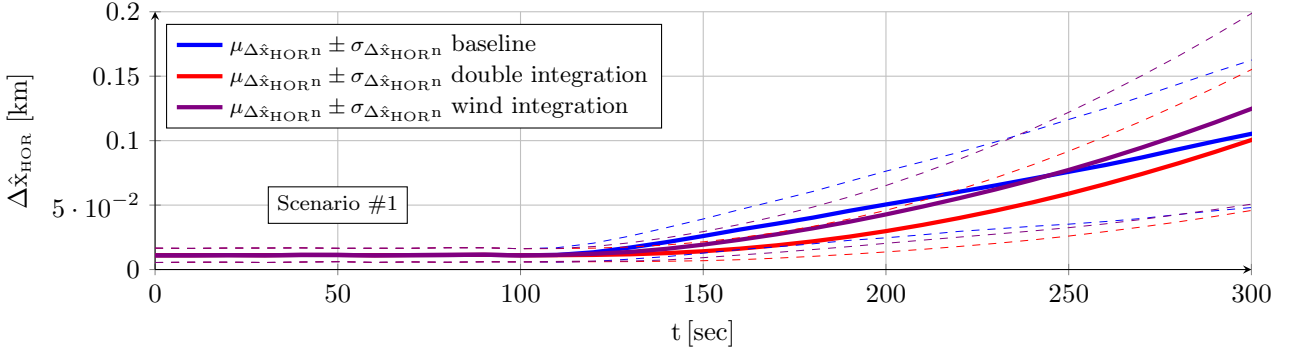

Figures \ref{fig:Ana_NSE_Results_xhor_integr} and \ref{fig:Ana_NSE_Results_xhor_integr_detail}\footnote{Note that the blue lines of figures \ref{fig:Ana_NSE_Results_xhor_integr} and \ref{fig:Ana_NSE_Results_xhor_integr_detail} are the same as those of figure \ref{fig:Sim_NSE_Results_xhor}.} show the results obtained with these two approaches for scenario \nm{\#1} when compared with those of the baseline filter. The figures for scenario \nm{\#2} are not included as they would show exactly the same trends. Figure \ref{fig:Ana_NSE_Results_xhor_integr} shows the whole duration of scenario \nm{\#1} (\nm{\tEND = 3800 \lrsb{sec}}), where the two alternative executions are limited to \nm{\tEND = 1000 \, \lrsb{sec}} as explained above, while figure \ref{fig:Ana_NSE_Results_xhor_integr_detail} provides more detail into the first few minutes of GNSS-Denied conditions. They clearly show that although the alternative approaches may prove advantageous for a short period of time after the GNSS signals are lost at \nm{\tGNSS = 100 \lrsb{sec}}, the higher order integration errors soon take over and make these alternatives far inferior to the proposed position filter.

The reason for this behavior originates in the additional integrations required to estimate the horizontal position. This analysis validates the position filter approach of freezing the wind speed to its estimated value at \nm{\tGNSS} (\ref{eq:filter_pos_vWINDN}). While any wind variation from that time on accumulates as ground velocity and horizontal position errors, these are bounded by atmospheric physics, which is preferred to the quick and unrestrained position error growth caused by alternative approaches that rely on double integrations.


\subsubsection*{Different Vertical Position Algorithms}

The position filter does not integrate the vertical speed to obtain the vertical position or geometric altitude, but instead estimates it through (\ref{eq:GNC_Navigation_AirDataFilterGNSSDenied_H}) and (\ref{eq:GNC_Navigation_AirDataFilterGNSSDenied_h}) from the pressure altitude \nm{\Hpest} and temperature offset \nm{\DeltaTest} estimated by the air data filter, together with the pressure offset \nm{\Deltapest} estimated at the time the GNSS signals are lost (\ref{eq:filter_pos_Deltap}). This approach leaves the vertical position estimation susceptible to atmospheric pressure changes, and results in an estimation error that is bounded by atmospheric physics, but proportional to the amount of pressure offset \nm{\Deltap} change from \nm{\tGNSS} to the time the vertical position is being estimated. The baseline approach is compared in this section with two alternative implementations that rely on integrating the estimated vertical speed. 
\begin{itemize}
\item The first, named \say{integration}, obtains the geometric altitude in the same way as the longitude and latitude, this is, by integrating the ground speed \nm{\vNest} obtained per (\ref{eq:filter_pos_vN}). This results in:
\neweq{\hest = {\hat h}_{n-1} + \DeltatEST \cdot \hdotest =  {\hat h}_{n-1} - \DeltatEST \cdot \vNestiii} {eq:Ana_PosFilter_h_integr_h}

\item The second alternative, named \say{airspeed integration}, imposes zero vertical wind and only integrates the vertical component of the airspeed. This approach takes into account the fact that the vertical wind is always very close to zero in the long term and only appears for short periods of time caused by gusts or when flying close to the terrain.
\begin{eqnarray}
\nm{\vNestiii} & \nm{\approx} & \nm{\vTASNestiii = \lrsb{\qNBest \otimes \vTASB\lrp{\vtasest, \, \alphaest, \, \betaest} \otimes \qNBastest}_{\sss 3}}\label{eq:Ana_PosFilter_h_tas_integr_vtas} \\
\nm{\hest} & = & \nm{{\hat h}_{n-1} + \DeltatEST \cdot \hdotest =  {\hat h}_{n-1} - \DeltatEST \cdot \vNestiii} \label {eq:Ana_PosFilter_h_tas_integr_h}
\end{eqnarray}
\end{itemize}

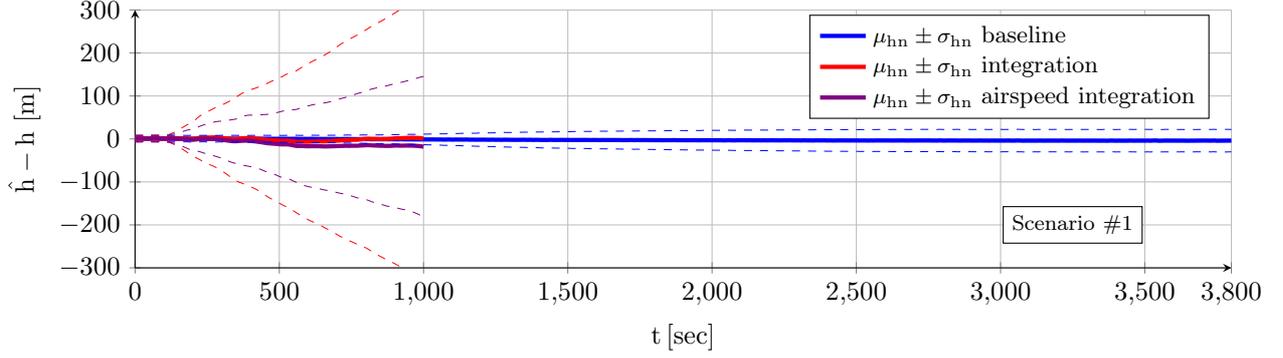
\begin{figure}[h]
\centering
\pgfplotsset{
	every axis legend/.append style={
		at={(0.98,0.78)},
		anchor=east,
	},
}
\begin{tikzpicture}
\begin{axis}[
cycle list={{blue,no markers,ultra thick},
			{red,no markers,ultra thick},
			{violet,no markers,ultra thick},
			{blue,dashed,no markers,ultra thin},{blue,dashed,no markers,ultra thin},
            {red,dashed,no markers,ultra thin}, {red,dashed,no markers,ultra thin},
			{violet,dashed,no markers,ultra thin}, {violet,dashed,no markers,ultra thin}},
width=16.0cm,
height=5.0cm,
xmin=0, xmax=3800, xtick={0,500,...,3500,3800},
xlabel={\nm{t \lrsb{sec}}},
xmajorgrids,
ymin=-300, ymax=300, ytick={-300,-200,...,300},
ylabel={\nm{\hest - h \, \lrsb{m}}},
ymajorgrids,
axis lines=left,
axis line style={-stealth},
legend entries={\nm{\mun{h} \pm \sigman{h}} \small{baseline},
                \nm{\mun{h} \pm \sigman{h}} \small{integration},
				\nm{\mun{h} \pm \sigman{h}} \small{airspeed integration}},
legend style={font=\footnotesize},
legend columns=1,
legend cell align=left,
]
\pgfplotstableread{figs/ch10_ana/versus_integr/versus_integr_pos_h_m_part1.txt}\mytable
\addplot table [header=false, x index=0,y index=1] {\mytable};
\pgfplotstableread{figs/ch10_ana/versus_integr/versus_integr_pos_h_m_part2.txt}\mytable
\addplot table [header=false, x index=0,y index=1] {\mytable};
\addplot table [header=false, x index=4,y index=5] {\mytable};
\pgfplotstableread{figs/ch10_ana/versus_integr/versus_integr_pos_h_m_part1.txt}\mytable
\addplot table [header=false, x index=0,y index=2] {\mytable};
\addplot table [header=false, x index=0,y index=3] {\mytable};
\pgfplotstableread{figs/ch10_ana/versus_integr/versus_integr_pos_h_m_part2.txt}\mytable
\addplot table [header=false, x index=0,y index=2] {\mytable};
\addplot table [header=false, x index=0,y index=3] {\mytable};
\addplot table [header=false, x index=4,y index=6] {\mytable};
\addplot table [header=false, x index=4,y index=7] {\mytable};
\path node [draw, shape=rectangle, fill=white] at (3250,-200) {\footnotesize Scenario \nm{\#1}};
\end{axis}   
\end{tikzpicture}
\caption{Influence of filter algorithms on vertical position NSE for scenario \nm{\#1} (100 runs)}
\label{fig:Ana_NSE_Results_h_integr}
\end{figure}

Figure \ref{fig:Ana_NSE_Results_h_integr}\footnote{Note that the blue lines of figures \ref{fig:Ana_NSE_Results_h_integr} are the same as those of figure \ref{fig:Sim_NSE_Results_Position_h}.} shows that although the second alternative is better than the first one, their performances are in both cases clearly inferior to that of the baseline. As in the previous section, the reason is that the unbounded integration of a noisy signal accumulates errors and needs to be avoided. In the case of the baseline, there is no integration, and the vertical position error depends on the ionospheric effects plus the variation in pressure offset since the GNSS signals are lost. Both are bounded by atmospheric physics and hence so is the resulting error. Note that the geometric altitude errors are so significant that on some executions they destabilize the filter, and hence it has been necessary to shorten the scenario \nm{\#1} simulations to \nm{\tEND = 1000 \, \lrsb{sec}}. The scenario \nm{\#2} figure is not included as it would show exactly the same trends. 

This analysis validates the approach of obtaining the vertical position or geometric altitude by means of the pressure altitude estimation by the air data filter while avoiding any integrations. Although atmospheric pressure changes then transform into vertical position errors, they are bounded by atmospheric physics and do not drift with time.


\subsubsection*{Different Attitude Algorithms}

The key characteristic of the proposed navigation filter is its ability to obtain a bounded and drift-less body attitude estimation in GNSS-Denied conditions while in turbulent flight, which is achieved by minimizing the negative influence of the ground velocity and position estimation errors. In particular, the attitude filter discards the airspeed and turbulence time derivatives viewed in the body frame (\nm{\vTASBdot = 0}, \nm{\vTURBBdot = 0}) and the wind field time derivative viewed in NED (\nm{\vWINDNdot = 0}), resulting in (\ref{eq:filter_att_obs_fIBB}). The discarded derivatives roughly oscillate around zero and hence their absence can be absorbed into the filter measurement noise, resulting in a filter that although noisier, maintains its unbiased nature and hence follows to a higher degree the EKF design assumptions \cite{Simon2006}.

This section analyzes two different alternatives. Although simpler to establish, both incur in biased observations of the specific force that result in higher body attitude estimation errors when executing maneuvers, as shown below. The first option, named \say{Zero \nm{\FB} Time Derivatives}, discards the full ground speed time derivative viewed in the body frame (\nm{\vTASBdot = 0}, \nm{\vTURBBdot = 0}, \nm{\vWINDBdot = 0}), resulting in:
\begin{eqnarray}
\nm{\fIBB} & \sim & \nm{\wEBBskew \; \lrp{\vTASB + \vTURBB + \vWINDB} + \qNBast \otimes \lrp{\acorN - \gcN} \otimes \qNB} \label{eq:Ana_AttFilter_fibb_theory_option2} \\
\nm{\fIBBtilde} & = & \nm{\wNBBskew \, \lrsb{\qNBast \otimes \vN \otimes \qNB} + \qNBast \otimes \lrp{\wENNskew \, \vN + \acorN - \gcNMODEL} \otimes \qNB + \EACC} \label{eq:Ana_AttFilter_fibb_option2}
\end{eqnarray}

The second option, named \say{Zero \nm{\FN} Time Derivatives}, also discards the full ground speed time derivative but this time viewed in NED (\nm{\vTASNdot = 0}, \nm{\vTURBNdot = 0}, \nm{\vWINDNdot = 0}), resulting in:
\begin{eqnarray}
\nm{\fIBB} & \sim & \nm{\qNBast \otimes \lrsb{\wENNskew \; \lrp{\vTASN + \vTURBN + \vWINDN}} \otimes \qNB + \qNBast \otimes \lrp{\acorN - \gcN} \otimes \qNB} \label{eq:Ana_AttFilter_fibb_theory_option3} \\
\nm{\fIBBtilde} & = & \nm{\qNBast \otimes \lrp{\wENNskew \, \vN + \acorN - \gcNMODEL} \otimes \qNB + \EACC} \label{eq:Ana_AttFilter_fibb_option3}
\end{eqnarray}

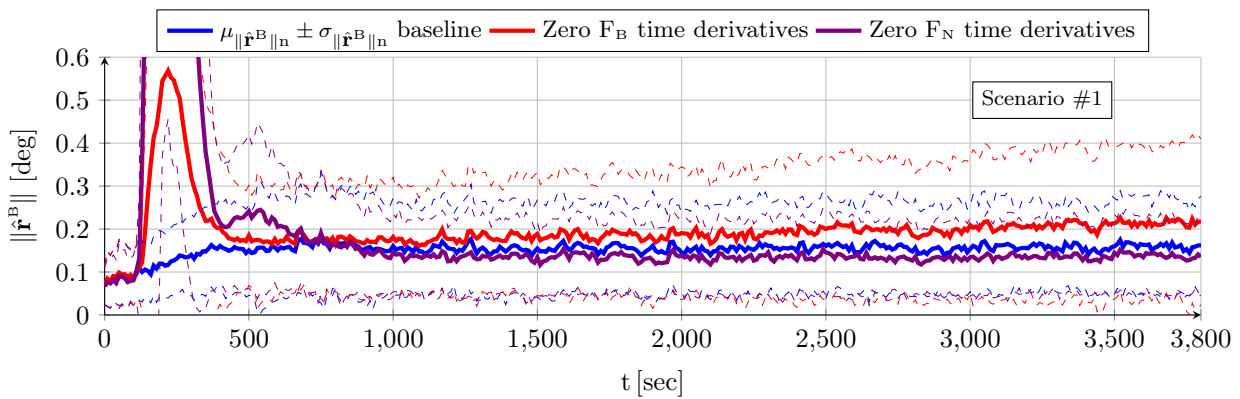
\begin{figure}[h]
\centering
\pgfplotsset{
	every axis legend/.append style={
		at={(0.5,1.02)},
		anchor=south,
	},
}
\begin{tikzpicture}
\begin{axis}[
cycle list={{blue,no markers,ultra thick},
			{red,no markers,ultra thick},
			{violet,no markers,ultra thick},
			{blue,dashed,no markers,ultra thin},{blue,dashed,no markers,ultra thin},
            {red,dashed,no markers,ultra thin}, {red,dashed,no markers,ultra thin}, 
            {violet,dashed,no markers,ultra thin}, {violet,dashed,no markers,ultra thin}},
width=16.0cm,
height=5.0cm,
xmin=0, xmax=3800, xtick={0,500,...,3500,3800},
xlabel={\nm{t \lrsb{sec}}},
xmajorgrids,
ymin=0, ymax=0.6, ytick={0,0.1,0.2,0.3,0.4,0.5,0.6},
ylabel={\nm{\DeltarBestnorm \, \lrsb{deg}}},
ymajorgrids,
axis lines=left,
axis line style={-stealth},
legend entries={\nm{\mun{\DeltarBestnorm} \pm \sigman{\DeltarBestnorm}} \small{baseline},
                \small{Zero \nm{\FB} time derivatives},
				\small{Zero \nm{\FN} time derivatives}},
legend columns=3,
legend style={font=\footnotesize},
legend cell align=left,
]
\pgfplotstableread{figs/ch10_ana/versus_att_filter/versus_att_filter_euler_deg.txt}\mytable
\addplot table [header=false, x index=0,y index=1] {\mytable};
\addplot table [header=false, x index=0,y index=4] {\mytable};
\addplot table [header=false, x index=0,y index=7] {\mytable};
\addplot table [header=false, x index=0,y index=2] {\mytable};
\addplot table [header=false, x index=0,y index=3] {\mytable};
\addplot table [header=false, x index=0,y index=5] {\mytable};
\addplot table [header=false, x index=0,y index=6] {\mytable};
\addplot table [header=false, x index=0,y index=8] {\mytable};
\addplot table [header=false, x index=0,y index=9] {\mytable};
\path node [draw, shape=rectangle, fill=white] at (3250,0.5) {\footnotesize Scenario \nm{\#1}};
\end{axis}
\end{tikzpicture}
\caption{Influence of filter algorithms on body attitude NSE for scenario \nm{\#1} (100 runs)}
\label{fig:Ana_NSE_Results_euler_att_filter}
\end{figure}

\begin{figure}[h]
\centering
\pgfplotsset{
	every axis legend/.append style={
		at={(0.5,1.02)},
		anchor=south,
	},
}
\begin{tikzpicture}
\begin{axis}[
cycle list={{blue,no markers,ultra thick}, 
            {green,no markers,thick},
            {red,no markers,ultra thick}, 
			{violet,no markers,ultra thick},
            {blue,dashed,no markers,ultra thin}, {blue,dashed,no markers,ultra thin},
            {red,dashed,no markers,ultra thin}, {red,dashed,no markers,ultra thin}, 
			{violet,dashed,no markers,ultra thin}, {violet,dashed,no markers,ultra thin}}, 
width=16.0cm,
height=5.0cm,
xmin=0, xmax=500, xtick={0,50,...,500},
xlabel={\nm{t \lrsb{sec}}},
xmajorgrids,
ymin=0, ymax=3.0, ytick={0,0.5,1,1.5,2,2.5,3},
ylabel={\nm{\DeltarBestnorm \, \lrsb{deg}}},
ymajorgrids,
axis lines=left,
axis line style={-stealth},
legend entries={\nm{\mun{\DeltarBestnorm} \pm \sigman{\DeltarBestnorm}} \small{baseline},
                \small{Number of executions turning / 10},
				\small{Zero \nm{\FB} time derivatives},
				\small{Zero \nm{\FN} time derivatives}},
legend columns=2,
legend style={font=\footnotesize},
legend cell align=left,
]
\pgfplotstableread{figs/ch10_ana/versus_att_filter/versus_att_filter_euler_deg.txt}\mytable
\addplot table [header=false, x index=0,y index=1] {\mytable};
\pgfplotstableread{figs/ch10_ana/versus_att_filter/versus_att_filter_turn_location.txt}\mytable
\addplot table [header=false, x index=0,y index=1] {\mytable};
\pgfplotstableread{figs/ch10_ana/versus_att_filter/versus_att_filter_euler_deg.txt}\mytable
\addplot table [header=false, x index=0,y index=4] {\mytable};
\addplot table [header=false, x index=0,y index=7] {\mytable};
\addplot table [header=false, x index=0,y index=2] {\mytable};
\addplot table [header=false, x index=0,y index=3] {\mytable};
\addplot table [header=false, x index=0,y index=5] {\mytable};
\addplot table [header=false, x index=0,y index=6] {\mytable};
\addplot table [header=false, x index=0,y index=8] {\mytable};
\addplot table [header=false, x index=0,y index=9] {\mytable};
\path node [draw, shape=rectangle, fill=white] at (50,2.5)  {\footnotesize Scenario \nm{\#1}};
\end{axis}
\end{tikzpicture}
\caption{Detail of influence of filter algorithms on body attitude NSE for scenario \nm{\#1} (100 runs)}
\label{fig:Ana_NSE_Results_euler_att_filter_detail}
\end{figure}
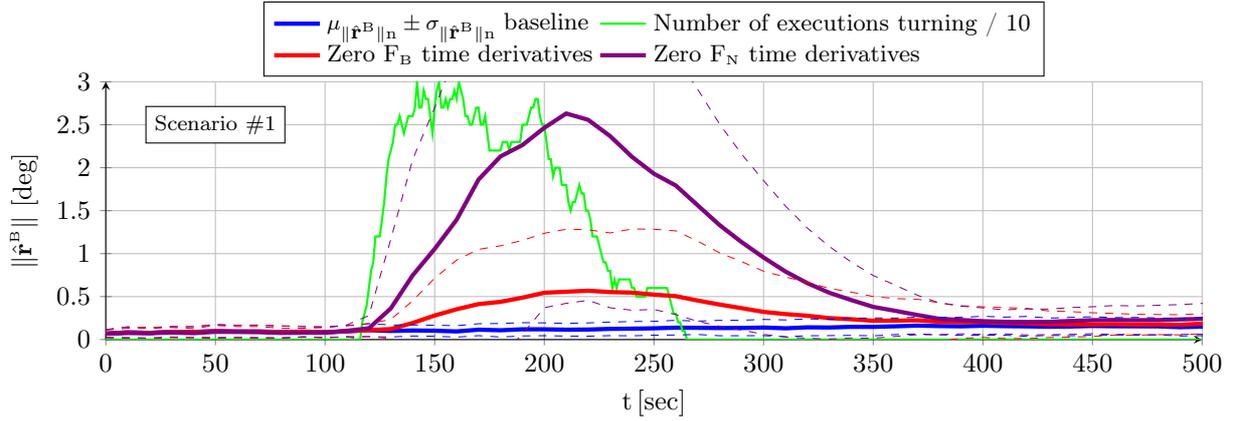

\begin{figure}[h]
\centering
\pgfplotsset{
	every axis legend/.append style={
		at={(0.5,1.02)},
		anchor=south,
	},
}
\begin{tikzpicture}
\begin{axis}[
cycle list={{blue,no markers,ultra thick}, 
            {green,no markers,thick},
            {red,no markers,ultra thick}, 
			{violet,no markers,ultra thick},
            {blue,dashed,no markers,ultra thin}, {blue,dashed,no markers,ultra thin},
            {red,dashed,no markers,ultra thin}, {red,dashed,no markers,ultra thin}, 
			{violet,dashed,no markers,ultra thin}, {violet,dashed,no markers,ultra thin}}, 
width=14.0cm,
height=5.0cm,
xmin=0, xmax=500, xtick={0,50,...,500},
xlabel={\nm{t \lrsb{sec}}},
xmajorgrids,
ymin=0, ymax=6, ytick={0,1,2,3,4,5,6},
ylabel={\nm{\DeltarBestnorm \, \lrsb{deg}}},
ymajorgrids,
axis lines=left,
axis line style={-stealth},
legend entries={\nm{\mun{\DeltarBestnorm} \pm \sigman{\DeltarBestnorm}} \small{baseline},
                \small{Number of executions turning / 10},
				\small{Zero \nm{\FB} time derivatives},
				\small{Zero \nm{\FN} time derivatives}},
legend columns=2,
legend style={font=\footnotesize},
legend cell align=left,
]
\pgfplotstableread{figs/ch10_ana/versus_att_filter_alter/versus_att_filter_alter_euler_deg.txt}\mytable
\addplot table [header=false, x index=0,y index=1] {\mytable};
\pgfplotstableread{figs/ch10_ana/versus_att_filter_alter/versus_att_filter_alter_turn_location.txt}\mytable
\addplot table [header=false, x index=0,y index=1] {\mytable};
\pgfplotstableread{figs/ch10_ana/versus_att_filter_alter/versus_att_filter_alter_euler_deg.txt}\mytable
\addplot table [header=false, x index=0,y index=4] {\mytable};
\addplot table [header=false, x index=0,y index=7] {\mytable};
\addplot table [header=false, x index=0,y index=2] {\mytable};
\addplot table [header=false, x index=0,y index=3] {\mytable};
\addplot table [header=false, x index=0,y index=5] {\mytable};
\addplot table [header=false, x index=0,y index=6] {\mytable};
\addplot table [header=false, x index=0,y index=8] {\mytable};
\addplot table [header=false, x index=0,y index=9] {\mytable};
\path node [draw, shape=rectangle, fill=white] at (75,5) {\footnotesize Scenario \nm{\#2}};
\end{axis}
\end{tikzpicture}
\caption{Influence of filter algorithms on body attitude NSE for scenario \nm{\#2} (100 runs)}
\label{fig:Ana_NSE_Results_euler_att_filter_alter}
\end{figure}
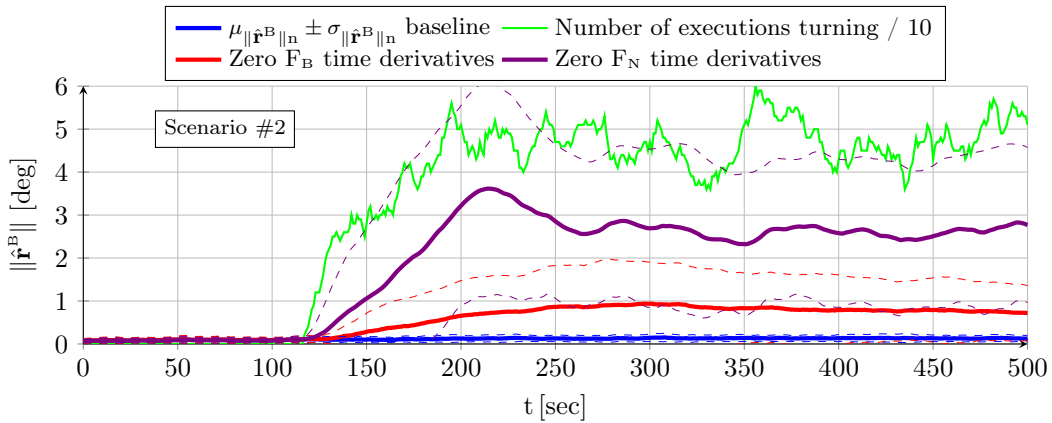

The aggregated body attitude NSE for the baseline configuration as well as the two alternatives are shown in table \ref{tab:inflence_attitude_errors}\footnote{Note that the baseline columns of table \ref{tab:inflence_attitude_errors} coincide with the right hand side of table \ref{tab:attitude_errors}.}, while their variations with time are depicted in figures \ref{fig:Ana_NSE_Results_euler_att_filter}, \ref{fig:Ana_NSE_Results_euler_att_filter_detail}\footnote{Note that figure \ref{fig:Ana_NSE_Results_euler_att_filter_detail} represents an augmented view of the first \nm{500 \lrsb{sec}} of figure \ref{fig:Ana_NSE_Results_euler_att_filter}.}, and \ref{fig:Ana_NSE_Results_euler_att_filter_alter}\footnote{Note that blue lines of figures \ref{fig:Ana_NSE_Results_euler_att_filter} and \ref{fig:Ana_NSE_Results_euler_att_filter_detail} coincide with those of figure \ref{fig:Sim_NSE_Results_euler}, and the blue lines of figure \ref{fig:Ana_NSE_Results_euler_att_filter_alter} coincide with those of figure \ref{fig:Sim_NSE_Results_euler_alter}.}.
\begin{center}
\begin{tabular}{llrrrrrrrrr}
\hline
\multicolumn{2}{l}{Scenario} & \multicolumn{3}{c}{Baseline} & \multicolumn{3}{c}{Zero \nm{\FB} Diffs.} & \multicolumn{3}{c}{Zero \nm{\FN} Diffs.} \\
\multicolumn{2}{c}{\nm{\DeltarBestnorm \ \lrsb{deg}}} & mean & std & max & mean & std & max & mean & std & max \\
\hline
\multirow{3}{*}{\nm{\#1}} & mean  & \textbf{0.152} &0.078 & 0.521 & \textbf{0.199} & 0.137 & 1.116 & \textbf{0.223} & 0.462 & 4.501 \\
                          & std   & 0.072 & 0.021 & 0.151 & 0.126 & 0.104 & 0.971 & 0.084 & 0.264 & 2.374 \\
                          & max   & 0.448 & 0.168 & 1.321 & 0.742 & 0.588 & 5.152 & 0.487 & 1.107 & 9.819 \\
\hline
\multirow{3}{*}{\nm{\#2}} & mean  & \textbf{0.118} &0.074 & 0.38 & \textbf{0.565} & 0.508 & 1.77 & \textbf{1.902} & 1.934 & 6.88 \\
                          & std   & 0.023 & 0.015 & 0.07 & 0.405 & 0.404 & 1.28 & 0.407 & 0.317 & 1.11 \\
                          & max   & 0.206 & 0.124 & 0.63 & 1.860 & 1.943 & 5.45 & 3.161 & 2.860 & 9.82 \\
\hline
\end{tabular}
\end{center} 
\captionof{table}{Influence of filter algorithms on aggregated body attitude NSE (100 runs)} \label{tab:inflence_attitude_errors}

If the beginning of figure \ref{fig:Ana_NSE_Results_euler_att_filter} is excluded, focusing exclusively on that portion of scenario \nm{\#1} in which the bearing is constant, the two alternatives result in body attitude estimation errors that are quite similar to those of the baseline configuration. As a matter of fact, the \say{Zero \nm{\FN} Time Derivatives} configuration performs consistently better than the baseline, while the \say{Zero \nm{\FB} Time Derivatives} configuration results are worse and also present a slight drift with time that however is slow enough so as to not represent any problem from a flight stability point of view.

However, the two alternative configurations incur in significant errors in the turn or change of bearing present in scenario \nm{\#1} (refer to \cite{SIMULATION} for details, or section \ref{sec:Simulation} for a summary of the scenario), which starts at \nm{\tTURN} and lasts until the final bearing \nm{\chiEND} is reached. To better visualize the aircraft turns, figure \ref{fig:Ana_NSE_Results_euler_att_filter_detail}, in addition to focusing on the first \nm{500 \lrsb{sec}} of the scenario \nm{\#1}, adds an extra line (green) that contains the number of runs at every time instant in which the aircraft is executing the turning maneuver\footnote{For example, a value of 2.8 at time \nm{t_s} means that twenty-eight out of the hundred runs are executing its turn at time \nm{t_s}, while seventy-two either have not started it or have already concluded the turning maneuver.}.

Note that the body attitude error for the two alternatives does not only increase quickly when the aircraft is turning, but continues doing so for at least a minute after the turn concludes, and then takes several minutes for it to return to levels comparable to what they were before initiating the turning maneuver. A close examination of the data reveals that the maximum error is directly related to the duration of the turn as the errors never stop increasing until after the aircraft has concluded its turn. Also, different executions in which turns are performed at bank angles different than \nm{\xiTURN \pm 10 \lrsb{deg}} result in errors that are proportional not only to the turn duration (or change of bearing) but also to the bank angle employed.

This behavior is emphasized in scenario \nm{\#2}, which is based on a series of consecutive turns with very little time in between (refer to \cite{SIMULATION} for details, or section \ref{sec:Simulation} for a summary of the scenario). Figure \ref{fig:Ana_NSE_Results_euler_att_filter_alter} clearly shows the benefits of the baseline configuration, as the estimation error in the two alternatives accumulates because it does not have time after a given turn to return to its pre-turn levels before the next turning maneuver is initiated.

The reason why the performance of the three configurations is similar when not executing any maneuvers but markedly different when turning lies in the nature of the airspeed, turbulence, and wind field derivatives with time. When in stationary straight flight, all three roughly oscillate around zero when viewed in any frame and the effect of including or removing them from (\ref{eq:filter_att_obs_fIBB}) is slight; this is not the case when turning, in which case the airspeed and turbulence are approximately constant only when viewed in body, while the wind field needs to be viewed in NED. Although it is not possible to fully observe these derivatives in GNSS-Denied conditions, the baseline configuration manages to avoid introducing significant biases into the filter at the expense of increasing its white noise, while the two alternative do add biases, resulting in the filter producing incorrect body attitude estimations to compensate them.

The proposed attitude filter observation algorithms result in a better estimation performance when faced with the lack of observability inherent to GNSS-Denied navigation. By neglecting the different speeds (airspeed, wind, turbulence) derivatives with time when viewed in certain reference frames, it is possible to maintain the unbiased nature of the filter at the expense of increasing its noise.


\section{Influence of Sensor Quality on GNSS-Denied Navigation} \label{sec:Influence}

This section analyzes the influence of the quality of the different aircraft sensors on the performance of the aircraft navigation system in GNSS-Denied conditions.  It discusses the influence of the bias drift oscillation band for the inertial sensors, as well as that of the performance parameters of gyroscopes, accelerometers, magnetometers, air speed sensors, and atmospheric sensors.

The analysis relies on repeating the Monte Carlo simulations for both scenarios composed of one hundred runs each with the only change of replacing the performance parameters shown in \cite{SENSORS} and section \ref{sec:Simulation}, referred to as the \say{baseline} configuration, with others representing higher or lower quality sensors, and then comparing the NSE obtained by the navigation system. 

The main objective of this analysis is to determine the influence that each of the sensors has on the aircraft GNSS-Denied navigation capabilities. By ensuring that the result obtained when employing sensors of inferior quality are qualitatively the same and quantitatively slightly inferior to those obtained with the baseline configuration, the simulation results are safeguarded from possible errors introduced in \cite{SENSORS} when modeling the performances of the different sensors.


\subsubsection*{Influence of Inertial Sensors Bias Stability}

In the description of the different error sources present in the output of the inertial sensors (accelerometers and gyroscopes) contained in \cite{SENSORS}, the bias drift is modeled as a random walk process obtained by the integration of a white noise signal. However, as the bias drift is mostly a warm-up process that stabilizes with the sensor temperature, the random walk is not allowed to vary freely (which would not be realistic) but is constrained to a band of \nm{\pm \, 100 \cdot \sigma_u \cdot \Deltat^{1/2}}, this is, it is only allowed to accumulate a hundred times in the same direction. 
\begin{center}
\begin{tabular}{clrrrp{0.1cm}rrrp{0.1cm}rrr}
\hline
\multicolumn{2}{l}{Scenario} & \multicolumn{3}{c}{Baseline} & & \multicolumn{3}{c}{\nm{\pm \, 300 \cdot \sigma_u  \cdot \Deltat^{1/2}}} & & \multicolumn{3}{c}{\nm{\pm \, 1000 \cdot \sigma_u  \cdot \Deltat^{1/2}}} \\
\multicolumn{2}{c}{\nm{\DeltarBestnorm \ \lrsb{deg}}} & mean & std & max & & mean & std & max & & mean & std & max \\
\hline
\multirow{3}{*}{\nm{\#1}} & mean & \textbf{0.152} &0.078 & 0.521 & & \textbf{0.164} & 0.089 & 0.567 & & \textbf{0.175} & 0.096 & 0.590 \\
                          & std  & 0.072 & 0.021 & 0.151 & & 0.070 & 0.023 & 0.151 & & 0.074 & 0.024 & 0.154 \\
                          & max  & 0.448 & 0.168 & 1.321 & & 0.451 & 0.173 & 1.339 & & 0.469 & 0.178 & 1.399 \\
\hline
\multirow{3}{*}{\nm{\#2}} & mean & \textbf{0.118} &0.074 & 0.376 & & \textbf{0.125} & 0.077 & 0.394 & & \textbf{0.128} & 0.079 & 0.399 \\
                          & std  & 0.023 & 0.015 & 0.069 & & 0.027 & 0.016 & 0.080 & & 0.030 & 0.017 & 0.081 \\
                          & max  & 0.206 & 0.124 & 0.632 & & 0.234 & 0.122 & 0.661 & & 0.239 & 0.144 & 0.672 \\
\hline
\end{tabular}
\end{center}
\captionof{table}{Influence of inertial biases oscillation band on aggregated body attitude NSE (100 runs)} \label{tab:Ana_NSE_Results_euler_bias}

Although limiting the random walk oscillation is the proper way to model its influence, the \nm{\pm \, 100} value is arbitrary and if optimistic may facilitate the tracking of the gyroscope and accelerometer errors (\nm{\EGYR} and \nm{\EACC}) by the navigation filter, limiting the negative consequences of failing to appropriately estimate these errors. For that reason, this analysis relaxes this limitation in two different configurations, to \nm{\pm \, 300} and \nm{\pm \, 1000} (simultaneously for the gyroscopes and accelerometers), and evaluates the sensitivity of the results to this change.

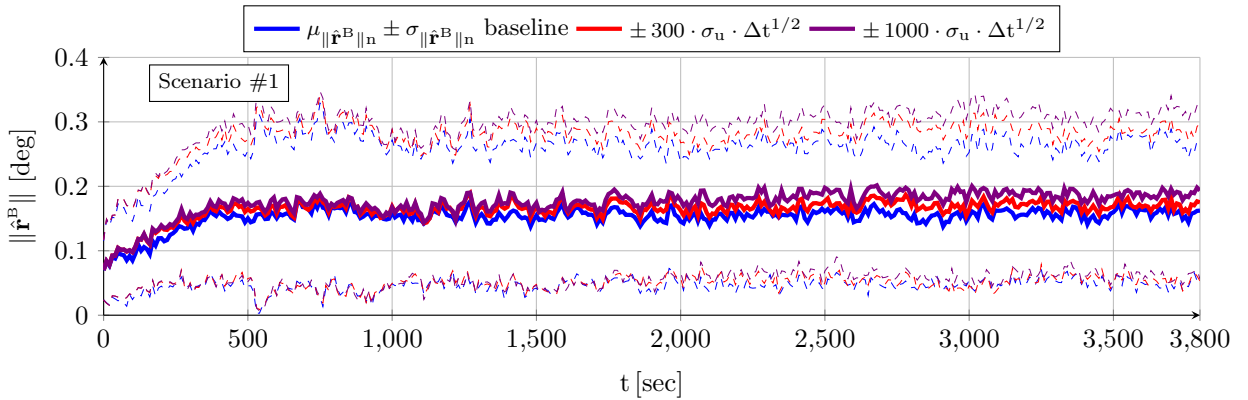
\begin{figure}[h]
\centering
\pgfplotsset{
	every axis legend/.append style={
		at={(0.5,1.02)},
		anchor=south,
	},
}
\begin{tikzpicture}
\begin{axis}[
cycle list={{blue,no markers,ultra thick},
			{red,no markers,ultra thick},
			{violet,no markers,ultra thick},
			{blue,dashed,no markers,ultra thin},{blue,dashed,no markers,ultra thin},
            {red,dashed,no markers,ultra thin}, {red,dashed,no markers,ultra thin}, 
			{violet,dashed,no markers,ultra thin}, {violet,dashed,no markers,ultra thin}}, 
width=16.0cm,
height=5.0cm,
xmin=0, xmax=3800, xtick={0,500,...,3500,3800},
xlabel={\nm{t \lrsb{sec}}},
xmajorgrids,
ymin=0, ymax=0.4, ytick={0,0.1,0.2,0.3,0.4},
ylabel={\nm{\DeltarBestnorm \, \lrsb{deg}}},
ymajorgrids,
axis lines=left,
axis line style={-stealth},
legend entries={\nm{\mun{\DeltarBestnorm} \pm \sigman{\DeltarBestnorm}} \small{baseline},
				\nm{\pm \, 300 \cdot \sigma_u \cdot \Deltat^{1/2}},
				\nm{\pm \, 1000 \cdot \sigma_u \cdot \Deltat^{1/2}}},
legend columns=3,
legend style={font=\footnotesize},
legend cell align=left,
]
\pgfplotstableread{figs/ch10_ana/versus_bias/versus_bias_euler_deg.txt}\mytable
\addplot table [header=false, x index=0,y index=1] {\mytable};
\addplot table [header=false, x index=0,y index=4] {\mytable};
\addplot table [header=false, x index=0,y index=7] {\mytable};
\addplot table [header=false, x index=0,y index=2] {\mytable};
\addplot table [header=false, x index=0,y index=3] {\mytable};
\addplot table [header=false, x index=0,y index=5] {\mytable};
\addplot table [header=false, x index=0,y index=6] {\mytable};
\addplot table [header=false, x index=0,y index=8] {\mytable};
\addplot table [header=false, x index=0,y index=9] {\mytable};

\path node [draw, shape=rectangle, fill=white] at (400,0.36) {\footnotesize Scenario \nm{\#1}};
\end{axis}   
\end{tikzpicture}
\caption{Influence of inertial biases oscillation band on body attitude NSE for scenario \nm{\#1} (100 runs)}
\label{fig:Ana_NSE_Results_euler_bias}
\end{figure}

\begin{figure}[h]
\centering
\pgfplotsset{
	every axis legend/.append style={
		at={(0.5,1.02)},
		anchor=south,
	},
}
\begin{tikzpicture}
\begin{axis}[
cycle list={{blue,no markers,ultra thick},
			{red,no markers,ultra thick},
			{violet,no markers,ultra thick},
			{blue,dashed,no markers,ultra thin},{blue,dashed,no markers,ultra thin},
            {red,dashed,no markers,ultra thin}, {red,dashed,no markers,ultra thin}, 
			{violet,dashed,no markers,ultra thin}, {violet,dashed,no markers,ultra thin}}, 
width=14.0cm,
height=5.0cm,
xmin=0, xmax=500, xtick={0,50,...,500},
xlabel={\nm{t \lrsb{sec}}},
xmajorgrids,
ymin=0, ymax=0.3, ytick={0,0.05,0.1,0.15,0.2,0.25,0.3},
ylabel={\nm{\DeltarBestnorm \, \lrsb{deg}}},
ymajorgrids,
axis lines=left,
axis line style={-stealth},
legend entries={\nm{\mun{\DeltarBestnorm} \pm \sigman{\DeltarBestnorm}} \small{baseline},
				\nm{\pm \, 300 \cdot \sigma_u \cdot \Deltat^{1/2}},
				\nm{\pm \, 1000 \cdot \sigma_u \cdot \Deltat^{1/2}}},
legend columns=3,
legend style={font=\footnotesize},
legend cell align=left,
]
\pgfplotstableread{figs/ch10_ana/versus_bias/versus_bias_alter_euler_deg.txt}\mytable
\addplot table [header=false, x index=0,y index=1] {\mytable};
\addplot table [header=false, x index=0,y index=4] {\mytable};
\addplot table [header=false, x index=0,y index=7] {\mytable};
\addplot table [header=false, x index=0,y index=2] {\mytable};
\addplot table [header=false, x index=0,y index=3] {\mytable};
\addplot table [header=false, x index=0,y index=5] {\mytable};
\addplot table [header=false, x index=0,y index=6] {\mytable};
\addplot table [header=false, x index=0,y index=8] {\mytable};
\addplot table [header=false, x index=0,y index=9] {\mytable};
\path node [draw, shape=rectangle, fill=white] at (75,0.25) {\footnotesize Scenario \nm{\#2}};
\end{axis}   
\end{tikzpicture}
\caption{Influence of inertial biases oscillation band on body attitude NSE for scenario \nm{\#2} (100 runs)}
\label{fig:Ana_NSE_Results_euler_bias_alter}
\end{figure}
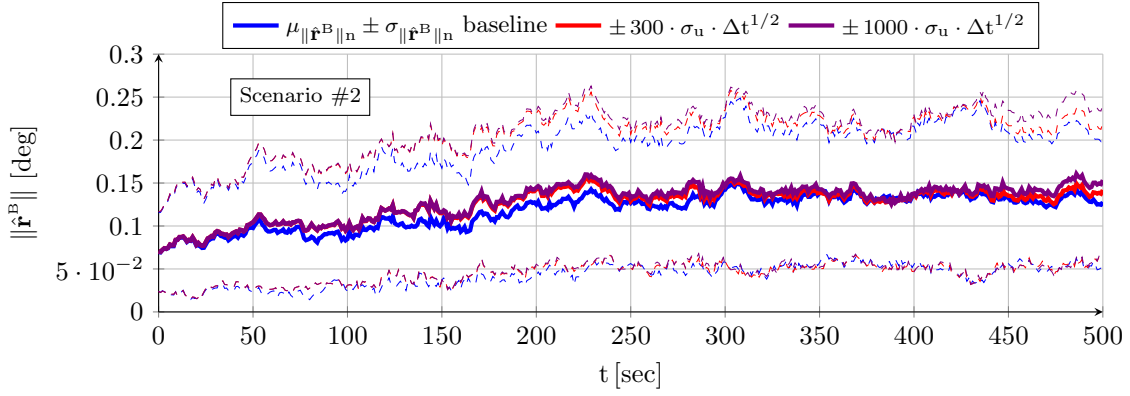

Relaxing the width of the band in which the inertial sensors bias drift can oscillate results in a quantitatively inferior estimation of the body attitude, as shown in table \ref{tab:Ana_NSE_Results_euler_bias}, together with figures \ref{fig:Ana_NSE_Results_euler_bias} and \ref{fig:Ana_NSE_Results_euler_bias_alter}\footnote{Note that the baseline columns of table \ref{tab:Ana_NSE_Results_euler_bias} coincide with the right hand columns of table \ref{tab:attitude_errors}, and the blue lines of figures \ref{fig:Ana_NSE_Results_euler_bias} and \ref{fig:Ana_NSE_Results_euler_bias_alter} coincide with those of figures \ref{fig:Sim_NSE_Results_euler} and \ref{fig:Sim_NSE_Results_euler_alter}, respectively.}, although qualitatively the estimation remains bounded and drift-less.

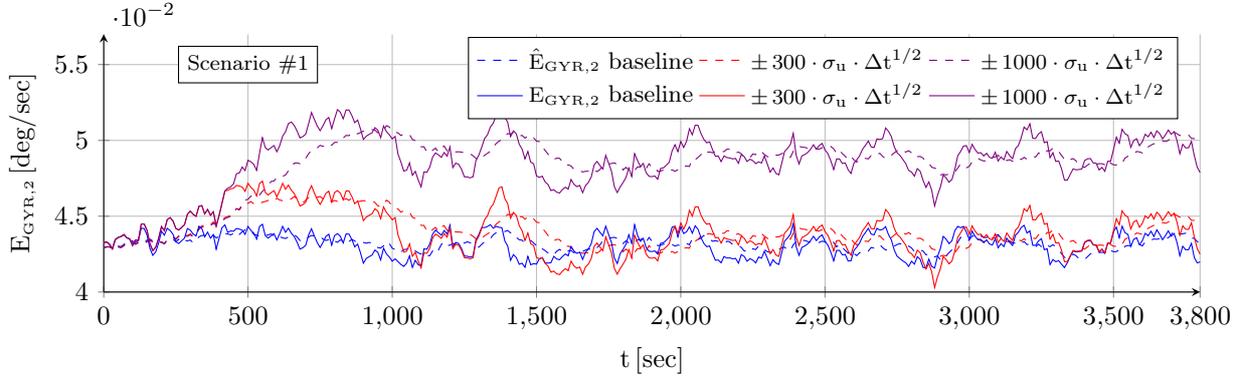
\begin{figure}[h]
\centering
\pgfplotsset{
	every axis legend/.append style={
		at={(0.98,0.83)},
		anchor=east,
	},
}
\begin{tikzpicture}
\begin{axis}[
cycle list={{blue,dashed,no markers},{red,dashed,no markers},{violet,dashed,no markers},
            {blue,no markers},{red,no markers},{violet,no markers}},
width=16.0cm,
height=5.0cm,
xmin=0, xmax=3800, xtick={0,500,...,3500,3800},
xlabel={\nm{t \lrsb{sec}}},
xmajorgrids,
ymin=0.04, ymax=0.057, ytick={0.04,0.045,0.05,0.055},
ylabel={\nm{\EGYRii \lrsb{deg/sec}}},
ymajorgrids,
axis lines=left,
axis line style={-stealth},
legend entries={estimated, truth},
legend entries={\nm{\EGYRestii} \small{baseline}, \nm{\pm \, 300 \cdot \sigma_u \cdot \Deltat^{1/2}}, \nm{\pm \, 1000 \cdot \sigma_u \cdot \Deltat^{1/2}},
                \nm{\EGYRii} \small{baseline},    \nm{\pm \, 300 \cdot \sigma_u \cdot \Deltat^{1/2}}, \nm{\pm \, 1000 \cdot \sigma_u \cdot \Deltat^{1/2}}},
legend columns=3,
legend style={font=\footnotesize},
legend cell align=left,
]
\pgfplotstableread{figs/ch10_ana/versus_bias/versus_bias_single_Egyr_dps.txt}\mytable
\addplot table [header=false, x index=0,y index=1] {\mytable};
\addplot table [header=false, x index=0,y index=3] {\mytable};
\addplot table [header=false, x index=0,y index=5] {\mytable};
\addplot table [header=false, x index=0,y index=2] {\mytable};
\addplot table [header=false, x index=0,y index=4] {\mytable};
\addplot table [header=false, x index=0,y index=6] {\mytable};
\path node [draw, shape=rectangle, fill=white] at (500,0.055) {\footnotesize Scenario \nm{\#1}};
\end{axis}   
\end{tikzpicture}
\caption{Influence of inertial biases oscillation band on full gyroscope error for run \nm{\#10} of scenario \nm{\#1}}
\label{fig:Ana_NSE_Results_single_Egyr_bias}
\end{figure}

The capability of the attitude filter to estimate the body attitude is based on its ability to track the full gyroscope error \nm{\EGYR}. Figure \ref{fig:Ana_NSE_Results_single_Egyr_bias} shows how the filter tracks \nm{\EGYR} with a slight delay that is a compromise between allocating the changes observed in the measured inertial angular velocity \nm{\wIBBtilde} to the aircraft angular velocity \nm{\wNBBest} or the full gyroscope error \nm{\EGYRest}. This balance is determined by the selection of the attitude filter covariances, and is optimized for the baseline configuration. It is possible to adjust the covariances so the filter tracks the gyroscope errors more closely (with less delay), but this also implies significantly higher errors when maneuvering in which a higher percentage of the real \nm{\wNBB} would be allocated to the gyroscope errors \nm{\EGYRest}, as on the short term the filter can not distinguish between real angular velocities and gyroscope errors. 

If the sensors bias drift oscillates within a wider band, the tracking delay at times may translate into bigger \nm{\EGYRest} errors, in particular when the real error \nm{\EGYR} accumulates a significant change over a relatively small period of time. This can be better understood by looking at a specific case, such as in figure \ref{fig:Ana_NSE_Results_single_Egyr_bias}, which shows the \nm{\second} component of the gyroscope error \nm{\EGYRii} and its estimation \nm{\hat{E}_{\sss GYR,2}} resulting from run \nm{\#10} of scenario \nm{\#1}, for the three configurations. Note that between approximately \nm{400} and \nm{800 \lrsb{sec}}, the random walk quickly accumulates creating ever higher bias drift values, and this is precisely where the tracking delay causes higher \nm{\EGYRest} errors, and hence bigger \nm{\DeltarBestnorm} errors. 
\begin{center} 
\begin{tabular}{clrrp{0.1cm}rrp{0.1cm}rr}
\hline
\multicolumn{2}{l}{Scenario} & \multicolumn{2}{c}{Baseline} & & \multicolumn{2}{c}{\nm{\pm \,300 \cdot \sigma_u \cdot \Deltat^{1/2}}} & & \multicolumn{2}{c}{\nm{\pm \, 1000 \cdot \sigma_u \cdot \Deltat^{1/2}}} \\
\multicolumn{2}{l}{\nm{\vNest \lrp{\tEND} - \vN \lrp{\tEND} \lrsb{m/sec}}} & \nm{\vNesti - \vNi} & \nm{\vNestii - \vNii} & & \nm{\vNesti - \vNi} & \nm{\vNestii - \vNii} & & \nm{\vNesti - \vNi} & \nm{\vNestii - \vNii} \\
\hline
\multirow{3}{*}{\nm{\#1}} & mean & +0.325 & +0.465 & & +0.325 & +0.465 & & +0.325 & +0.460 \\
                          & std  & \textbf{2.588} & \textbf{2.204} & & \textbf{2.592} & \textbf{2.203} & & \textbf{2.593} & \textbf{2.204} \\
                          & max  & +9.321 & +7.392 & & +9.336 & +7.368 & & +9.293 & +7.436\\
\hline
\multirow{3}{*}{\nm{\#2}} & mean & -0.083 & -0.139 & & -0.084 & -0.140 & & -0.085 & -0.143 \\
                          & std  & \textbf{0.652} & \textbf{0.629} & & \textbf{0.656} & \textbf{0.634} & & \textbf{0.655} & \textbf{0.634} \\
                          & max  & -1.916 & -1.792 & & -1.906 & -1.791 & & -1.897 & -1.792\\
\hline
\end{tabular}
\end{center}
\captionof{table}{Influence of inertial biases oscillation band on aggregated final ground speed NSE (100 runs)} \label{tab:Ana_NSE_Results_vned_bias}

\begin{center}
\begin{tabular}{clrrrp{0.1cm}rrp{0.1cm}rr}
\hline
\multicolumn{2}{l}{Scenario} & \multicolumn{3}{c}{Baseline} & & \multicolumn{2}{c}{\nm{\pm \, 300 \cdot \sigma_u \cdot \Deltat^{1/2}}} & & \multicolumn{2}{c}{\nm{\pm \, 1000 \cdot \sigma_u \cdot \Deltat^{1/2}}} \\
\multicolumn{2}{l}{Error} & Distance \nm{\lrsb{m}} & \multicolumn{2}{c}{\nm{\Deltaxhorest \lrp{\tEND} \lrsb{m,\%}}} & & \multicolumn{2}{c}{\nm{\Deltaxhorest \lrp{\tEND}\lrsb{m,\%}}} & & \multicolumn{2}{c}{\nm{\Deltaxhorest \lrp{\tEND}\lrsb{m,\%}}} \\
\hline
\multirow{3}{*}{\nm{\#1}} & mean & 108622.6 & 7427.7 & \textbf{7.18} & & 7423.5 & \textbf{7.18} & & 7422.6 & \textbf{7.17} \\
                          & std  & 19935.3 & 4987.8 & 5.73 & & 4997.0 & 5.73 & & 5004.4 & 5.74 \\
                          & max  & 172842.3 & 25287.9 & 32.38 & & 25291.9 & 32.32 & & 25393.6 & 32.24 \\
\hline
\multirow{3}{*}{\nm{\#2}} & mean & 14198.4 & 215.6 & \textbf{1.52} & & 217.0 & \textbf{1.53} & & 217.1 & \textbf{1.53} \\
                          & std  & 1176.1 & 119.4 & 0.86 & & 120.2 & 0.87 & & 120.2 & 0.87 \\
                          & max  & 18253.0 & 586.3 & 4.38 & & 586.2 & 4.38 & & 586.2 & 4.38 \\
\hline
\end{tabular}
\end{center}
\captionof{table}{Influence of inertial biases oscillation band on aggregated final horizontal position NSE (100 runs)} \label{tab:Ana_NSE_Results_xhor_bias}

As shown in tables \ref{tab:Ana_NSE_Results_vned_bias} and \ref{tab:Ana_NSE_Results_xhor_bias}\footnote{Note that the baseline columns of tables \ref{tab:Ana_NSE_Results_vned_bias} and \ref{tab:Ana_NSE_Results_xhor_bias} coincide with those of tables \ref{tab:v_errors} and \ref{tab:hor_errors}.}, as long as the body attitude estimation remains bounded and drift-less, the influence of the attitude estimation accuracy on the horizontal velocity and position estimations is rather small. Note that the ground velocity and horizontal position results in the three configurations are identical for all practical purposes both for scenario \nm{\#1}, which is mostly stationary but contains varying wind, as for \nm{\#2}, which contains continuous maneuvers with a constant wind field. Although no data is shown, the influence on the vertical position estimation accuracy is also negligible.

The summary of this section is that the dependency of the results on the width of the band in which the inertial sensors bias drift oscillates is very small. The body attitude estimation results vary slightly but these changes are not significant enough to imply any changes in the horizontal velocity or position estimation accuracy.


\subsubsection*{Influence of Gyroscope Quality}

This section analyzes the sensitivity on the GNSS-Denied navigation results of employing gyroscopes of different quality or performance grade than those of the baseline described in \cite{SENSORS} and table \ref{tab:Sensors_gyr_acc}. Table \ref{tab:Ana_NSE_Results_gyr_other} shows the performances of the configuration named \say{GYR Better} in the same format as that employed in section \ref{sec:Simulation}, which correspond to the high performance MEMS gyroscopes inside the Sensonor STIM300 IMU \cite{STIM300}, and are better than those of the baseline. It also shows the performance parameters of two fictitious gyroscopes that perform worse than the baseline. The configurations are named \say{GYR Worse} and \say{GYR Worst}, and their objective is to provide a wide range of gyroscope performances with which to evaluate the sensitivity of the navigation algorithms to the gyroscopes grade. Additionally, they represent a safety measure in case of errors in the modeling of the baseline gyroscopes performances.
\begin{center}
\begin{tabular}{lc|lclclc}
	\hline
	& & \multicolumn{2}{c}{GYR Better} & \multicolumn{2}{c}{GYR Worse} & \multicolumn{2}{c}{GYR Worst} \\
	Error Source & Variable & \multicolumn{1}{c}{Value} & Unit & \multicolumn{1}{c}{Value} & Unit & \multicolumn{1}{c}{Value} & Unit \\
	\hline	
	Bias Drift					& \nm{\sigmauGYR}	& \nm{1.38 \cdot 10^{-5}}	& [\nm{deg/sec^{1.5}}] & \nm{5.00 \cdot 10^{-4}}	& [\nm{deg/sec^{1.5}}] & \nm{1.50 \cdot 10^{-3}}	& [\nm{deg/sec^{1.5}}] \\
	System Noise				& \nm{\sigmavGYR}	& \nm{2.50 \cdot 10^{-3}}	& [\nm{deg/sec^{0.5}}] & \nm{8.00 \cdot 10^{-3}}	& [\nm{deg/sec^{0.5}}] & \nm{2.50 \cdot 10^{-2}}	& [\nm{deg/sec^{0.5}}] \\
	Scale Factor				& \nm{\sGYR}		& \nm{5.00 \cdot 10^{-6}}	& [-] & \nm{5.00 \cdot 10^{-5}}	& [-]                  & \nm{1.00 \cdot 10^{-4}}	& [-] \\
	Cross Coupling				& \nm{\mGYR}		& \nm{1.50 \cdot 10^{-5}}	& [-] & \nm{1.50 \cdot 10^{-4}}	& [-]                  & \nm{4.50 \cdot 10^{-4}}	& [-] \\
	Bias Offset					& \nm{\BzeroGYR}	& \nm{3.00 \cdot 10^{-2}}	& [\nm{deg/sec}] & \nm{7.50 \cdot 10^{-1}}	& [\nm{deg/sec}]       & \nm{1.50 \cdot 10^{0}}	& [\nm{deg/sec}] \\
	\hline
\end{tabular}
\captionof{table}{Performances of ``GYR Better'', ``GYR Worse'', and ``GYR Worst'' gyroscopes} \label{tab:Ana_NSE_Results_gyr_other}
\end{center}

\begin{figure}[h]
\centering
\pgfplotsset{
	every axis legend/.append style={
		at={(0.5,1.07)},
		anchor=south,
	},
}
\begin{tikzpicture}
\begin{axis}[
cycle list={{blue,no markers,ultra thick},
			{green,no markers,ultra thick},
			{red,no markers,ultra thick},
			{violet,no markers,ultra thick},
			{blue,dashed,no markers,ultra thin},{blue,dashed,no markers,ultra thin},
            {green,dashed,no markers,ultra thin}, {green,dashed,no markers,ultra thin}, 
            {red,dashed,no markers,ultra thin}, {red,dashed,no markers,ultra thin}, 
			{violet,dashed,no markers,ultra thin}, {violet,dashed,no markers,ultra thin}}, 
width=16.0cm,
height=5.0cm,
xmin=0, xmax=3800, xtick={0,500,...,3500,3800},
xlabel={\nm{t \lrsb{sec}}},
xmajorgrids,
ymin=0, ymax=0.7, ytick={0,0.1,0.2,0.3,0.4,0.5,0.6,0.7},
ylabel={\nm{\DeltarBestnorm \, \lrsb{deg}}},
ymajorgrids,
axis lines=left,
axis line style={-stealth},
legend entries={\nm{\mun{\DeltarBestnorm} \pm \sigman{\DeltarBestnorm}} \small{baseline},
				\small{GYR better},
				\small{GYR worse},
				\small{GYR worst}},
legend columns=4,
legend style={font=\footnotesize},
legend cell align=left,
]
\pgfplotstableread{figs/ch10_ana/versus_gyr/versus_gyr_euler_deg.txt}\mytable
\addplot table [header=false, x index=0,y index=1] {\mytable};
\addplot table [header=false, x index=0,y index=4] {\mytable};
\addplot table [header=false, x index=0,y index=7] {\mytable};
\addplot table [header=false, x index=0,y index=10] {\mytable};
\addplot table [header=false, x index=0,y index=2] {\mytable};
\addplot table [header=false, x index=0,y index=3] {\mytable};
\addplot table [header=false, x index=0,y index=5] {\mytable};
\addplot table [header=false, x index=0,y index=6] {\mytable};
\addplot table [header=false, x index=0,y index=8] {\mytable};
\addplot table [header=false, x index=0,y index=9] {\mytable};
\addplot table [header=false, x index=0,y index=11] {\mytable};
\addplot table [header=false, x index=0,y index=12] {\mytable};
\path node [draw, shape=rectangle, fill=white] at (3250,0.49) {\footnotesize Scenario \nm{\#1}};
\end{axis}   
\end{tikzpicture}
\caption{Influence of gyroscopes quality on body attitude NSE for scenario \nm{\#1} (100 runs)}
\label{fig:Ana_NSE_Results_euler_gyr}
\end{figure}
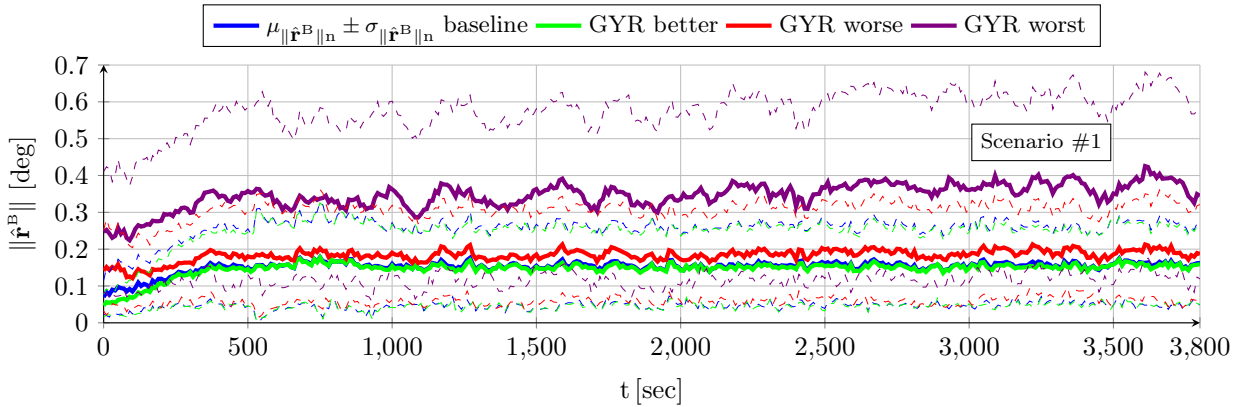

\begin{figure}[h]
\centering
\begin{tikzpicture}
\begin{axis}[
cycle list={{blue,no markers,ultra thick},
			{green,no markers,ultra thick},
			{red,no markers,ultra thick},
			{violet,no markers,ultra thick},
			{blue,dashed,no markers,ultra thin},{blue,dashed,no markers,ultra thin},
            {green,dashed,no markers,ultra thin}, {green,dashed,no markers,ultra thin}, 
            {red,dashed,no markers,ultra thin}, {red,dashed,no markers,ultra thin}, 
			{violet,dashed,no markers,ultra thin}, {violet,dashed,no markers,ultra thin}}, 
width=14.0cm,
height=5.0cm,
xmin=0, xmax=500, xtick={0,50,...,500},
xlabel={\nm{t \lrsb{sec}}},
xmajorgrids,
ymin=0, ymax=0.6, ytick={0,0.1,0.2,0.3,0.4,0.5,0.6},
ylabel={\nm{\DeltarBestnorm \, \lrsb{deg}}},
ymajorgrids,
axis lines=left,
axis line style={-stealth},
]
\pgfplotstableread{figs/ch10_ana/versus_gyr/versus_gyr_alter_euler_deg.txt}\mytable
\addplot table [header=false, x index=0,y index=1] {\mytable};
\addplot table [header=false, x index=0,y index=4] {\mytable};
\addplot table [header=false, x index=0,y index=7] {\mytable};
\addplot table [header=false, x index=0,y index=10] {\mytable};
\addplot table [header=false, x index=0,y index=2] {\mytable};
\addplot table [header=false, x index=0,y index=3] {\mytable};
\addplot table [header=false, x index=0,y index=5] {\mytable};
\addplot table [header=false, x index=0,y index=6] {\mytable};
\addplot table [header=false, x index=0,y index=8] {\mytable};
\addplot table [header=false, x index=0,y index=9] {\mytable};
\addplot table [header=false, x index=0,y index=11] {\mytable};
\addplot table [header=false, x index=0,y index=12] {\mytable};
\path node [draw, shape=rectangle, fill=white] at (420,0.42) {\footnotesize Scenario \nm{\#2}};
\end{axis}   
\end{tikzpicture}
\caption{Influence of gyroscopes quality on body attitude NSE for scenario \nm{\#2} (100 runs)}
\label{fig:Ana_NSE_Results_euler_gyr_alter}
\end{figure}
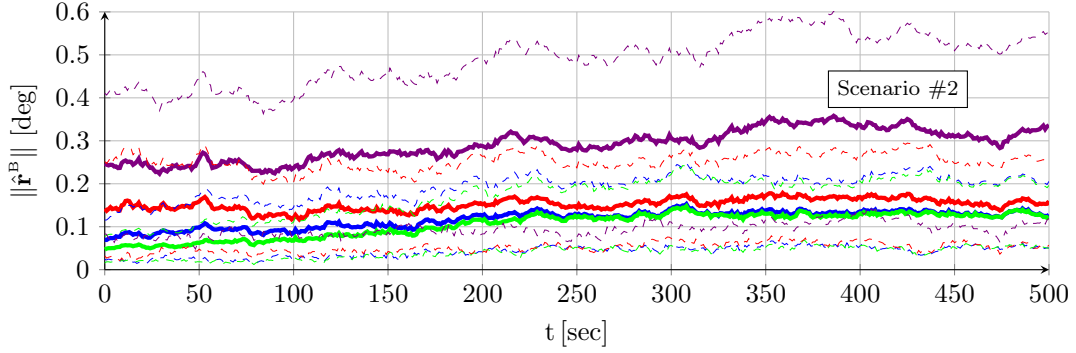

Table \ref{tab:Ana_NSE_Results_euler_gyr}, together with figures \ref{fig:Ana_NSE_Results_euler_gyr} and \ref{fig:Ana_NSE_Results_euler_gyr_alter}\footnote{Note that the baseline columns of table \ref{tab:Ana_NSE_Results_euler_gyr} coincide with the right hand side of table \ref{tab:attitude_errors}, and the blue lines of figures \ref{fig:Ana_NSE_Results_euler_gyr} and \ref{fig:Ana_NSE_Results_euler_gyr_alter} coincide with those of figures \ref{fig:Sim_NSE_Results_euler} and \ref{fig:Sim_NSE_Results_euler_alter}, respectively.}, show the aggregated body attitude NSE after one hundred runs of both scenarios with each of the three configurations described above, in addition to the baseline. There exists a clear correlation between the body attitude NSE and the quality or grade of the gyroscopes, but the proposed attitude filter algorithms manage in all four cases to obtain estimations that are bounded and contain no drift.
\begin{center}
\begin{tabular}{llrrrrrrrrrrrr}
\hline
\multicolumn{2}{l}{Scenario} & \multicolumn{3}{c}{Baseline} & \multicolumn{3}{c}{GYR Better} & \multicolumn{3}{c}{GYR Worse} & \multicolumn{3}{c}{GYR Worst} \\
\multicolumn{2}{l}{\nm{\DeltarBestnorm \nm{\lrsb{deg}}}} & mean & std & max & mean & std & max & mean & std & max & mean & std & max \\
\hline
\multirow{3}{*}{\nm{\#1}} & mean  & \textbf{0.152} &0.078 & 0.52 & \textbf{0.147} & 0.075 & 0.50 & \textbf{0.184} &0.105 & 0.63 & \textbf{0.346} & 0.224 & 1.17 \\
                          & std   & 0.072 & 0.021 & 0.15 & 0.072 & 0.022 & 0.16 & 0.066 & 0.022 & 0.13 & 0.059 & 0.037 & 0.17 \\
                          & max   & 0.448 & 0.168 & 1.32 & 0.450 & 0.169 & 1.35 & 0.452 & 0.189 & 1.23 & 0.532 & 0.319 & 1.70 \\
\hline
\multirow{3}{*}{\nm{\#2}} & mean  & \textbf{0.118} &0.074 & 0.38 & \textbf{0.105} & 0.069 & 0.35 & \textbf{0.152} &0.094 & 0.47 & \textbf{0.291} & 0.177 & 0.80 \\
                          & std   & 0.023 & 0.015 & 0.07 & 0.020 & 0.016 & 0.07 & 0.035 & 0.021 & 0.10 & 0.081 & 0.055 & 0.19 \\
                          & max   & 0.206 & 0.124 & 0.63 & 0.147 & 0.129 & 0.63 & 0.283 & 0.158 & 0.78 & 0.574 & 0.383 & 1.53 \\
\hline
\end{tabular}
\end{center}
\captionof{table}{Influence of gyroscopes quality on aggregated body attitude NSE (100 runs)} \label{tab:Ana_NSE_Results_euler_gyr}

Even taking into consideration that the algorithms are optimized for the baseline configuration (in particular, an adjustment of the attitude filter covariances would likely result in slight improvements for the non baseline configurations), it can be observed that there is little potential for improvement, while on the other side employing gyroscopes of lesser grade does result in a worse estimation of the body attitude. The small improvements obtained when employing gyroscopes that are significantly better than those of the baseline is attributed to the limitations of GNSS-Denied navigation in turbulent conditions, and in particular to the multiple inaccuracies present in the filter specific force observation equation (\ref{eq:filter_att_obs_fIBB}) that prevent the tracking of the full accelerometer error \nm{\EACC} in GNSS-Denied conditions.

As the body attitude is bounded and drift-less, the gyroscopes grade does not have any influence on the estimation accuracy of the ground velocity or the horizontal position. Although not shown, and similarly to the case of the bias drift oscillation band analyzed in the previous section, the results for all four configurations are virtually the same in both scenarios. Note that this is the case even with the significant differences of gyroscope quality considered (two orders of magnitude in bias drift from \say{GYR Better} to \say{GYR Worst}, and smaller but also significant differences in all other parameters), and considering the differences in mission, weather, and wind between the two scenarios. The influence of the gyroscope grade on the vertical position estimation accuracy is also negligible.

\subsubsection*{Influence of Accelerometer Quality}

This section analyzes the sensitivity on the GNSS-Denied navigation results of employing accelerometers of different quality or performance grade than those of the baseline described in \cite{SENSORS} and table \ref{tab:Sensors_gyr_acc}. Table \ref{tab:Ana_NSE_Results_acc_other} shows the performances of the configuration named \say{ACC Better}, which correspond to the high performance MEMS accelerometers inside the Sensonor STIM300 IMU \cite{STIM300}. It also shows the performance parameters of two fictitious accelerometers that perform worse than the baseline. The configurations are named \say{ACC Worse} and \say{ACC Worst}, and their objective is to provide a wide range of accelerometer performances with which to evaluate the sensitivity of the navigation algorithms to the accelerometer grade. Additionally, they represent a safety measure in case of errors in the modeling of the baseline accelerometers performances.
\begin{center}
\begin{tabular}{lc|lclclc}
	\hline
	& & \multicolumn{2}{c}{ACC Better} & \multicolumn{2}{c}{ACC Worse} & \multicolumn{2}{c}{ACC Worst} \\
	Error Source & Variable & \multicolumn{1}{c}{Value} & Unit & \multicolumn{1}{c}{Value} & Unit & \multicolumn{1}{c}{Value} & Unit \\
	\hline
	Bias Drift		& \nm{\sigmauACC}	& \nm{4.90 \cdot 10^{-5}}	& [\nm{m/sec^{2.5}}] & \nm{8.50 \cdot 10^{-5}}	& [\nm{m/sec^{2.5}}] & \nm{1.20 \cdot 10^{-4}}	& [\nm{m/sec^{2.5}}] \\
	System Noise	& \nm{\sigmavACC}	& \nm{3.30 \cdot 10^{-4}}	& [\nm{m/sec^{1.5}}] & \nm{5.00 \cdot 10^{-4}}	& [\nm{m/sec^{1.5}}] & \nm{6.50 \cdot 10^{-4}}	& [\nm{m/sec^{1.5}}] \\
	Scale Factor	& \nm{\sACC}		& \nm{1.50 \cdot 10^{-5}}	& [-] & \nm{8.50 \cdot 10^{-5}}	& [-]                & \nm{1.40 \cdot 10^{-4}}	& [-] \\
	Cross Coupling	& \nm{\mACC}		& \nm{1.50 \cdot 10^{-5}}	& [-] & \nm{5.00 \cdot 10^{-5}}	& [-]                & \nm{9.50 \cdot 10^{-5}}	& [-] \\
	Bias Offset		& \nm{\BzeroACC}	& \nm{1.96 \cdot 10^{-2}}	& [\nm{m/sec^2}] & \nm{4.50 \cdot 10^{-1}}	& [\nm{m/sec^2}]     & \nm{8.50 \cdot 10^{-1}}	& [\nm{m/sec^2}] \\
	\hline
\end{tabular}
\captionof{table}{Performances of ``ACC Better'', ``ACC Worse'', and ``ACC Worst'' accelerometers} \label{tab:Ana_NSE_Results_acc_other}
\end{center}

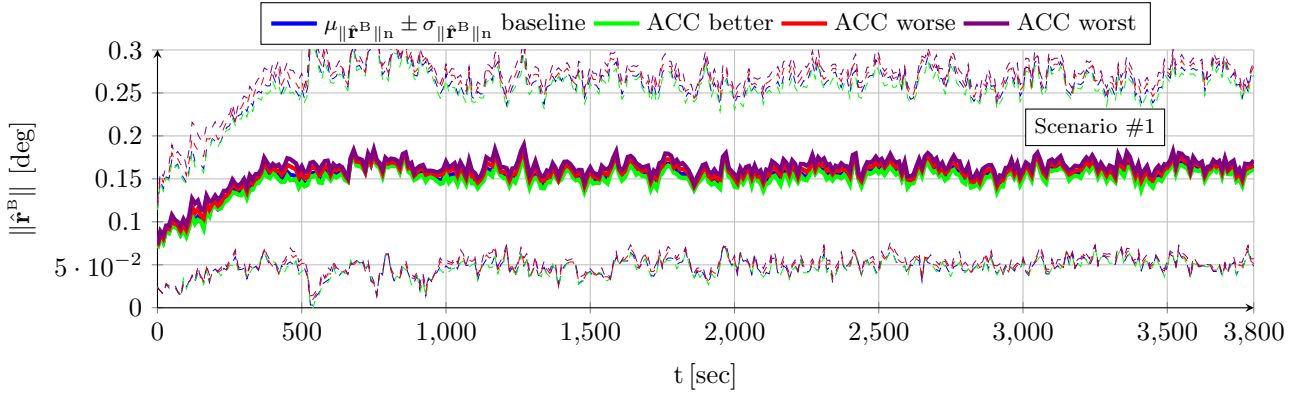
\begin{figure}[h]
\centering
\pgfplotsset{
	every axis legend/.append style={
		at={(0.5,1.02)},
		anchor=south,
	},
}
\begin{tikzpicture}
\begin{axis}[
cycle list={{blue,no markers,ultra thick},
			{green,no markers,ultra thick},
			{red,no markers,ultra thick},
			{violet,no markers,ultra thick},
			{blue,dashed,no markers,ultra thin},{blue,dashed,no markers,ultra thin},
            {green,dashed,no markers,ultra thin}, {green,dashed,no markers,ultra thin}, 
            {red,dashed,no markers,ultra thin}, {red,dashed,no markers,ultra thin}, 
			{violet,dashed,no markers,ultra thin}, {violet,dashed,no markers,ultra thin}}, 
width=16.0cm,
height=5.0cm,
xmin=0, xmax=3800, xtick={0,500,...,3500,3800},
xlabel={\nm{t \lrsb{sec}}},
xmajorgrids,
ymin=0, ymax=0.3, ytick={0,0.05,0.1,0.15,0.2,0.25,0.3},
ylabel={\nm{\DeltarBestnorm \, \lrsb{deg}}},
ymajorgrids,
axis lines=left,
axis line style={-stealth},
legend entries={\nm{\mun{\DeltarBestnorm} \pm \sigman{\DeltarBestnorm}} \small{baseline},
				\small{ACC better},
				\small{ACC worse},
				\small{ACC worst}},
legend columns=4,
legend style={font=\footnotesize},
legend cell align=left,
]
\pgfplotstableread{figs/ch10_ana/versus_acc/versus_acc_euler_deg.txt}\mytable
\addplot table [header=false, x index=0,y index=1] {\mytable};
\addplot table [header=false, x index=0,y index=4] {\mytable};
\addplot table [header=false, x index=0,y index=7] {\mytable};
\addplot table [header=false, x index=0,y index=10] {\mytable};
\addplot table [header=false, x index=0,y index=2] {\mytable};
\addplot table [header=false, x index=0,y index=3] {\mytable};
\addplot table [header=false, x index=0,y index=5] {\mytable};
\addplot table [header=false, x index=0,y index=6] {\mytable};
\addplot table [header=false, x index=0,y index=8] {\mytable};
\addplot table [header=false, x index=0,y index=9] {\mytable};
\addplot table [header=false, x index=0,y index=11] {\mytable};
\addplot table [header=false, x index=0,y index=12] {\mytable};
\path node [draw, shape=rectangle, fill=white] at (3250,0.21) {\footnotesize Scenario \nm{\#1}};
\end{axis}   
\end{tikzpicture}
\caption{Influence of accelerometers quality on body attitude NSE for scenario \nm{\#1} (100 runs)}
\label{fig:Ana_NSE_Results_euler_acc}
\end{figure}

\begin{figure}[h]
\centering
\pgfplotsset{
	every axis legend/.append style={
		at={(0.5,1.02)},
		anchor=south,
	},
}
\begin{tikzpicture}
\begin{axis}[
cycle list={{blue,no markers,ultra thick},
			{green,no markers,ultra thick},
			{red,no markers,ultra thick},
			{violet,no markers,ultra thick},
			{blue,dashed,no markers,ultra thin},{blue,dashed,no markers,ultra thin},
            {green,dashed,no markers,ultra thin}, {green,dashed,no markers,ultra thin}, 
            {red,dashed,no markers,ultra thin}, {red,dashed,no markers,ultra thin}, 
			{violet,dashed,no markers,ultra thin}, {violet,dashed,no markers,ultra thin}}, 
width=14.0cm,
height=5.0cm,
xmin=0, xmax=500, xtick={0,50,...,500},
xlabel={\nm{t \lrsb{sec}}},
xmajorgrids,
ymin=0, ymax=0.3, ytick={0,0.05,0.1,0.15,0.2,0.25,0.3},
ylabel={\nm{\DeltarBestnorm \, \lrsb{deg}}},
ymajorgrids,
axis lines=left,
axis line style={-stealth},
legend entries={\nm{\mun{\DeltarBestnorm} \pm \sigman{\DeltarBestnorm}} \small{baseline},
				\small{ACC better},
				\small{ACC worse},
				\small{ACC worst}},
legend columns=4,
legend style={font=\footnotesize},
legend cell align=left,
]
\pgfplotstableread{figs/ch10_ana/versus_acc/versus_acc_alter_euler_deg.txt}\mytable
\addplot table [header=false, x index=0,y index=1] {\mytable};
\addplot table [header=false, x index=0,y index=4] {\mytable};
\addplot table [header=false, x index=0,y index=7] {\mytable};
\addplot table [header=false, x index=0,y index=10] {\mytable};
\addplot table [header=false, x index=0,y index=2] {\mytable};
\addplot table [header=false, x index=0,y index=3] {\mytable};
\addplot table [header=false, x index=0,y index=5] {\mytable};
\addplot table [header=false, x index=0,y index=6] {\mytable};
\addplot table [header=false, x index=0,y index=8] {\mytable};
\addplot table [header=false, x index=0,y index=9] {\mytable};
\addplot table [header=false, x index=0,y index=11] {\mytable};
\addplot table [header=false, x index=0,y index=12] {\mytable};
\path node [draw, shape=rectangle, fill=white] at (70,0.25) {\footnotesize Scenario \nm{\#2}};
\end{axis}   
\end{tikzpicture}
\caption{Influence of accelerometers quality on body attitude NSE for scenario \nm{\#2} (100 runs)}
\label{fig:Ana_NSE_Results_euler_acc_alter}
\end{figure}
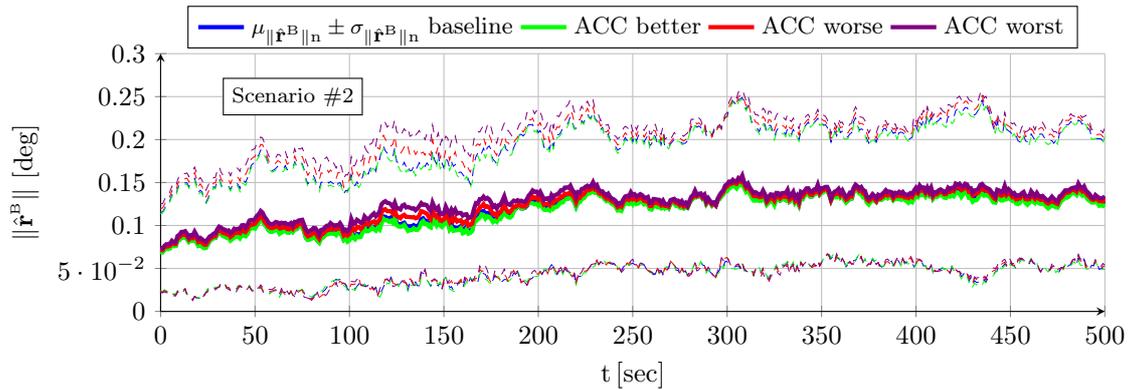

Table \ref{tab:Ana_NSE_Results_euler_acc}, together with figures \ref{fig:Ana_NSE_Results_euler_acc} and \ref{fig:Ana_NSE_Results_euler_acc_alter}\footnote{Note that the baseline columns of table \ref{tab:Ana_NSE_Results_euler_acc} coincide with the right hand side of table \ref{tab:attitude_errors}, and the blue lines of figures \ref{fig:Ana_NSE_Results_euler_acc} and \ref{fig:Ana_NSE_Results_euler_acc_alter} coincide with those of figures \ref{fig:Sim_NSE_Results_euler} and \ref{fig:Sim_NSE_Results_euler_alter}, respectively.}, show the aggregated body attitude NSE after one hundred runs of both scenarios with each of the three configurations described above, in addition to the baseline. Although technically present, the dependency of the body attitude estimation accuracy with the grade of the accelerometers is so small than it can be considered nonexistent for all practical purposes.
\begin{center}
\begin{tabular}{llrrrrrrrrrrrr}
\hline
\multicolumn{2}{l}{Scenario} & \multicolumn{3}{c}{Baseline} & \multicolumn{3}{c}{ACC Better} & \multicolumn{3}{c}{ACC Worse} & \multicolumn{3}{c}{ACC Worst} \\
\multicolumn{2}{l}{\nm{\DeltarBestnorm \lrsb{deg}}} & mean & std & max & mean & std & max & mean & std & max & mean & std & max \\
\hline
\multirow{3}{*}{\nm{\#1}} & mean  & \textbf{0.152} &0.078 & 0.52 & \textbf{0.150} & 0.078 & 0.52 & \textbf{0.157} &0.080 & 0.53 & \textbf{0.162} & 0.081 & 0.53 \\
                          & std   & 0.072 & 0.021 & 0.15 & 0.069 & 0.020 & 0.15 & 0.073 & 0.021 & 0.15 & 0.074 & 0.021 & 0.15 \\
                          & max   & 0.448 & 0.168 & 1.32 & 0.447 & 0.168 & 1.32 & 0.452 & 0.170 & 1.31 & 0.462 & 0.172 & 1.31 \\
\hline
\multirow{3}{*}{\nm{\#2}} & mean  & \textbf{0.118} &0.074 & 0.38 & \textbf{0.116} & 0.073 & 0.37 & \textbf{0.121} &0.076 & 0.38 & \textbf{0.126} & 0.078 & 0.39 \\
                          & std   & 0.023 & 0.015 & 0.07 & 0.021 & 0.015 & 0.07 & 0.026 & 0.016 & 0.07 & 0.031 & 0.017 & 0.08 \\
                          & max   & 0.206 & 0.124 & 0.63 & 0.179 & 0.131 & 0.65 & 0.254 & 0.125 & 0.63 & 0.307 & 0.138 & 0.72 \\
\hline
\end{tabular}
\end{center}
\captionof{table}{Influence of accelerometers quality on aggregated body attitude NSE (100 runs)} \label{tab:Ana_NSE_Results_euler_acc}

The reason for this behavior lies on the low accuracy of the (\ref{eq:filter_att_obs_fIBB}) attitude filter observations in GNSS-Denied conditions, which do not allow the navigation filter to track the full accelerometer error \nm{\EACC}. Discarding the airspeed and turbulence time derivatives viewed in the body frame (\nm{\vTASBdot, \, \vTURBBdot}) as well as the wind time derivative viewed in NED (\nm{\vWINDNdot}) is what provides the attitude filter with relatively unbiased, albeit noisier, observations that enable it to track the body attitude without drift for long periods of time. Although not considered in the observation equation, these time derivatives exist and are included in the measurements. In particular, the wind turbulence accelerations considered in both scenarios are much higher than the effect of the accelerometers bias drift and system noise in all four accelerometer configurations. The accelerometer grade hence would only play a meaningful role in the body attitude estimation results if flying in very calm conditions with barely any turbulence or if their noise and drift levels were so high that they became comparable to the turbulence levels.

In this sense the neglected time derivatives within the navigation filter specific force observation equation (\ref{eq:filter_att_obs_fIBB}), of which the wind turbulence acceleration is by far the most significant, can be considered as additional sources of error within the accelerometers with which to measure the specific force. If their oscillations are bigger than the accelerometers intrinsic errors, as is the case, then the effect of the later on the measurements is barely noticeable.

As explained in previous sections, the estimation of the ground velocity and horizontal position does not depend on the body attitude as long as the later is bounded and drift-less, which is the case in this analysis. No results are shown for the estimation of the ground velocity or horizontal position in either scenario because they are virtually identical for each of the four configurations. Note that this is the case even with the significant differences of quality among the accelerometers considered (approximately an order of magnitude in most performance parameters between \say{ACC Better} and \say{ACC Worst}), and taking into account the differences in mission, weather, and wind between the two scenarios. The influence of the accelerometer grade on the vertical position estimation accuracy is also negligible. 


\subsubsection*{Influence of Magnetometer Quality}

This section analyzes the sensitivity on the GNSS-Denied navigation results of employing magnetometers of different quality or performance grade than those of the baseline described in \cite{SENSORS} and table \ref{tab:Sensors_mag}. Table \ref{tab:Ana_NSE_Results_mag_other} shows the performances of two fictitious magnetometers with performances better (\say{MAG Better}) and worse (\say{MAG Worse}) than those of the baseline. As in previous sections, the objective is to obtain a sufficiently wide range of magnetometer performance grades with which to evaluate the sensitivity of the navigation results when executing both scenarios.
\begin{center}
\begin{tabular}{lc|lclc}
	\hline
	& & \multicolumn{2}{c}{MAG Better} & \multicolumn{2}{c}{MAG Worse} \\
	Error Source & Variable & \multicolumn{1}{c}{Value} & Unit & \multicolumn{1}{c}{Value} & Unit \\	
	\hline
	System Noise				& \nm{\sigmavMAG}	& \nm{3.00 \cdot 10^{0}}	& [\nm{nT \cdot sec^{0.5}}] & \nm{1.00 \cdot 10^{1}}	& [\nm{nT \cdot sec^{0.5}}] \\
	Scale Factor \& Soft Iron	& \nm{\sMAG}		& \nm{5.00 \cdot 10^{-4}}	& [-]                       & \nm{1.25 \cdot 10^{-3}}	& [-] \\
	Cross Coupling \& Soft Iron	& \nm{\mMAG}		& \nm{7.00 \cdot 10^{-4}}	& [-]                       & \nm{1.50 \cdot 10^{-3}}	& [-] \\
	Hard Iron 					& \nm{\BhiMAG}		& \nm{1.25 \cdot 10^{2}} 	& [\nm{nT}]                 & \nm{3.50 \cdot 10^{2}} 	& [\nm{nT}] \\
	Bias Offset					& \nm{\BzeroMAG}	& \nm{3.00 \cdot 10^{2}} 	& [\nm{nT}]                 & \nm{7.50 \cdot 10^{2}} 	& [\nm{nT}] \\
	\hline
\end{tabular}
\captionof{table}{Performances of ``MAG Better'' and ``MAG Worse'' magnetometers} \label{tab:Ana_NSE_Results_mag_other}
\end{center}

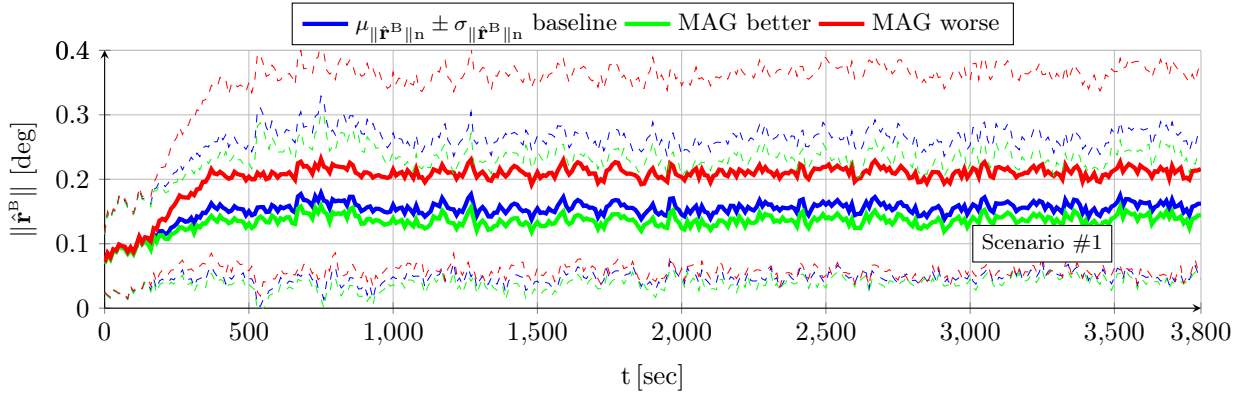
\begin{figure}[h]
\centering
\pgfplotsset{
	every axis legend/.append style={
		at={(0.5,1.02)},
		anchor=south,
	},
}
\begin{tikzpicture}
\begin{axis}[
cycle list={{blue,no markers,ultra thick},
			{green,no markers,ultra thick},
			{red,no markers,ultra thick},
			{blue,dashed,no markers,ultra thin},{blue,dashed,no markers,ultra thin},
            {green,dashed,no markers,ultra thin}, {green,dashed,no markers,ultra thin}, 
            {red,dashed,no markers,ultra thin}, {red,dashed,no markers,ultra thin}}, 
width=16.0cm,
height=5.0cm,
xmin=0, xmax=3800, xtick={0,500,...,3500,3800},
xlabel={\nm{t \lrsb{sec}}},
xmajorgrids,
ymin=0, ymax=0.4, ytick={0,0.1,0.2,0.3,0.4,0.4},
ylabel={\nm{\DeltarBestnorm \, \lrsb{deg}}},
ymajorgrids,
axis lines=left,
axis line style={-stealth},
legend entries={\nm{\mun{\DeltarBestnorm} \pm \sigman{\DeltarBestnorm}} \small{baseline},
				\small{MAG better},
				\small{MAG worse}},
legend columns=3,
legend style={font=\footnotesize},
legend cell align=left,
]
\pgfplotstableread{figs/ch10_ana/versus_mag/versus_mag_euler_deg.txt}\mytable
\addplot table [header=false, x index=0,y index=1] {\mytable};
\addplot table [header=false, x index=0,y index=4] {\mytable};
\addplot table [header=false, x index=0,y index=7] {\mytable};
\addplot table [header=false, x index=0,y index=2] {\mytable};
\addplot table [header=false, x index=0,y index=3] {\mytable};
\addplot table [header=false, x index=0,y index=5] {\mytable};
\addplot table [header=false, x index=0,y index=6] {\mytable};
\addplot table [header=false, x index=0,y index=8] {\mytable};
\addplot table [header=false, x index=0,y index=9] {\mytable};
\path node [draw, shape=rectangle, fill=white] at (3250,0.1) {\footnotesize Scenario \nm{\#1}};
\end{axis}   
\end{tikzpicture}
\caption{Influence of magnetometers quality on body attitude NSE for scenario \nm{\#1} (100 runs)}
\label{fig:Ana_NSE_Results_euler_mag}
\end{figure}

\begin{figure}[h]
\centering
\pgfplotsset{
	every axis legend/.append style={
		at={(0.5,1.02)},
		anchor=south,
	},
}
\begin{tikzpicture}
\begin{axis}[
cycle list={{blue,no markers,ultra thick},
			{green,no markers,ultra thick},
			{red,no markers,ultra thick},
			{blue,dashed,no markers,ultra thin},{blue,dashed,no markers,ultra thin},
            {green,dashed,no markers,ultra thin}, {green,dashed,no markers,ultra thin}, 
            {red,dashed,no markers,ultra thin}, {red,dashed,no markers,ultra thin}}, 
width=14.0cm,
height=5.0cm,
xmin=0, xmax=500, xtick={0,50,...,500},
xlabel={\nm{t \lrsb{sec}}},
xmajorgrids,
ymin=0, ymax=0.3, ytick={0,0.05,0.1,0.15,0.2,0.25,0.3},
ylabel={\nm{\DeltarBestnorm \, \lrsb{deg}}},
ymajorgrids,
axis lines=left,
axis line style={-stealth},
legend entries={\nm{\mun{\DeltarBestnorm} \pm \sigman{\DeltarBestnorm}} \small{baseline},
				\small{MAG better},
				\small{MAG worse}},
legend columns=3,
legend style={font=\footnotesize},
legend cell align=left,
]
\pgfplotstableread{figs/ch10_ana/versus_mag/versus_mag_alter_euler_deg.txt}\mytable
\addplot table [header=false, x index=0,y index=1] {\mytable};
\addplot table [header=false, x index=0,y index=4] {\mytable};
\addplot table [header=false, x index=0,y index=7] {\mytable};
\addplot table [header=false, x index=0,y index=2] {\mytable};
\addplot table [header=false, x index=0,y index=3] {\mytable};
\addplot table [header=false, x index=0,y index=5] {\mytable};
\addplot table [header=false, x index=0,y index=6] {\mytable};
\addplot table [header=false, x index=0,y index=8] {\mytable};
\addplot table [header=false, x index=0,y index=9] {\mytable};
\path node [draw, shape=rectangle, fill=white] at (100,0.25)  {\footnotesize Scenario \nm{\#2}};
\end{axis}   
\end{tikzpicture}
\caption{Influence of magnetometers quality on body attitude NSE for scenario \nm{\#2} (100 runs)}
\label{fig:Ana_NSE_Results_euler_mag_alter}
\end{figure}
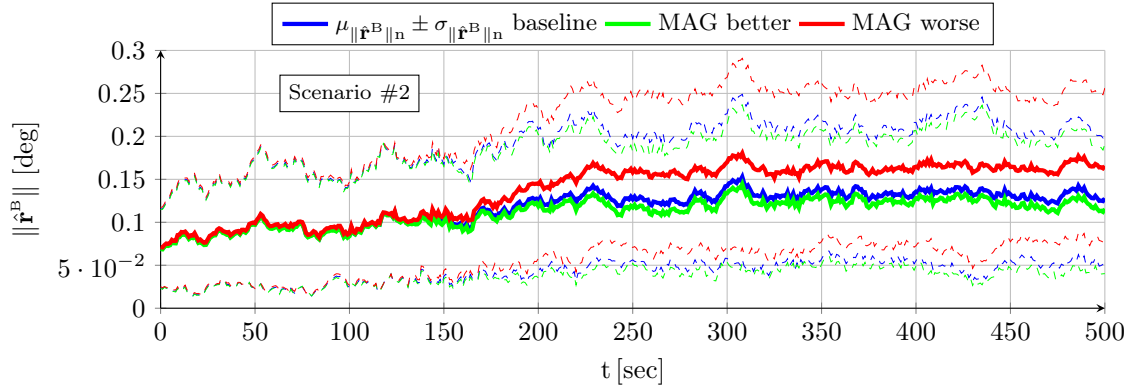

Table \ref{tab:Ana_NSE_Results_euler_mag}, together with figures \ref{fig:Ana_NSE_Results_euler_mag} and \ref{fig:Ana_NSE_Results_euler_mag_alter}\footnote{Note that the baseline columns of table \ref{tab:Ana_NSE_Results_euler_mag} coincide with the right hand side of table \ref{tab:attitude_errors}, and the blue lines of figures \ref{fig:Ana_NSE_Results_euler_mag} and \ref{fig:Ana_NSE_Results_euler_mag_alter} coincide with those of figures \ref{fig:Sim_NSE_Results_euler} and \ref{fig:Sim_NSE_Results_euler_alter}, respectively.}, show the body attitude NSE for the benchmark and alternate scenarios in the same format as in previous sections for each of the two configurations described above, in addition to the baseline. There exists a clear correlation between the magnitude of the attitude estimation errors and the quality or grade of the magnetometers, but the proposed attitude filter algorithms manage in all three cases to obtain estimations that are bounded and contain no drift.
\begin{center}
\begin{tabular}{llrrrrrrrrr}
\hline
\multicolumn{2}{l}{Scenario} & \multicolumn{3}{c}{Baseline} & \multicolumn{3}{c}{MAG Better} & \multicolumn{3}{c}{MAG Worse}\\
\multicolumn{2}{l}{\nm{\DeltarBestnorm \lrsb{deg}}} & mean & std & max & mean & std & max & mean & std & max\\
\hline
\multirow{3}{*}{\nm{\#1}} & mean  & \textbf{0.152} &0.078 & 0.521 & \textbf{0.134} & 0.075 & 0.499 & \textbf{0.203} & 0.087 & 0.579 \\
                          & std   & 0.072 & 0.021 & 0.151 & 0.060 & 0.019 & 0.141 & 0.120 & 0.030 & 0.191 \\
                          & max   & 0.448 & 0.168 & 1.321 & 0.452 & 0.170 & 1.274 & 0.855 & 0.263 & 1.453 \\
\hline
\multirow{3}{*}{\nm{\#2}} & mean  & \textbf{0.118} &0.074 & 0.376 & \textbf{0.112} & 0.072 & 0.367 & \textbf{0.138} &0.082 & 0.411 \\
                          & std   & 0.023 & 0.015 & 0.069 & 0.022 & 0.014 & 0.069 & 0.029 & 0.020 & 0.078 \\
                          & max   & 0.206 & 0.124 & 0.632 & 0.198 & 0.122 & 0.617 & 0.227 & 0.138 & 0.676 \\
\hline
\end{tabular}
\end{center}
\captionof{table}{Influence of magnetometers quality on aggregated body attitude NSE (100 runs)} \label{tab:Ana_NSE_Results_euler_mag}

Comparing these results with those obtained above when varying the grade of the gyroscopes, they are quite similar, in the sense that there is relatively little potential for improvement but significant downside to using magnetometers of inferior grade. This is attributed to the limitations of GNSS-Denied navigation in turbulent conditions, and in particular to the multiple inaccuracies present in the filter specific force observation equation (\ref{eq:filter_att_obs_fIBB}) that prevent the tracking of the full accelerometer error \nm{\EACC} in GNSS-Denied conditions.

As in the gyroscopes case, the differences in body attitude estimation accuracy are too small to have any influence on the accuracy of the ground velocity and horizontal positions. Although the results are not shown, they are virtually the same for all three configurations in both scenarios. The influence of the magnetometer grade on the vertical position estimation accuracy is also negligible.

\subsubsection*{Influence of Airspeed Sensors Quality}

This section analyzes the influence on the navigation results of employing a Pitot tube and air vanes of different quality to those used in the baseline and described in \cite{SENSORS} and table \ref{tab:Sensors_vtasb_air}. To do so, this analysis employs two different fictitious settings, named \say{TAS-AOA-AOS Better} and \say{TAS-AOA-AOS Worse}, whose system noise and bias offsets are listed in table \ref{tab:Ana_NSE_Results_vtas_better_worse}. The former configuration represents the performances of higher SWaP or better quality sensors, while the later may be more realistic for smaller platforms in which a lower SWaP system is required.
\begin{center}
\begin{tabular}{ll|rcrc}
	\hline
	& & \multicolumn{2}{c}{TAS-AOA-AOS Better} & \multicolumn{2}{c}{TAS-AOA-AOS Worse} \\
	Error Source & Variable & \multicolumn{1}{c}{Value} & Unit & \multicolumn{1}{c}{Value} & Unit \\	
	\hline
	Airspeed System Noise	        & \nm{\sigmaTAS}	& \nm{1.50 \cdot 10^{-1}}	& [\nm{m/sec}] & \nm{6.66 \cdot 10^{-1}}	& [\nm{m/sec}] \\
	Airspeed Bias Offset			& \nm{\BzeroTAS}	& \nm{1.50 \cdot 10^{-1}}	& [\nm{m/sec}] & \nm{6.66 \cdot 10^{-1}}	& [\nm{m/sec}] \\
	Angle of Attack System Noise    & \nm{\sigmaAOA}	& \nm{1.00 \cdot 10^{-1}}	& [\nm{deg}]   & \nm{6.66 \cdot 10^{-1}}	& [\nm{deg}]   \\
	Angle of Attack Bias Offset		& \nm{\BzeroAOA}	& \nm{1.00 \cdot 10^{-1}}	& [\nm{deg}]   & \nm{6.66 \cdot 10^{-1}}	& [\nm{deg}]   \\
	Angle of Sideslip System Noise	& \nm{\sigmaAOS}	& \nm{1.00 \cdot 10^{-1}}	& [\nm{deg}]   & \nm{6.66 \cdot 10^{-1}}	& [\nm{deg}]   \\
	Angle of Sideslip Bias Offset	& \nm{\BzeroAOS}	& \nm{1.00 \cdot 10^{-1}}	& [\nm{deg}]   & \nm{6.66 \cdot 10^{-1}}	& [\nm{deg}]   \\
	\hline
\end{tabular}
\captionof{table}{Performances of ``TAS-AOA-AOS Better'' and ``TAS-AOA-AOS Worse'' air data systems} \label{tab:Ana_NSE_Results_vtas_better_worse}
\end{center}

As previously explained, the air data filter is capable of mostly eliminating the system noise when estimating the components of the airspeed vector, but can not remove the bias offsets, which are incorporated into the estimations. These are biased on a trajectory basis, zero mean when aggregated, and always bounded by the sensor quality. Table \ref{tab:Ana_NSE_Results_vtas_alpha_beta_vtasb} validates these conclusions and shows progressively better results as the quality of the airspeed (TAS), angle of attack (AOA), and angle of sideslip (AOS) sensors improves. Only the scenario \nm{\#1} results are shown, as the behavior of the air data filter is similar on both scenarios.
\begin{center}
\begin{tabular}{lrrrrrrrrr}
\hline
Scenario \nm{\#1} & \multicolumn{3}{c}{Baseline} & \multicolumn{3}{c}{TAS-AOA-AOS Better} & \multicolumn{3}{c}{TAS-AOA-AOS Worse} \\
\nm{\vtasest - \vtas \, \lrsb{10^{-1} \cdot m/sec}} & mean & std & max & mean & std & max & mean & std & max \\
\hline
mean & -0.30 & \textbf{0.78} & 5.78 & -0.14 & \textbf{0.53} & 3.41 & -0.45 & \textbf{0.96} & 7.79 \\
std  & 3.01 & 0.01 & 1.78 & 1.36 & 0.00 & 0.80 & 4.52 & 0.01 & 2.68 \\
max  & -7.74 & 0.80 & -11.07 & -3.49 & 0.55 & -5.87 & -11.63 & 1.00 & -16.18 \\
\hline
\nm{\alphaest - \alpha \, \lrsb{10^{-1} \cdot deg}} & mean & std & max & mean & std & max & mean & std & max \\
\hline
mean  & +0.32 & \textbf{0.94} & 6.83 & +0.14 & \textbf{0.66} & 4.18 & +0.48 & \textbf{1.17} & 9.13 \\
std   &  3.25 & 0.04 & 2.02 & 1.46 & 0.04 & 0.90 & 4.88 & 0.03 & 2.99 \\
max   & -8.39 & 1.04 & +12.44 & -3.78 & 0.76 & +7.17 & -12.59 & 1.26 & +17.74 \\
\hline
\nm{\betaest - \beta \, \lrsb{10^{-1} \cdot deg}} & mean & std & max & mean & std & max & mean & std & max \\
\hline
mean & -0.25 & \textbf{1.75} & 10.47 & -0.11 & \textbf{1.08} & 6.02 & -0.38 & \textbf{2.23} & 13.95 \\
std  & 3.35 & 0.24 & 2.07 & 1.51 & 0.07 & 0.91 & 5.03 & 0.67 & 4.27 \\
max  & +8.66 & 3.02 & +15.92 & +3.90 & 1.37 & +8.50 & +13.00 & 6.85 & +39.49 \\
\hline
\end{tabular}
\end{center}
\captionof{table}{Influence of airspeed sensor quality on aggregated airspeed vector NSE for scenario \nm{\#1} (100 runs)} \label{tab:Ana_NSE_Results_vtas_alpha_beta_vtasb}

\begin{figure}[h]
\centering
\pgfplotsset{
	every axis legend/.append style={
		at={(0.5,1.02)},
		anchor=south,
	},
}
\begin{tikzpicture}
\begin{axis}[
cycle list={{blue,no markers,ultra thick},
			{green,no markers,ultra thick},
			{red,no markers,ultra thick},
			{blue,dashed,no markers,ultra thin},{blue,dashed,no markers,ultra thin},
            {green,dashed,no markers,ultra thin}, {green,dashed,no markers,ultra thin}, 
            {red,dashed,no markers,ultra thin}, {red,dashed,no markers,ultra thin}}, 
width=16.0cm,
height=5.0cm,
xmin=0, xmax=3800, xtick={0,500,...,3500,3800},
xlabel={\nm{t \lrsb{sec}}},
xmajorgrids,
ymin=0, ymax=0.3, ytick={0,0.05,0.1,0.15,0.2,0.25,0.3},
ylabel={\nm{\DeltarBestnorm \, \lrsb{deg}}},
ymajorgrids,
axis lines=left,
axis line style={-stealth},
legend entries={\nm{\mun{\DeltarBestnorm} \pm \sigman{\DeltarBestnorm}} \small{baseline},
				\small{TAS-AOA-AOS better},
				\small{TAS-AOA-AOS worse}},
legend columns=3,
legend style={font=\footnotesize},
legend cell align=left,
]
\pgfplotstableread{figs/ch10_ana/versus_vtasb/versus_vtasb_euler_deg.txt}\mytable
\addplot table [header=false, x index=0,y index=1] {\mytable};
\addplot table [header=false, x index=0,y index=4] {\mytable};
\addplot table [header=false, x index=0,y index=7] {\mytable};
\addplot table [header=false, x index=0,y index=2] {\mytable};
\addplot table [header=false, x index=0,y index=3] {\mytable};
\addplot table [header=false, x index=0,y index=5] {\mytable};
\addplot table [header=false, x index=0,y index=6] {\mytable};
\addplot table [header=false, x index=0,y index=8] {\mytable};
\addplot table [header=false, x index=0,y index=9] {\mytable};
\path node [draw, shape=rectangle, fill=white] at (3200,0.10) {\footnotesize Scenario \nm{\#1}};
\end{axis}   
\end{tikzpicture}
\caption{Influence of air speed sensor quality on body attitude NSE for scenario \nm{\#1} (100 runs)}
\label{fig:Ana_NSE_Results_euler_vtasb}
\end{figure}
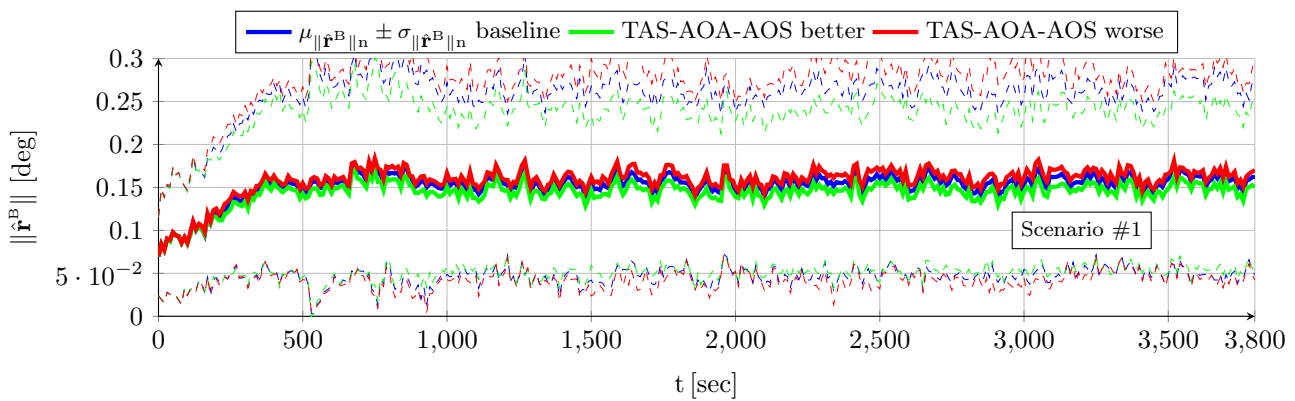

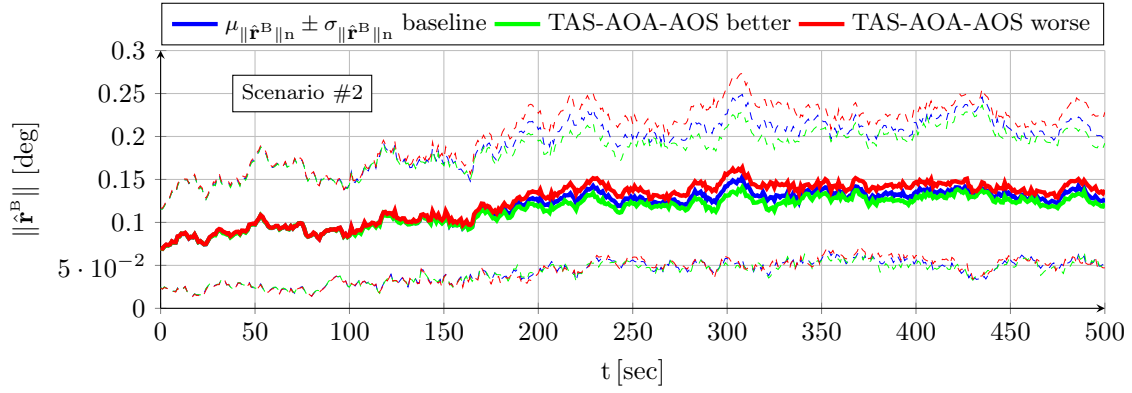
\begin{figure}[h]
\centering
\pgfplotsset{
	every axis legend/.append style={
		at={(0.5,1.02)},
		anchor=south,
	},
}
\begin{tikzpicture}
\begin{axis}[
cycle list={{blue,no markers,ultra thick},
			{green,no markers,ultra thick},
			{red,no markers,ultra thick},
			{blue,dashed,no markers,ultra thin},{blue,dashed,no markers,ultra thin},
            {green,dashed,no markers,ultra thin}, {green,dashed,no markers,ultra thin}, 
            {red,dashed,no markers,ultra thin}, {red,dashed,no markers,ultra thin}}, 
width=14.0cm,
height=5.0cm,
xmin=0, xmax=500, xtick={0,50,...,500},
xlabel={\nm{t \lrsb{sec}}},
xmajorgrids,
ymin=0, ymax=0.3, ytick={0,0.05,0.1,0.15,0.2,0.25,0.3},
ylabel={\nm{\DeltarBestnorm \, \lrsb{deg}}},
ymajorgrids,
axis lines=left,
axis line style={-stealth},
legend entries={\nm{\mun{\DeltarBestnorm} \pm \sigman{\DeltarBestnorm}} \small{baseline},
				\small{TAS-AOA-AOS better},
				\small{TAS-AOA-AOS worse}},
legend columns=3,
legend style={font=\footnotesize},
legend cell align=left,
]
\pgfplotstableread{figs/ch10_ana/versus_vtasb/versus_vtasb_alter_euler_deg.txt}\mytable
\addplot table [header=false, x index=0,y index=1] {\mytable};
\addplot table [header=false, x index=0,y index=4] {\mytable};
\addplot table [header=false, x index=0,y index=7] {\mytable};
\addplot table [header=false, x index=0,y index=2] {\mytable};
\addplot table [header=false, x index=0,y index=3] {\mytable};
\addplot table [header=false, x index=0,y index=5] {\mytable};
\addplot table [header=false, x index=0,y index=6] {\mytable};
\addplot table [header=false, x index=0,y index=8] {\mytable};
\addplot table [header=false, x index=0,y index=9] {\mytable};
\path node [draw, shape=rectangle, fill=white] at (75,0.25) {\footnotesize Scenario \nm{\#2}};
\end{axis}   
\end{tikzpicture}
\caption{Influence of air speed sensor quality on body attitude NSE for scenario \nm{\#2} (100 runs)}
\label{fig:Ana_NSE_Results_euler_vtasb_alter}
\end{figure}

The influence of the airspeed sensors quality on the body attitude for both scenarios is summarized in table \ref{tab:Ana_NSE_Results_euler_vtasb}, together with figures \ref{fig:Ana_NSE_Results_euler_vtasb} and \ref{fig:Ana_NSE_Results_euler_vtasb_alter}\footnote{Note that the baseline columns of table \ref{tab:Ana_NSE_Results_euler_vtasb} coincide with the right hand side of table \ref{tab:attitude_errors}, and the blue lines of figures \ref{fig:Ana_NSE_Results_euler_vtasb} and \ref{fig:Ana_NSE_Results_euler_vtasb_alter} coincide with those of figures \ref{fig:Sim_NSE_Results_euler} and \ref{fig:Sim_NSE_Results_euler_alter}, respectively.}. There is a clear correlation between the sensors quality and the aggregated body attitude NSE, which is caused by the accuracy of the attitude filter observations, in particular those of \nm{{\vec p}_n} (\ref{eq:filter_att_obs_pvec}).
\begin{center}
\begin{tabular}{llrrrrrrrrr}
\hline
\multicolumn{2}{l}{Scenario} & \multicolumn{3}{c}{Baseline} & \multicolumn{3}{c}{TAS-AOA-AOS Better} & \multicolumn{3}{c}{TAS-AOA-AOS Worse}\\
\multicolumn{2}{l}{\nm{\DeltarBestnorm \nm{\lrsb{deg}}}} & mean & std & max & mean & std & max & mean & std & max\\
\hline
\multirow{3}{*}{\nm{\#1}} & mean & \textbf{0.152} &0.078 & 0.521 & \textbf{0.144} & 0.076 & 0.505 & \textbf{0.158} & 0.080 & 0.532 \\
                          & std  & 0.072 & 0.021 & 0.151 & 0.056 & 0.017 & 0.135 & 0.084 & 0.026 & 0.165 \\
                          & max  & 0.448 & 0.168 & 1.321 & 0.364 & 0.138 & 1.319 & 0.504 & 0.205 & 1.296 \\
\hline
\multirow{3}{*}{\nm{\#2}} & mean & \textbf{0.118} &0.074 & 0.376 & \textbf{0.113} & 0.070 & 0.365 & \textbf{0.124} &0.079 & 0.396 \\
                          & std  & 0.023 & 0.015 & 0.069 & 0.021 & 0.014 & 0.066 & 0.027 & 0.018 & 0.080 \\
                          & max  & 0.206 & 0.124 & 0.632 & 0.188 & 0.110 & 0.562 & 0.224 & 0.135 & 0.681 \\
\hline
\end{tabular}
\end{center}
\captionof{table}{Influence of air speed sensor quality on aggregated body attitude NSE (100 runs)}\label{tab:Ana_NSE_Results_euler_vtasb}

The influence of the Pitot tube and air vanes quality on the ground velocity estimation is shown in table \ref{tab:Ana_NSE_Results_vned_vtasb}\footnote{Note that the baseline columns of table \ref{tab:Ana_NSE_Results_vned_vtasb} coincide with those of table \ref{tab:v_errors}.}. It is small for scenario \nm{\#1}, in which the effects of the airspeed vector are masked by the much bigger influence of the wind change from the time the GNSS are lost, but significant for scenario \nm{\#2}, in which the wind remains constant throughout the trajectory. In any case, this behavior is different to that obtained when varying the quality of gyroscopes, accelerometers, and magnetometers in previous sections, and the reason lies in the use of \nm{\vTASNest} for the estimation of the ground velocity \nm{\vNest} in the position filter. The influence of both \nm{\vTASBest} and \nm{\qNBest}, which together make up \nm{\vTASNest}, is rather small because of the back and forth compensation performed by the position filter when it loses the GNSS signals, but small errors in \nm{\vTASBest} have a bigger influence in \nm{\vTASNest} than those of \nm{\qNBest}, specially in the case of pitch and roll where the angles are very small. For this reason, improvements on the body attitude estimation \nm{\qNBest} by themselves do not imply any noticeable change in the estimation of the ground velocity, as shown in previous sections, while when combined with improvements in the airspeed vector estimation \nm{\vTASBest}, improvements in the aggregated ground velocity NSE become noticeable. 
\begin{center}
\begin{tabular}{llrrrrrr}
\hline
\multicolumn{2}{l}{Scenario} & \multicolumn{2}{c}{Baseline} & \multicolumn{2}{c}{TAS-AOA-AOS Better} & \multicolumn{2}{c}{TAS-AOA-AOS Worse} \\
\multicolumn{2}{l}{\nm{\vNest \lrp{\tEND} - \vN \lrp{\tEND} \lrsb{m/sec}}} & \nm{\vNesti - \vNi} & \nm{\vNestii - \vNii} & \nm{\vNesti - \vNi} & \nm{\vNestii - \vNii} & \nm{\vNesti - \vNi} & \nm{\vNestii - \vNii} \\
\hline
\multirow{3}{*}{\nm{\#1}} & mean  &  +0.325 & +0.465 & +0.335 & +0.477 & +0.311 & +0.450 \\
                          & std   & \textbf{2.588} & \textbf{2.204} & \textbf{2.582} & \textbf{2.163} & \textbf{2.601} & \textbf{2.253} \\
                          & max   & +9.321 & +7.392 & +9.132 & +7.568 & +9.448 & +7.267\\
\hline
\multirow{3}{*}{\nm{\#2}} & mean & -0.083 & -0.139 & -0.047 & -0.102 & -0.114 & -0.165 \\
                          & std  & \textbf{0.652} & \textbf{0.629} & \textbf{0.570} & \textbf{0.575} & \textbf{0.771} & \textbf{0.721} \\
                          & max  & -1.916 & -1.792 & -1.497 & +1.404 & +2.357 & -2.163\\
\hline
\end{tabular}
\end{center}
\captionof{table}{Influence of air speed sensor quality on final aggregated ground speed NSE (100 runs)} \label{tab:Ana_NSE_Results_vned_vtasb}

\begin{center}
\begin{tabular}{llrrrrrrr} 
\hline
\multicolumn{2}{c}{Scenario} & & \multicolumn{2}{c}{Baseline} & \multicolumn{2}{c}{TAS-AOA-AOS Better} & \multicolumn{2}{c}{TAS-AOA-AOS Worse} \\
& &  Distance \nm{\lrsb{m}} & \multicolumn{2}{c}{\nm{\Deltaxhorest \lrp{\tEND} \lrsb{m,\%}}} & \multicolumn{2}{c}{\nm{\Deltaxhorest \lrp{\tEND} \lrsb{m,\%}}} & \multicolumn{2}{c}{\nm{\Deltaxhorest \lrp{\tEND} \lrsb{m,\%}}} \\
\hline
\multirow{3}{*}{\nm{\#1}} & mean & 108622.6 & 7427.7 & \textbf{7.18} & 7291.2 & \textbf{7.05} & 7667.0 & \textbf{7.41} \\
                          & std  & 19935.3 & 4987.8 & 5.73 & 4913.1 & 5.65 & 5030.5 & 5.70 \\
                          & max  & 172842.3 & 25287.9 & 32.38 & 25291.8 & 31.30 & 25320.0 & 33.39 \\
\hline
\multirow{3}{*}{\nm{\#2}} & mean & 14198.4 & 215.6 & \textbf{1.52} & 198.0 & \textbf{1.40} & 245.6 & \textbf{1.74} \\
                          & std  & 1176.1 & 119.4 & 0.86 & 105.8 & 0.75 & 137.0 & 1.00 \\
                          & max  & 18253.0 & 586.3 & 4.38 & 513.2 & 3.65 & 672.4 & 5.33 \\
\hline
\end{tabular}
\end{center}
\captionof{table}{Influence of air speed sensor quality on final aggregated horizontal position NSE (100 runs)} \label{tab:Ana_NSE_Results_xhor_vtasb}

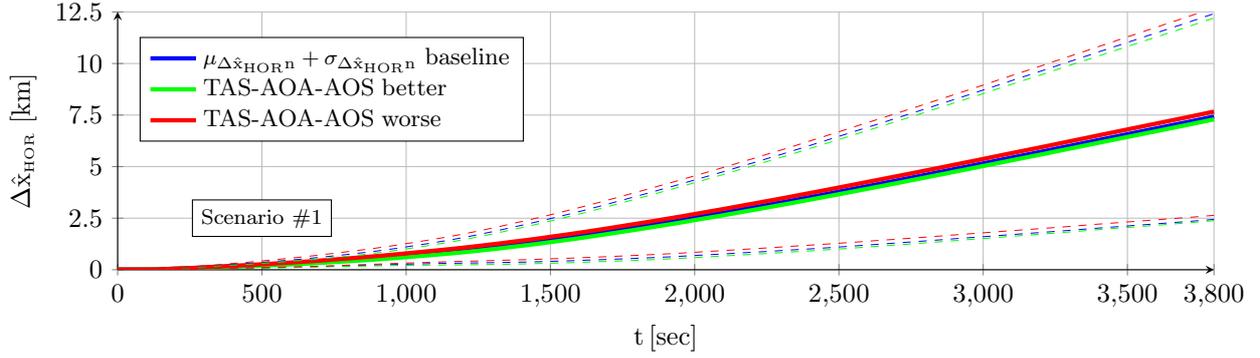
\begin{figure}[h]
\centering
\pgfplotsset{
	every axis legend/.append style={
		at={(0.02,0.7)},
		anchor=west,
	},
}
\begin{tikzpicture}
\begin{axis}[
cycle list={{blue,no markers,ultra thick},
			{green,no markers,ultra thick},
			{red,no markers,ultra thick},
			{blue,dashed,no markers,ultra thin},{blue,dashed,no markers,ultra thin},
            {green,dashed,no markers,ultra thin}, {green,dashed,no markers,ultra thin}, 
			{red,dashed,no markers,ultra thin}, {red,dashed,no markers,ultra thin}},
width=16.0cm,
height=5.0cm,
xmin=0, xmax=3800, xtick={0,500,...,3500,3800},
xlabel={\nm{t \lrsb{sec}}},
xmajorgrids,
ymin=0, ymax=12.5, ytick={0,2.5,...,12.5},
ylabel={\nm{\Deltaxhorest \, \lrsb{km}}},
ymajorgrids,
axis lines=left,
axis line style={-stealth},
legend entries={\nm{\mun{\Deltaxhorest} + \sigman{\Deltaxhorest}} \small{baseline},
				\small{TAS-AOA-AOS better},
				\small{TAS-AOA-AOS worse}},
legend columns=1,
legend style={font=\footnotesize},
legend cell align=left,
]
\pgfplotstableread{figs/ch10_ana/versus_vtasb/versus_vtasb_pos_hor_m_pc.txt}\mytable
\addplot table [header=false, x index=0,y index=1] {\mytable};
\addplot table [header=false, x index=0,y index=4] {\mytable};
\addplot table [header=false, x index=0,y index=7] {\mytable};
\addplot table [header=false, x index=0,y index=2] {\mytable};
\addplot table [header=false, x index=0,y index=3] {\mytable};
\addplot table [header=false, x index=0,y index=5] {\mytable};
\addplot table [header=false, x index=0,y index=6] {\mytable};
\addplot table [header=false, x index=0,y index=8] {\mytable};
\addplot table [header=false, x index=0,y index=9] {\mytable};
\path node [draw, shape=rectangle, fill=white] at (500,2.5) {\footnotesize Scenario \nm{\#1}};
\end{axis}   
\end{tikzpicture}
\caption{Influence of air speed sensor quality on horizontal position NSE for scenario \nm{\#1} (100 runs)}
\label{fig:Ana_NSE_Results_xhor_vtasb}
\end{figure}

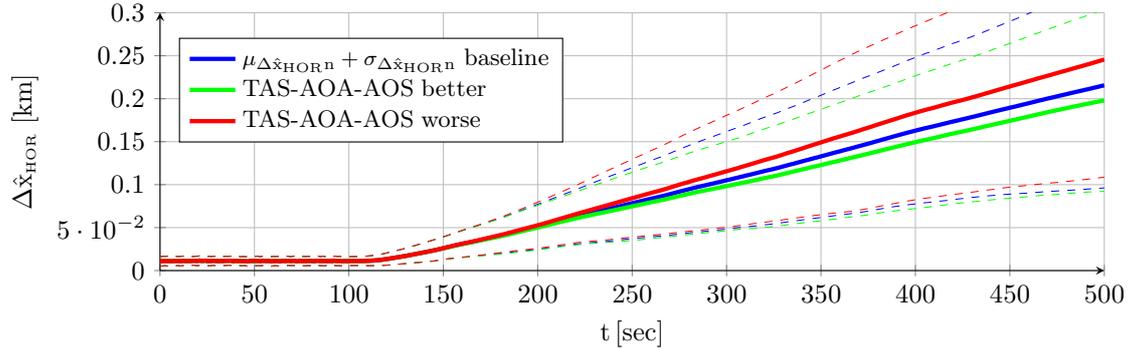
\begin{figure}[h]
\centering
\pgfplotsset{
	every axis legend/.append style={
		at={(0.02,0.7)},
		anchor=west,
	},
}
\begin{tikzpicture}
\begin{axis}[
cycle list={{blue,no markers,ultra thick},
			{green,no markers,ultra thick},
			{red,no markers,ultra thick},
			{blue,dashed,no markers,ultra thin},{blue,dashed,no markers,ultra thin},
            {green,dashed,no markers,ultra thin}, {green,dashed,no markers,ultra thin}, 
			{red,dashed,no markers,ultra thin}, {red,dashed,no markers,ultra thin}},
width=14.0cm,
height=5.0cm,
xmin=0, xmax=500, xtick={0,50,...,500},
xlabel={\nm{t \lrsb{sec}}},
xmajorgrids,
ymin=0, ymax=0.3, ytick={0,0.05,0.1,0.15,0.2,0.25,0.3},
ylabel={\nm{\Deltaxhorest \, \lrsb{km}}},
ymajorgrids,
axis lines=left,
axis line style={-stealth},
legend entries={\nm{\mun{\Deltaxhorest} + \sigman{\Deltaxhorest}} \small{baseline},
				\small{TAS-AOA-AOS better},
				\small{TAS-AOA-AOS worse}},
legend columns=1,
legend style={font=\footnotesize},
legend cell align=left,
]
\pgfplotstableread{figs/ch10_ana/versus_vtasb/versus_vtasb_alter_pos_hor_m_pc.txt}\mytable
\addplot table [header=false, x index=0,y index=1] {\mytable};
\addplot table [header=false, x index=0,y index=4] {\mytable};
\addplot table [header=false, x index=0,y index=7] {\mytable};
\addplot table [header=false, x index=0,y index=2] {\mytable};
\addplot table [header=false, x index=0,y index=3] {\mytable};
\addplot table [header=false, x index=0,y index=5] {\mytable};
\addplot table [header=false, x index=0,y index=6] {\mytable};
\addplot table [header=false, x index=0,y index=8] {\mytable};
\addplot table [header=false, x index=0,y index=9] {\mytable};
\path node [draw, shape=rectangle, fill=white] at (100,0.2) {\footnotesize Scenario \nm{\#2}};
\end{axis}   
\end{tikzpicture}
\caption{Influence of air speed sensor quality on horizontal position NSE for scenario \nm{\#2} (100 runs)}
\label{fig:Ana_NSE_Results_xhor_vtasb_alter}
\end{figure}

Table \ref{tab:Ana_NSE_Results_xhor_vtasb}, together with figures \ref{fig:Ana_NSE_Results_xhor_vtasb} and \ref{fig:Ana_NSE_Results_xhor_vtasb_alter}\footnote{Note that the baseline columns of table \ref{tab:Ana_NSE_Results_xhor_vtasb} can also be found in table \ref{tab:hor_errors}, while the blue lines of figures \ref{fig:Ana_NSE_Results_xhor_vtasb} and \ref{fig:Ana_NSE_Results_xhor_vtasb_alter} are the same as those of figures \ref{fig:Sim_NSE_Results_xhor} and \ref{fig:Sim_NSE_Results_xhor_alter}, respectively.} show the the influence of the quality of the airspeed vector sensors on the final aggregated horizontal position NSE. The slight changes in ground velocity estimation accuracy are integrated and translate into small changes in the accuracy of the horizontal position estimation, although they are small when compared to the influence of the wind speed changes since the time the GNSS signals are lost. Although no data is shown, the influence of the airspeed sensors on the vertical position estimation accuracy is negligible.

\subsubsection*{Influence of Atmospheric Sensors Quality}

This section analyzes the influence on the navigation results of employing atmospheric sensors (barometer and thermometer) of different quality to those used in the baseline configuration and described in \cite{SENSORS} and table \ref{tab:Sensors_vtasb_air}. As neither the body attitude nor the horizontal position estimation errors depend on the quality of these sensors, this analysis employs two different fictitious settings, named \say{OSP-OAT Worse} and \say{OSP-OAT Worst}, whose system noise and bias offsets are listed in table \ref{tab:Ana_NSE_Results_air_worse_worst}, with the objective of determining if the lack of influence is also the case with sensors significantly worse than those employed in the baseline configuration.
\begin{center}
\begin{tabular}{ll|rcrc}
	\hline
	& & \multicolumn{2}{c}{OSP-OAT Worse} & \multicolumn{2}{c}{OSP-OAT Worst} \\
	Error Source & Variable & \multicolumn{1}{c}{Value} & Unit & \multicolumn{1}{c}{Value} & Unit \\	
	\hline
	Pressure System Noise		& \nm{\sigmaOSP}	& \nm{1.50 \cdot 10^{+2}}	& [\nm{pa}]        & \nm{3.00 \cdot 10^{+2}}	& [\nm{pa}]\\
	Pressure Bias Offset   	    & \nm{\BzeroOSP}	& \nm{1.50 \cdot 10^{+2}}	& [\nm{pa}]        & \nm{3.00 \cdot 10^{+2}}	& [\nm{pa}] \\	
	Temperature System Noise	& \nm{\sigmaOAT}	& \nm{1.50 \cdot 10^{-1}}   & [\nm{^{\circ}K}] & \nm{5.00 \cdot 10^{-1}}    & [\nm{^{\circ}K}] \\
	Temperature Bias Offset		& \nm{\BzeroOAT}	& \nm{1.50 \cdot 10^{-1}}	& [\nm{^{\circ}K}] & \nm{5.00 \cdot 10^{-1}}	& [\nm{^{\circ}K}] \\
	\hline
\end{tabular}
\captionof{table}{Performances of ``OSP-OAT Worse'' and ``OSP-OAT Worst'' atmospheric sensors} \label{tab:Ana_NSE_Results_air_worse_worst}
\end{center}

As explained above, the air data filter is capable of mostly eliminating the system noise when estimating the pressure altitude and the atmospheric temperature, but can not remove the bias offsets, which are incorporated into the estimations. These are biased on a trajectory basis, zero mean when aggregated, and always bounded by the sensor quality. Table \ref{tab:Ana_NSE_Results_T_Hp_DeltaT_air} validates these conclusions for scenario \nm{\#1} and shows progressively worse results as the quality of the pressure (OSP) and temperature (OAT) sensors diminishes. A similar conclusion can be applied to the temperature offset estimation \nm{\DeltaTest}, which is just a combination of the previous two (\ref{eq:filter_air_DeltaT}) and hence shows the same dependencies. Only the scenario \nm{\#1} results are shown, as the behavior of the air data filter is similar on both scenarios.
\begin{center}
\begin{tabular}{lrrrrrrrrr}
\hline
Aggregated Errors & \multicolumn{3}{c}{Baseline} & \multicolumn{3}{c}{OSP-OAT Worse} & \multicolumn{3}{c}{OSP-OAT Worst} \\
\nm{\Test - T \, \lrsb{10^{-1} \cdot ^{\circ}K}} & mean & std & max & mean & std & max & mean & std & max \\
\hline
mean & -0.06 & \textbf{0.16} & 1.14 & -0.17 & \textbf{0.27} & 2.48 & -0.58 & \textbf{0.50} & 6.30 \\
std  & 0.51 & 0.00 & 0.30 & 1.52 & 0.00 & 0.90 & 5.07 & 0.00 & 3.00 \\
max  & -1.52 & 0.16 & -2.21 & -4.56 & 0.28 & -5.70 & -15.20 & 0.51 & -17.26 \\
\hline
\nm{\Hpest - \Hp \, \lrsb{m}} & mean & std & max & mean & std & max & mean & std & max \\
\hline
mean & +0.97 & \textbf{0.93} & 12.92 & +1.45 & \textbf{1.29} & 18.91 & +2.88 & \textbf{2.26} & 36.33 \\
std  & 11.26 & 0.05 & 6.73 & 16.89 & 0.07 & 10.13 & 33.77 & 0.16 & 20.27 \\
max  & +29.60 & 1.11 & +33.59 & +44.37 & 1.57 & +50.00 & +88.54 & 2.86 & +98.82 \\
\hline
\nm{\DeltaTest - \DeltaT \, \lrsb{10^{-1} \cdot ^{\circ}K}} & mean & std & max & mean & std & max & mean & std & max \\
\hline
mean & +0.01 & \textbf{0.17} & 1.52 & -0.08 & \textbf{0.29} & 2.85 & -0.39 & \textbf{0.52} & 6.82 \\
std  & 0.92 & 0.00 & 0.56 & 1.94 & 0.00 & 1.16 & 5.66 & 0.01 & 3.38 \\
max  & -2.59 & 0.17 & -3.34 & -6.16 & 0.29 & -7.35 & -18.40 & 0.54 & -20.61 \\
\hline
\end{tabular}
\end{center}
\captionof{table}{Influence of atmospheric sensors quality on aggregated \nm{T}, \nm{\Hp}, and \nm{\DeltaT} NSE for scenario \nm{\#1} (100 runs)} \label{tab:Ana_NSE_Results_T_Hp_DeltaT_air}

The body attitude estimation is for all practical purposes independent of the atmospheric sensors quality, as shown in table \ref{tab:Ana_NSE_Results_euler_air}\footnote{Note that the baseline columns of table \ref{tab:Ana_NSE_Results_euler_air} coincide with the right hand side of table \ref{tab:attitude_errors}.}. Although no data is shown, the influence on the ground velocity and horizontal position estimation accuracy is negligible.
\begin{center}
\begin{tabular}{llrrrrrrrrr}
\hline
\multicolumn{2}{l}{Scenario} & \multicolumn{3}{c}{Baseline} & \multicolumn{3}{c}{OSP-OAT Worse} & \multicolumn{3}{c}{OSP-OAT Worst} \\
\multicolumn{2}{l}{\nm{\DeltarBestnorm \lrsb{deg}}} & mean & std & max & mean & std & max & mean & std & max \\
\hline
\multirow{3}{*}{\nm{\#1}} & mean & \textbf{0.152} &0.078 & 0.521 & \textbf{0.153} & 0.079 & 0.523 & \textbf{0.156} & 0.080 & 0.530 \\
                          & std  & 0.072 & 0.021 & 0.151 & 0.072 & 0.021 & 0.151 & 0.072 & 0.021 & 0.150 \\
                          & max  & 0.448 & 0.168 & 1.321 & 0.448 & 0.169 & 1.313 & 0.450 & 0.170 & 1.295 \\
\hline
\end{tabular}
\end{center}
\captionof{table}{Influence of atmospheric sensors quality on aggregated body attitude NSE for scenario \nm{\#1} (100 runs)} \label{tab:Ana_NSE_Results_euler_air}

The previous analysis indicates that the vertical position estimation error depends on two factors: the ionospheric errors and the change in pressure offset \nm{\Deltap} from the time the GNSS signals are lost. The results obtained with different atmospheric sensors, shown in table \ref{tab:Ana_NSE_Results_h_air}, validate these conclusions as they do not show any influence on the accuracy of the vertical position estimations.
\begin{center}
\begin{tabular}{llccc}
\hline
\multicolumn{2}{l}{\nm{\hest \lrp{\tEND} - h \lrp{\tEND} \ \lrsb{m}}} & Baseline & OSP-OAT Worse & OSP-OAT Worst \\
\hline
\multirow{3}{*}{\nm{\#1}} & mean  & -3.89 & -3.87 & -4.01 \\
                          & std   & \textbf{26.03} & \textbf{26.01} & \textbf{26.07} \\
                          & max   & -70.49 & -69.75 & -68.00 \\
\hline						  
\end{tabular}
\end{center}
\captionof{table}{Influence of atmospheric sensors quality on aggregated vertical position NSE for scenario \nm{\#1} (100 runs)} \label{tab:Ana_NSE_Results_h_air}


\section{Summary and Conclusions} \label{sec:Conclusions}

This article proposes an inertial navigation algorithm that takes advantage of sensors such as magnetometers, Pitot tube, and air vanes, already present in autonomous fixed wing low SWaP aircraft, to improve their GNSS-Denied navigation capabilities, as well s to facilitate the fusion between the inertial filter and visual odometry algorithms. The results obtained when applying the proposed algorithms to high fidelity Monte Carlo simulations of two scenarios representative of the challenges of GNSS-Denied navigation indicate the following:
\begin{itemize}
\item The \textbf{body attitude} estimation shows no drift with time. It is lightly biased for each trajectory (for each Euler angle as well as for the total attitude error); when aggregated for multiple executions, the estimation is unbiased (zero mean) for each Euler angle and biased for the total error. The aggregated errors are highly repeatable and can be considered bounded. There is no need for maneuvers, and the existing ones do not introduce error spikes. The aircraft can hence remain aloft in GNSS-Denied conditions for as long as it has fuel. 
	
\item The \textbf{vertical position} estimation shows an error that depends on two factors: the ionospheric effects (which also apply when GNSS signals are available) and the change in atmospheric pressure offset \nm{\Deltap} from its value at the time the GNSS signals are lost. Both factors are random and hence the final error is unbiased (zero mean) when aggregated; both factors are also bounded by atmospheric physics, and hence so is the estimation error. The estimation shows no drift with time and the magnitude of the bounded error indicates it is valid for all navigation purposes except landing.

\item The \textbf{ground velocity} estimation shows an error that mostly depends on the wind change from its value at the time the GNSS signals are lost. Airspeed vector and body attitude estimation accuracy also play a minor role. The three factors are random and hence the final error is unbiased (zero mean) when aggregated. The wind change is bounded by atmospheric physics, while the airspeed vector and body attitude errors are bounded by sensor quality; the estimation error hence is also bounded. The estimation shows no drift with time.
	
\item The \textbf{horizontal position} estimation exhibits a drift with time that depends exclusively on the ground velocity estimation error; when this error is constant, the drift is linear with time. The drift results in an unbounded error that is not valid for navigation purposes due to excessive collision risk.
\end{itemize}

When comparing the GNSS-Denied navigation results obtained with the proposed filter against those resulting from more traditional approaches to inertial navigation designed to rely on velocity and position observations provided by the GNSS receiver that suddenly are not present, there are significant improvements in the estimation of body attitude (specially during maneuvers), vertical position (where drift with time is replaced by a bounded error), and horizontal position (where the pace of drift or error growth with time is drastically reduced). The following observations can be made:
\begin{itemize}
\item The proposed approach of \textbf{freezing the wind speed estimation} \nm{\vNest} at its value when the GNSS signals are lost is the preferred choice for reducing horizontal position errors as it relies on a single integration. Alternative approaches using a double integration may be superior for the first few minutes of GNSS-Denied conditions, but overall result in a much faster growth of the horizontal position error with time.

\item The proposed approach of employing the pressure altitude estimation and \textbf{freezing the atmospheric pressure offset estimation} \nm{\Deltapest} at its value when the GNSS signals are lost is the preferred choice for reducing vertical position errors as it does not rely on any integrations. Alternative approaches require at least one integration and result in unbounded estimation errors that grow with time. 

\item The proposed approach of \textbf{discarding the airspeed and turbulence time derivatives} viewed in the body frame (\nm{\vTASBdot = 0}, \nm{\vTURBBdot = 0}) as well as the \textbf{wind field time derivative} viewed in NED (\nm{\vWINDNdot = 0}) is the preferred choice for reducing body attitude estimation errors, specially during maneuvers. Alternative approaches are more sensitive to the lack of absolute velocity observations in GNSS-Denied conditions and result in significant attitude estimation errors when maneuvering.
\end{itemize}

A sensitivity analysis of the influence of the GNSS-Denied navigation results with respect to the quality or grade of the onboard sensors proves that the above results are generic and not specific to certain sensor performance parameters. The main conclusions are the following:
\begin{itemize}
\item The \textbf{vertical position} estimation error is independent of the quality of all sensors and depends exclusively on ionospheric effects and the pressure offset change since the time the GNSS signals are lost. 

\item The \textbf{horizontal position} estimation error shows a slight dependency on the quality of the Pitot tube and air vanes, but not that of any other sensor. It depends mostly on the wind change since the time the GNSS signals are lost. 
	
\item The body \textbf{attitude estimation} error depends on the quality of gyroscopes, magnetometers, Pitot tube, and air vanes, but not that of accelerometers and air data sensors. Taking the baseline as reference, there is little potential for improvement, but a clear correlation between lower sensor grade and higher errors. Its bounded nature however is independent of sensor grade. Note that body attitude estimation accuracy is key for the fusion between inertial and visual odometry algorithms.
\end{itemize}

The results indicate that the proposed navigation filter, which relies on algorithms specifically designed for GNSS-Denied conditions, is capable of significantly improving the navigation of autonomous fixed wing low SWaP aircraft that suddenly can not rely on the signals provided by GNSS. It does so by making use of sensors already present onboard the aircraft but traditionally used for control and not for navigation, by taking advantage of the particularities of fixed wing flight, and by employing the GNSS signals until they become unavailable to track certain atmospheric parameters. Although the inherent horizontal position drift can not be fully eliminated, it is significantly reduced; additionally, the body attitude can be estimated accurately and drift-less even when maneuvering, which facilitates the inclusion of cameras into the sensor mix and visual algorithms into the navigation filter.

\appendix 

\section{Time Derivative of the Aircraft Velocity} \label{sec:DerivativeVelocity}

This appendix derives expressions for the time derivative of the aircraft ground velocity that form the basis for the establishment of the attitude filter equations in section \ref{sec:Filter}. Equation (\ref{eq:MOT_Comp_LinearAccFinal}) represents the relationship among the relative linear accelerations of the origins of three different frames, where \say{0} is considered inertial, with their relative velocities \nm{\vec v} and positions \nm{\vec T}, as well as their relative angular velocities \nm{\vec \omega} and accelerations \nm{\vec \alpha}. Note that all magnitudes are evaluated from the frame corresponding to the first sub index to that of the second. Its obtainment is explained in \cite{SENSORS}:
\neweq{\vec a_{02} = \vec a_{12} + \lrp{\vec a_{01} + \widehat{\vec \alpha}_{01} \; \vec T_{12} + \widehat{\vec \omega}_{01} \; \widehat{\vec \omega}_{01} \;  \vec T_{12}} +  2 \; \widehat{\vec \omega}_{01} \; \vec v_{12}} {eq:MOT_Comp_LinearAccFinal}

Its application to the aircraft considering \nm{F_{2}} as the body frame \nm{\FB}, \nm{F_{1}} as the Earth Centered Earth Fixed ECEF frame \nm{\FE}, and \nm{F_{0}} as the inertial frame \nm{\FI}, all viewed in \nm{\FI}, results in:
\neweq{\vec a_{\sss IB}^{\sss I} = \vec a_{\sss EB}^{\sss I} + \lrp{ \vec a_{\sss IE}^{\sss I} + \alphaIEIskew \; \vec x_{\sss EB}^{\sss I} + \wIEIskew \; \wIEIskew \; \vec x_{\sss EB}^{\sss I} } +  2 \; \wIEIskew \; \vEBI} {eq:EquationsMotion_equations_force2}

The first term at the right hand side of (\ref {eq:EquationsMotion_equations_force2}) is the linear acceleration \nm{\vec a_{\sss EB}^{\sss I}}, which is the time derivative of the aircraft absolute velocity \nm{\vec v_{\sss EB}} viewed in the inertial frame. Making use of the rotation matrix \nm{\vec R} expressions for the transformation of vectors and its relationship with the angular velocity \cite{Sola2017,Shuster1993}, it can be converted into a function of the aircraft and motion angular velocities (\nm{\wEB = \wNB + \wEN}) and the aircraft absolute velocity \nm{\vec v = \vec v_{\sss EB}}:
\begin{eqnarray}
\nm{\vec a_{\sss EB}^{\sss I}} & = & \nm{\RIE \; \vec a_{\sss EB}^{\sss E} = \RIE \; \vEBEdot= \RIE \; \deriv{\lrp{\REB \; \vEBB}} = \RIE \; \lrp{\REBdot \; \vEBB + \REB \; \vEBBdot}} \nonumber \\
& = & \nm{\RIB \; \lrp{\wEBBskew \; \vEBB + \vEBBdot} \ \ \longrightarrow \ \ \vec a_{\sss EB}^{\sss B} = \wEBBskew \; \vEBB + \vEBBdot = \wEBBskew \; \vB + \vBdot} \label{eq:EquationsMotion_equations_force3}
\end{eqnarray}

Noting that \nm{\vEB = \vEN = \vvec}, the same process can be repeated replacing the \nm{\FB} body frame by the \nm{\FN} NED frame:
\begin{eqnarray}
\nm{\vec a_{\sss EB}^{\sss I}} & = & \nm{\RIE \; \vec a_{\sss EB}^{\sss E} = \RIE \; \vEBEdot = \RIE \; \vENEdot = \RIE \; \deriv{\lrp{\REN \; \vENN}} = \RIE \; \lrp{\RENdot \; \vENN + \REN \; \vENNdot}} \nonumber \\
& = & \nm{\RIN \; \lrp{\wENNskew \; \vENN + \vENNdot} \ \ \longrightarrow \ \ \vec a_{\sss EB}^{\sss N} = \wENNskew \; \vENN + \vENNdot = \wENNskew \; \vN + \vNdot} \label{eq:EquationsMotion_equations_force3_N}
\end{eqnarray}

The second term of (\ref{eq:EquationsMotion_equations_force2}), called the \emph{transport acceleration} \nm{\atrs}, can be simplified because the linear acceleration \nm{\vec a_{\sss IE}} of the Earth frame \nm{\FE} with respect to \nm{\FI} is zero as both have their origins located at the Earth center of mass, while the angular acceleration \nm{\vec \alpha_{\sss IE}} is also zero as \nm{\FE} rotates around \nm{\iEiii} at a constant rate \nm{\omegaE}.
\neweq{\atrsI = \wIEIskew \; \wIEIskew \; \vec x_{\sss EB}^{\sss I}} {eq:EquationsMotion_equations_force4} 

The transport acceleration hence coincides with the \emph{centripetal acceleration} caused by the Earth rotation around its symmetry axis, this is, it is the opposite of the centrifugal acceleration \nm{\ac}. Its \texttt{NED} form \nm{\atrsN} can be directly obtained from the Earth angular velocity and the aircraft coordinates, where \nm{\varphi} is the latitude, \nm{h} the altitude over the \texttt{WGS84} ellipsoid, and \nm{N} is the radius of curvature of the prime vertical:
\neweq{\atrsN = - \ac^{\sss N} = \wIENskew \; \wIENskew \; \vec x_{\sss EB}^{\sss N} = \lrsb{\omegaE^2 \; \lrp{N + h} \; \sin \varphi \; \cos \varphi, \, 0, \, \omegaE^2 \; \lrp{N + h} \, \cos^2 \varphi}^T}{eq:EquationsMotion_equations_force5}

The third term of (\ref{eq:EquationsMotion_equations_force2}) is called the \emph{Coriolis acceleration} \nm{\acor}. Its \texttt{NED} form \nm{\acorN} can also be obtained from the Earth angular velocity \nm{\wIE} and the aircraft absolute velocity \nm{\vEB = \vvec}:
\neweq{\acorN =  2 \; \wIENskew \; \vEBN = \lrsb{2 \; \omegaE \; \vNii \; \sin \varphi, \, 2 \; \omegaE \; \lrp{- \vNi \; \sin \varphi - \vNiii \; \cos \varphi}, \, 2 \; \omegaE \; \vNii \; \cos \varphi}^T}{eq:EquationsMotion_equations_force6}

The application of Newton's \nm{\second} law to the motion of the aircraft represented by its body frame results in
\neweq{\sum {\vec F}^{\sss I} = \deriv{\lrp{m \; \vIB}^{\sss I}} = m \; \vIBIdot = m \; \vec a_{\sss IB}^{\sss I}} {eq:EquationsMotion_equations_force1}

where the mass variation with time is considered slow enough as to be quasi stationary. The introduction of the previous terms into (\ref{eq:EquationsMotion_equations_force1}) while converting to \nm{\FB} or \nm{\FN} results in (\ref{eq:EquationsMotion_equations_force7}) and (\ref{eq:EquationsMotion_equations_force7_N}), where \nm{\FAER} and \nm{\FPRO} are the aircraft aerodynamic and propulsive forces, while \nm{\vec g} represents the gravitational acceleration:
\begin{eqnarray}
\nm{\sum \vec F^{\sss B}} & = & \nm{\RBI \; \sum \vec F^{\sss I} = \FAERB + \FPROB + m \ \qNBast \otimes \gN \otimes \qNB} \nonumber \\
& = & \nm{ m \, \lrp{\wEBBskew \, \vB + \vBdot + \qNBast \otimes \atrsN \otimes \qNB + \qNBast \otimes \acorN \otimes \qNB}}\label{eq:EquationsMotion_equations_force7} \\
\nm{\sum \vec F^{\sss N}} & = & \nm{\RNI \; \sum \vec F^{\sss I} = \qNB \otimes \lrp{\FAERB + \FPROB} \otimes \qNBast + m \ \gN} \nonumber \\
& = & \nm{ m \, \lrp{\wENNskew \, \vN + \vNdot + \atrsN + \acorN}}\label{eq:EquationsMotion_equations_force7_N}
\end{eqnarray}

As the gravity acceleration combines the gravitational and centrifugal accelerations (\nm{\gc = \vec g + \ac = \vec g - \atrs}) and \nm{\vEB = \vEN = \vvec}, the final expressions for the time derivative of the aircraft linear velocity are the following:
\begin{eqnarray}
\nm{\vBdot} & = & \nm{\dfrac{\FAERB + \FPROB}{m} - \wEBBskew \; \vB + \qNBast \otimes \lrp{\gcN - \acorN} \otimes \qNB} \label{eq:EquationsMotion_equations_force8} \\
\nm{\vNdot} & = & \nm{\qNB \otimes \dfrac{\FAERB + \FPROB}{m} \otimes \qNBast - \wENNskew \; \vN + \gcN - \acorN} \label{eq:EquationsMotion_equations_force8_N}
\end{eqnarray}

\bibliographystyle{ieeetr}   
\bibliography{inertial_navigation}

\end{document}